\documentclass[10pt,twocolumn,letterpaper]{article}

\usepackage[pagenumbers,algorithms]{wacv}   

\usepackage{microtype}
\usepackage{graphicx}
\usepackage{subcaption}
\usepackage{booktabs}
\usepackage{makecell}
\usepackage{multirow}
\usepackage{amsmath}
\usepackage{amssymb}
\usepackage{mathtools}
\usepackage{amsthm}
\usepackage[linesnumbered,ruled,vlined]{algorithm2e}
\usepackage[most]{tcolorbox}
\usepackage{pifont}
\usepackage{capt-of}
\usepackage{xcolor}
\usepackage[percent]{overpic}

\newif\ifrevisionhighlight
\revisionhighlightfalse
\definecolor{revcolor}{rgb}{0.75,0.15,0.15}
\newcommand{\rev}[1]{\ifrevisionhighlight{\color{revcolor}#1}\else#1\fi}

\newcommand{\addc}{\ifrevisionhighlight\color{revcolor}\fi}

\newcommand{\cmark}{\ding{51}}
\newcommand{\xmark}{\ding{55}}
\newcommand{\mbx}{\mathbf{x}}
\newcommand{\mbe}{\mathbf{e}}

\def\gN{{\mathcal{N}}}
\def\gX{{\mathcal{X}}}
\def\gL{{\mathcal{L}}}

\theoremstyle{plain}
\newtheorem{theorem}{Theorem}

\newcommand{\para}[1]{\noindent\textbf{#1}}

\newtheoremstyle{ibcdhead}%
  {\topsep}
  {\topsep}
  {\normalfont}
  {}
  {\bfseries}
  {.}
  {5pt plus 1pt minus 1pt}
  {\thmname{#1}\thmnumber{ #2}\thmnote{ {\normalfont\bfseries(#3)}}}
\theoremstyle{ibcdhead}
\newtheorem{proposition}[theorem]{Proposition}
\newtheorem{lemma}[theorem]{Lemma}
\newtheorem{corollary}[theorem]{Corollary}
\newtheorem{definition}[theorem]{Definition}
\newtheorem{assumption}[theorem]{Assumption}
\newtheorem{remark}[theorem]{Remark}
\newtheorem*{proprestate}{Proposition 1 (restated)}
\theoremstyle{ibcdhead}

\definecolor{wacvblue}{rgb}{0.21,0.49,0.74}
\usepackage[breaklinks,colorlinks,allcolors=wacvblue]{hyperref}
\usepackage[capitalize,noabbrev]{cleveref}

\def\confName{WACV}
\def\confYear{2027}

\title{Single-Step Bidirectional Unpaired Image Translation \\ Using Implicit Bridge Consistency Distillation}
\author{%
Suhyeon Lee$^{1}$\thanks{Equal contribution.} \qquad
Kwanyoung Kim$^{2}$\footnotemark[1] \qquad
Jong Chul Ye$^{1}$\\[2pt]
$^{1}$Kim Jaechul Graduate School of AI, KAIST \qquad $^{2}$Department of AI, GIST\\
{\tt\small \{suhyeon.lee, jong.ye\}@kaist.ac.kr \quad k0.kim@gist.ac.kr}%
}

\begin{document}
\maketitle
\begin{abstract}
Unpaired image-to-image translation has seen significant progress since the introduction of CycleGAN. However, methods based on diffusion models or Schr\"{o}dinger bridges have yet to be widely adopted in real-world applications due to their iterative sampling nature. To address this challenge, we propose Implicit Bridge Consistency Distillation (IBCD), which enables single-step bidirectional unpaired translation without using adversarial loss. IBCD reformulates consistency distillation by using a diffusion implicit bridge model that connects PF-ODE trajectories between distributions, \rev{parametrized on a signed time axis whose sign selects the target domain, so that one network translates in both directions in a single step}. Additionally, we introduce two key improvements: 1) distribution matching for consistency distillation and 2) adaptive weighting method based on distillation difficulty. Experimental results demonstrate that IBCD achieves state-of-the-art performance on benchmark datasets in a single generation step\rev{, and that it extends to brain MRI contrast translation}.
\end{abstract}

\section{Introduction}
Unpaired image-to-image (I2I) translation~\citep{zhu2017unpaired}, which transfers images between domains while preserving content without supervision, has gained continuous attention in academia and industry. This approach is particularly useful in real-world scenarios where paired data is hard to obtain, such as in medical and scientific imaging~\citep{kaji2019overview, chen2023deep}\rev{; brain MRI contrast translation is a representative case}. Despite the recent advances in modern zero-shot image editing methods~\citep{parmar2023zero, hertz2023delta, hertz2022prompt}, their applicability remains limited due to challenges such as the lack of domain-specific adaptation and the difficulty of preserving fine-grained details. Even the latest foundational models, such as GPT-Image-1, consistently underperform without domain-specific adaptation (Tab.~\ref{table:main}). Therefore, unpaired I2I translation remains essential for applications like image enhancement, artifact removal, and cross-modality translation in modern computer vision~\citep{safayani2026unpaired}.

Traditionally, CycleGAN~\citep{zhu2017unpaired} and its variants form the foundation for unpaired I2I translation~\citep{choi2018stargan, park2020contrastive, fu2019geometry, zheng2022ittr}. These methods use bidirectional generators for domain translation, along with domain-specific discriminators. Training combines adversarial loss, guided by discriminators, and cycle consistency loss from the bidirectional structure. While CycleGAN-based methods have advanced unpaired I2I translation, they still rely on adversarial loss, which can cause instability, convergence issues, and mode collapse~\citep{saad2024survey}.

The recent emergence of diffusion models (DMs) has significantly advanced unpaired I2I translation, thanks to their exceptional generative capabilities through iterative denoising. SDEdit~\citep{meng2022sdedit} performs image translation by solving the reverse process with a diffusion model trained on the target domain. This is achieved by introducing noise to the source image or mapping it to a noisy space using an inversion method~\citep{wu2023latent}. Additional regularizers balance the inherent realism-faithfulness tradeoff~\citep{zhao2022egsde, sun2023sddm}. In contrast, DDIB~\citep{su2022dual} addresses I2I translation by concatenating the ODE trajectories of two distinct DMs. However, these diffusion-based approaches are fundamentally limited by the inefficiency of numerous iterative steps.

\begin{figure}
    \centering
    \includegraphics[width=0.85\linewidth]{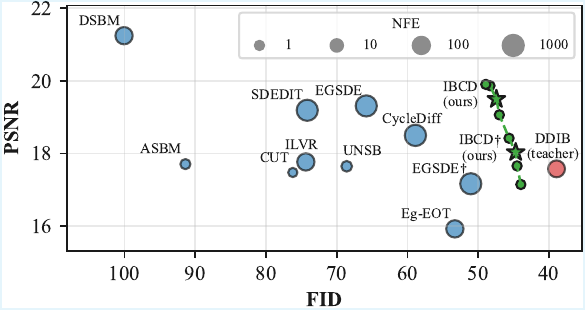}
    \vspace{-0.5em}
    \captionof{figure}{\rev{Peak signal-to-noise ratio (PSNR)} and \rev{Fr\'{e}chet inception distance (FID)} trade-off comparison with baselines on the Cat$\rightarrow$Dog task\rev{, where PSNR is measured against the \emph{source} image (Sec.~\ref{sec:psnr_caveat})}. \rev{Marker size represents the number of NFE.}}
    \vspace{-2em}
    \label{fig:intro-tradeoff}
\end{figure}

On the other hand, Schr\"{o}dinger Bridge (SB)~\citep{schrodinger1932theorie} and Optimal Transport (OT)~\citep{peyre2019computational} have emerged as promising frameworks for translating between unpaired arbitrary distributions. Recent advances have extended these concepts to deep learning, with methods such as NOT~\citep{korotin2022neural}, DSBM~\citep{shi2023diffusion}, and UNSB~\citep{kim2023unpaired} aiming to find efficient transport paths in unpaired settings. To further improve stability and guidance, adversarial training~\citep{gushchin2024adversarial}, energy-guided entropic regularizers~\citep{mokrov2024energyguided}, and displacement interpolation~\citep{choi2025improving} have been introduced. However, these approaches often remain limited by their reliance on iterative solvers or face challenges in scaling to high-resolution images.

To address the limitations, we aim at the development of a \textit{bidirectional} {\em single-step} generator that enables translation between two arbitrary distributions in \textit{unpaired} settings without relying on adversarial losses. Specifically, we propose Implicit Bridge Consistency Distillation (IBCD), a reformulation of the concept of consistency distillation (CD)~\citep{song2023consistency} that incorporates a diffusion implicit bridge model for translating between arbitrary data distributions. Unlike CD, which learns paths from Gaussian noise to data, IBCD connects trajectories from one arbitrary distribution to another one using a Probability Flow Ordinary Differential Equation (PF-ODE), allowing for flexible and efficient distribution translation in both directions.

However, simply extending CD can lead to increased distillation error due to error accumulation, as well as challenges related to model capacity and training scheme, which arise from integrating two trajectories and introducing bidirectionality. To address this, we propose a regularization method called Distribution Matching for Consistency Distillation (DMCD). Furthermore, we introduce a weighting scheme based on distillation difficulty, which applies a stronger DMCD penalty specifically to samples where the consistency loss alone proves insufficient. By integrating additional cycle translation loss with these advanced components, our approach significantly enhances the realism-faithfulness trade-off, achieving state-of-the-art performance in a single step, as shown in Fig.~\ref{fig:intro-tradeoff}. The main contributions of our work are as follows:
\begin{itemize}
\item Propose Implicit Bridge Consistency Distillation (IBCD), an unpaired image translation framework that enables bidirectional translation with a single NFE and achieves state-of-the-art results.
\item Introduce regularizing components, including Distribution Matching for Consistency Distillation (DMCD), an adaptive weighting scheme based on distillation difficulty \rev{(grounded by a bound we prove rather than chosen empirically)}, and cycle translation loss to mitigate inherent distillation errors.
\item Demonstrate the effectiveness of IBCD through extensive experiments on toy data, natural images, and \rev{brain MRI contrast translation}, covering a range of diverse modalities.
\end{itemize}

\vspace{0.6em}

\begin{figure*}[ht!]
    \centering
    \includegraphics[width=0.8\textwidth]{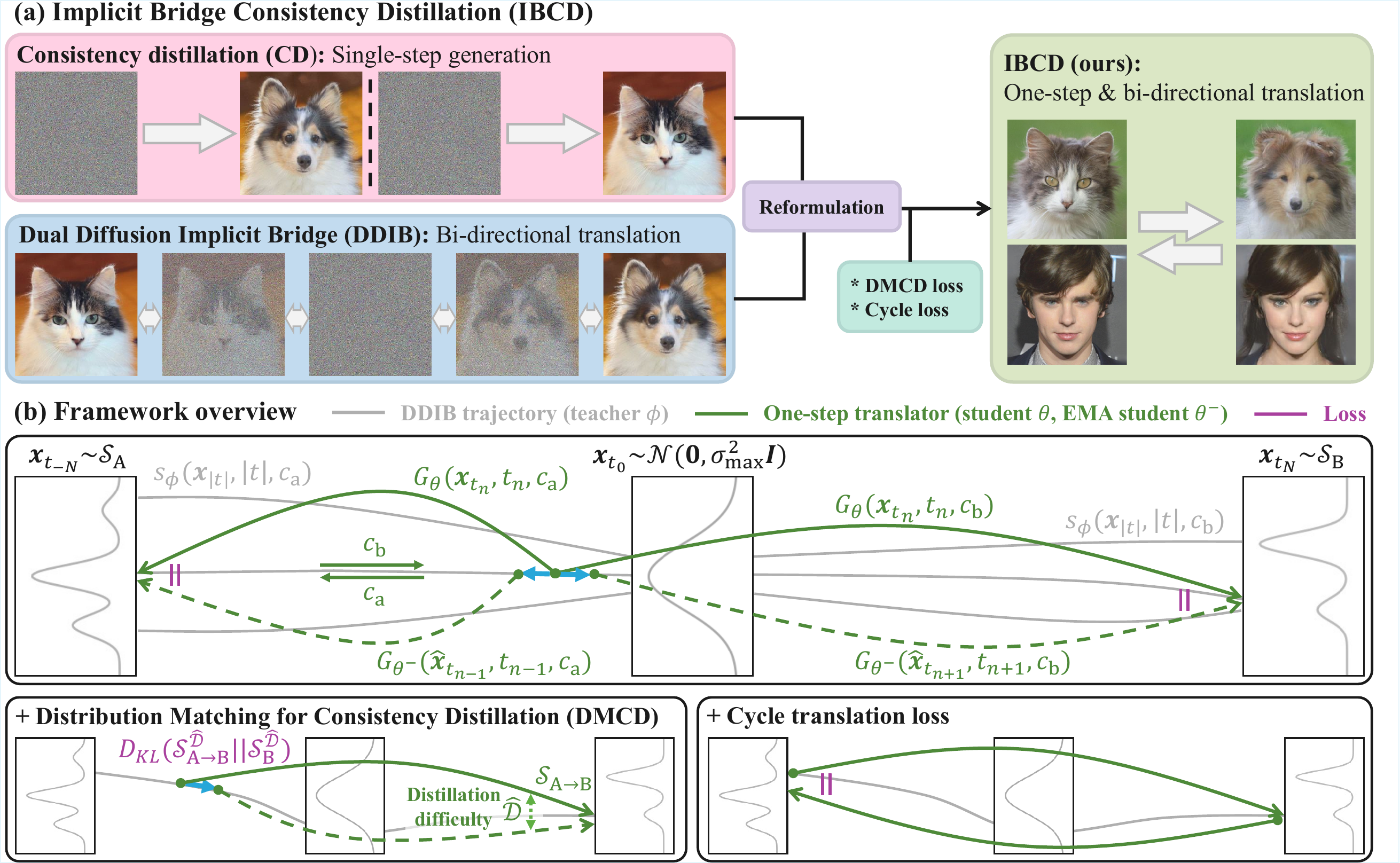}
    \vspace{-0.25em}
    \caption{
    (a) IBCD performs single-step bi-directional translation using a distillation framework that reformulates consistency distillation with a diffusion implicit bridge and introduces regularizers. (b) The IBCD framework bridges two distributions by connecting the PF-ODE paths of two pre-trained diffusion models through bidirectionally extended consistency distillation. To mitigate distillation errors, we introduce distribution matching for consistency distillation and a cycle translation loss.
    }
    \label{fig:method_overview}
    \vspace{-1em}
\end{figure*}

\section{Preliminaries}

\subsection{Image Translation with Diffusion Models}
\para{Diffusion Models (DM).} In DMs~\citep{ho2020denoising, song2021scorebased}, the predefined forward process with the time variable \( t \in [0, T] \) progressively corrupts data into pure Gaussian noise over a series of steps \( T \). Specifically, given a data distribution \( \mbx_0 \sim p(\mbx_0) := p_{\text{real}}(\mathbf{x}) \), the distribution \( \mbx_T \sim p(\mathbf{x}_T) \) approaches an isotropic normal distribution as noise is added according to the process \( p(\mathbf{x}_t \mid \mathbf{x}_0) = \mathcal{N}(\mathbf{x}_0, t^2 \mathbf{I}) \). The reverse process can be described by \rev{a PF-ODE}~\citep{song2021scorebased}:
\begin{align}
    \frac{d \mathbf{x}_t}{dt} = -t \nabla_{\mathbf{x}_t} \log p(\mathbf{x}_t) = \frac{\mathbf{x}_t - \mathbb{E}[\mathbf{x}_0|\mathbf{x}_t]}{t}, \label{eq:pf-ode}
\end{align}
where the second equality follows from Tweedie's formula, \( \mathbb{E}[\mathbf{x}_0|\mathbf{x}_t] = \mathbf{x}_t + t^2 \nabla_{\mathbf{x}_t} \log p(\mathbf{x}_t) \) \citep{efron2011tweedie,kim2021noise2score}. In practice, the neural network is trained to approximate the ground truth score function \( \mathbf{s}_{\phi}(\mbx_t, t) \approx \nabla_{\mathbf{x}_t} \log p(\mathbf{x}_t) \) or the denoiser $D_{\phi}(\mbx_t,t) \approx \mathbb{E}[\mbx_0 |\mbx_t]$ by denoising score matching~\citep{vincent2011connection}. Substituting the trained networks into Eq. (\ref{eq:pf-ode}), we obtain the denoised sample by numerical integration from \( T \) to \( 0 \):
\begin{equation}
\mbx_0     = \mbx_T + \int^{0}_{T} -t\cdot \textbf{s}_{\phi}(\mbx, t) \, dt  
= \mbx_T +  \int^{0}_{T} \frac{\mathbf{x}_t -D_{\phi}(\mbx_t,t)}{t} \, dt.
\label{eq:ep-ode}
\end{equation}
To solve Eq.~(\ref{eq:ep-ode}),  an ODE solver, denoted as \(\texttt{Solve}(\mbx_T;\phi, T,0)\) (with an initial state $\mbx_T$ at time $T$ and ending at time 0, DM parameterized by  $\phi$) can be applied. Examples include the Euler solver~\citep{song2021scorebased, ho2020denoising}, DPM-solver~\citep{lu2022dpm}, or the second-order Heun solver~\citep{karras2022elucidating}. The sampling process typically requires dozens to hundreds of neural function evaluations (NFE) to effectively minimize discretization error during ODE solving.

\para{Dual Diffusion Implicit Bridge (DDIB).} DDIB~\citep{su2022dual} is a simple yet effective method for I2I translation that leverages the connection between DMs and Schr\"{o}dinger bridge problem (SBPs), where DMs act as implicit optimal transport models. DDIB requires training two individual DMs for the two domains $\gX_\text{A}$ and $\gX_\text{B}$, denoted as $\mathbf{s}_{\phi_\text{A}}$ and $\mathbf{s}_{\phi_\text{B}}$, where the subscript names the domain a model is trained on. The sampling process solves the two ODEs in sequence:
\begin{equation}
    \mbx^\text{l} = \texttt{Solve}(\mbx^\text{A}; \phi_\text{A}, 0, T), \;  \mbx^\text{B} = \texttt{Solve}(\mbx^\text{l}; \phi_\text{B}, T, 0). \label{ddib}
\end{equation}
Here, $\mbx^\text{l}$ represents the latent code in the pure Gaussian noise domain, $\mbx^\text{A} \in \gX_\text{A}$ is the image in the source domain, and $\mbx^\text{B}$ is the estimated image in the target domain $\gX_\text{B}$. Thanks to the intermediate Gaussian distribution, DDIB automatically satisfies the cycle consistency property without any explicit regularization term~\citep{zhu2017unpaired, choi2018stargan}.

\subsection{Existing Single-Step Acceleration Approaches}
\para{Consistency Distillation (CD).} The aim of the consistency distillation (CD)~\citep{song2023consistency} is to learn the direct mapping from noise to clean data. Specifically, the model is designed to predict \( G_\theta(\mathbf{x}_t, t) = \mathbf{x}_0 \), and is constrained to be $\textit{self-consistent}$, meaning that outputs should be the same for any time point input within the same PF-ODE trajectory, $i.e.,$ $ G (\mathbf{x}_t, t) = G(\mathbf{x}_{t'}, t')$ for all $t, t' \in [\epsilon, T]$, with the boundary condition $G_\theta(\mathbf{x}_\epsilon, \epsilon)=\mathbf{x}_\epsilon$. Here, $\epsilon$ is a small positive number, to avoid numerical instability at $t=0$. By discretizing the time interval $[\epsilon, T]$ into $N-1$ sub-interval with boundaries $t_1 = \epsilon < t_2 < \cdots < t_N = T$, the resulting objective function for CD is given by:
\begin{equation}
\mathcal{L}_{\text{CD}}(\theta;\phi) = \mathbb{E} [\lambda(t_n)d(G_{\theta}(\mathbf{x}_{t_{n+1}}, t_{n+1}), G_{\theta^{-}}(\mathbf{\hat{x}}_{t_n}, t_n))],
\end{equation}
where $n \sim \mathcal{U}[1,N-1]$ and $\lambda(t_n)$ is weight hyperparameter, \( d(\cdot, \cdot) \) measures the distance between two samples. \( \theta^{-} \) is the exponential moving average (EMA) of the student parameter \( \theta \), and \( \phi \) represents the pre-trained teacher model, and $\mathcal{U}[\cdot]$ refers to the uniform distribution. The target \(\mathbf{\hat{x}}_{t_n}\) is obtained using a one-step ODE solver, \textit{i.e.}, \(\mathbf{\hat{x}}_{t_n} = \texttt{Solve}(\mbx_{t_{n+1}}; {\phi},t_{n+1}, t_n) \), from $\mbx_{t_{n+1}} \sim \mathcal{N}(\mbx_0, t_{n+1}^2 \mathbf{I})$. 

\para{Distribution Matching Distillation (DMD).}  DMD~\citep{yin2024one, wang2023prolificdreamer} minimizes the Kullback-Leibler (KL) divergence between the real data distribution, $p^{\text{real}}$, and the student sample distribution, $p_{\theta}^{\text{fake}}$ to distill the diffusion model $\mathbf{s}_{\phi}^{\text{real}}$ into a single-step generator $G_{\theta}(\mathbf{x}_T) = \mathbf{x}_0$.
Additionally, DMD introduces an auxiliary ``fake'' DM, $s_{\psi}^{\text{fake}}$, to approximate the score function of the student-generated sample distribution. This estimator is trained with denoising score matching, adapting in real-time as the student model progresses through training. The gradient of the DMD loss is then approximated as the difference between the two score functions:
\begin{multline}
\nabla_{\theta}D_{\text{KL}} (p_{\theta}^{\text{fake}}||p^{\text{real}}) \approx \nabla_\theta \mathcal{L}_{\text{DMD}} \\ = \underset{\mbx_t, t, \mbx_T} {\mathbb{E}} [w_t (s_{\psi}^{\text{fake}}(\mbx_t, t) - s_{\phi}^{\text{real}}(\mbx_t, t)) \nabla_{\theta}G_{\theta}(\mbx_T)], \label{eq:dmd}
\end{multline}
where $\mbx_t \sim \mathcal{N}(G_{\theta}(\mbx_T), t^2 \mathbf{I})$, $t\sim\mathcal{U}(T_{\text{min}}, T_{\text{max}})$, $\mbx_T \sim \mathcal{N}(\mathbf{0},T^2 \mathbf{I})$ and $w_t$ is a scalar weighting factor. DMD serves as an effective distillation loss that optimizes the student model from the view of the distribution, without relying on unstable adversarial loss~\citep{goodfellow2014generative}.

\section{Method}
\label{sec:method}

We aim to develop a one-step distillation method for bidirectional mapping between arbitrary distributions in an unpaired setting, using pretrained diffusion models. Specifically, given two domains $\gX_\text{A}$ and $\gX_\text{B}$ with unpaired datasets $\mathcal{S}_\text{A}$ and $\mathcal{S}_\text{B}$, our translator $G_{\theta}$ performs two translations: $G_{\theta}(\mbx^\text{A}, c_\text{B})$ for A$\rightarrow$B and $G_{\theta}(\mbx^\text{B}, c_\text{A})$ for B$\rightarrow$A, where \rev{$c_\text{A}$ and $c_\text{B}$ are \emph{domain embeddings}: one learned vector per domain, taken from the class-conditioning embedding table of the teacher and shared by every image of that domain, so they carry no per-example information}. The concept of our method is illustrated in Fig.~\ref{fig:method_overview} by contrasting it to CD and DDIB. In the following, we describe a distillation approach with distribution matching, adaptive weighting, and a cycle loss for bidirectional reconstruction.

\subsection{Implicit Bridge Consistency Distillation} \label{sec:ibcd}

\para{\rev{Setup and notation.}} \rev{The teacher is a \emph{single} class-conditional EDM~\citep{karras2022elucidating} diffusion model $\mathbf{s}_{\phi}(\mbx_t, t, c)$, trained jointly on both datasets by denoising score matching under EDM's log-normal noise schedule, without modification. We write $c \in \{c_\text{A}, c_\text{B}\}$ for the requested \emph{target} domain and $c'$ for the opposite one. Conditioning is thus an argument, not a parameter set: DDIB's two per-domain teachers are this single model evaluated at $c_\text{A}$ and at $c_\text{B}$, and the student $G_\theta(\mbx_t,t,c)$ likewise has one $\theta$ for both directions: $G_{\theta}(\mbx^\text{A}, \cdot, c_\text{B})$ gives A$\rightarrow$B and $G_{\theta}(\mbx^\text{B}, \cdot, c_\text{A})$ gives B$\rightarrow$A.}

\para{\rev{Signed-time discretization.}} \rev{Let $\sigma_0 = \sigma_{\text{max}} > \sigma_1 > \cdots > \sigma_{N-1} = \sigma_{\text{min}}$ be the $N$ noise levels of the EDM sampling schedule ($\sigma_{\text{min}}=0.002$, $\sigma_{\text{max}}=80$, $N=40$, $\rho=7$), and let $\sigma_N := 0$, so that $\sigma_0$ is the Gaussian latent shared by the two PF-ODEs and $\sigma_N$ is clean data.} Because DDIB concatenates two ODEs, an unsigned time would name each noise level twice. \rev{We therefore index the concatenated trajectory by a signed integer $i$ and define the student's timestep as}
{\addc
\begin{gather}
    i \in [\underbrace{-N, \cdots, -1}_{\gX_\text{A}}, \underbrace{0}_{\gX_\text{A} {\cap} \gX_\text{B}}, \underbrace{1, \cdots, N}_{\gX_\text{B}}], \nonumber \\
    t_i = \text{sign}(i) \, \sigma_{|i|}, \qquad \text{sign}(0) := +1.  \label{eq:t_i}
\end{gather}}
\rev{The sign of $t$ says which domain's diffusion a point belongs to and $|t|$ is its noise level: $i=0$ is the shared latent, and the two ends $t_{-N}$ and $t_{N}$ are the clean domains, both of magnitude $0$ (written $\text{-}0$ and $\text{+}0$ to record the side).}

\para{Boundary condition.} The student is required to be \textit{self-consistent} along Eq.~(\ref{eq:t_i}), \rev{\ie its output must agree for every input taken from one trajectory.} Unlike in CD, the boundary to which the student collapses depends on the target domain $c$:
\begin{multline}
     G(\mbx_{\epsilon(c)}, \epsilon(c), c)  =
        \mbx_{\epsilon(c)},
    \\ \text{where} \;
\epsilon(c) = \begin{cases}
        t_{-N+1} &= -\sigma_{\text{min}}, \quad \text{for} \; c = c_\text{A} \\
        t_{N-1} &= +\sigma_{\text{min}}, \quad \text{for} \; c = c_\text{B}
    \end{cases}. \label{eq:boundary}
\end{multline}
\rev{As in CD, the offset $\sigma_{\text{min}}$ avoids the numerical instability at zero noise.} Together with the IBCD loss below, this yields translation by injecting the desired domain condition, $G(\mbx_{\epsilon(c)}, \epsilon(c), c') = \mbx_{\epsilon(c')}$: from any point $\mbx_t$ of a trajectory, $G(\cdot,\cdot,c)$ returns the clean image that this trajectory reaches in domain $c$. Since EDM/CD is not defined for negative $t$ and does not meet this boundary condition, we reparametrize the student. \rev{Its preconditioning coefficients $c_\text{skip}$, $c_\text{out}$, $c_\text{in}$ become functions of $(t,c)$, extended symmetrically about $t=0$ with $\text{sign}(t)$ selecting the active branch, so that $c_\text{skip}=1$ and $c_\text{out}=0$ at $t=\epsilon(c)$ for either domain, and the timestep rescaler is made odd in $t$. Eq.~(\ref{eq:boundary}) then holds by construction on both sides of the axis, while the positive branch reduces to EDM/CD; App.~\ref{supp:extend-edm} gives the coefficients.}\footnote{The reparametrization applies to the student only; the teacher keeps the original EDM form.}

\para{\rev{Distillation targets and loss.}} As in CD, a training pair lies on one teacher trajectory and is obtained by a single solver step. \rev{We write $\texttt{Solve}(\mbx;\phi,c,|t|,|t'|)$ for one step of the PF-ODE of the teacher conditioned on domain $c$, from noise level $|t|$ to $|t'|$; only magnitudes enter, as a step never crosses the latent.} For the direction $\gX_\text{A} \rightarrow \gX_\text{B}$ (\ie $c=c_\text{B}$), the pairs used for distillation are
{\addc
\begin{align}
\hat \mbx_{t_{{i+1}}} = \texttt{Solve}(\mbx_{t_{{i}}};{\phi},c_\text{A},|t_{{i}}|, |t_{{i+1}}|), \nonumber \\ \hat \mbx_{t_{{j+1}}} = \texttt{Solve}(\mbx_{t_{{j}}};{\phi},c_\text{B},|t_{{j}}|,|t_{{j+1}}|), \label{eq:forward}
\end{align}}
where ${i} \sim \mathcal{U}[-N+1, -1]$, ${j} \sim \mathcal{U}[0, N-2]$, $\mbx_{t_{{i}}} \sim \gN(\mbx^\text{A},t^2_{{i}} \mathbf{I})$, $\mbx_{t_{{j}}} \sim \gN(\mbx^\text{B},t^2_{{j}} \mathbf{I})$\rev{; the two lines move the domain-A and the domain-B half of the minibatch one step towards $\gX_\text{B}$}. Symmetrically, for $\gX_\text{B} \rightarrow \gX_\text{A}$ (\ie $c=c_\text{A}$),
{\addc
\begin{align}
\hat \mbx_{t_{{i-1}}} = \texttt{Solve}(\mbx_{t_{{i}}};{\phi},c_\text{A},|t_{{i}}|,|t_{{i-1}}|), \nonumber \\
\hat \mbx_{t_{{j-1}}} = \texttt{Solve}(\mbx_{t_{{j}}};{\phi},c_\text{B},|t_{{j}}|,|t_{{j-1}}|), \label{eq:backward}
\end{align}}
with ${i} \sim \mathcal{U}[-N+2, 0]$, ${j} \sim \mathcal{U}[1, N-1]$. Given these targets, the IBCD objective is
\begin{equation}
\mathcal{L}_{\text{IBCD}}(\theta;\phi) =
\underset{\mathbf{t}_1, \mbx_{\mathbf{t}_1},c}{\mathbb{E}} [\lambda(\mathbf{t}_2) \, d(G_{\theta}(\mathbf{x}_{\mathbf{t}_1}, \mathbf{t}_1, c), G_{\theta^{-}}(\hat{\mbx}_{\mathbf{t}_2}, \mathbf{t}_2, c))]. \label{eq:ibcd}
\end{equation}
Here the target domain is drawn as $c \sim \mathcal{U}\{c_\text{A}, c_\text{B}\}$, and $\mathbf{t}_1 = [t_{{i}}; t_{{j}}]$, $\mathbf{t}_2 = [t_{{i}\pm1}; t_{{j}\pm1}]$, $\mbx_{\mathbf{t}_1} = [\mbx_{t_{{i}}}; \mbx_{t_{{j}}}]$ and $\hat \mbx_{\mathbf{t}_2} = [\hat \mbx_{t_{{i}\pm1}}; \hat \mbx_{t_{{j}\pm1}}]$ \rev{collect the inputs and the one-solver-step targets of Eqs.~(\ref{eq:forward}) and~(\ref{eq:backward}), where $[\,\cdot\,;\,\cdot\,]$ stacks the two halves of the minibatch and the sign is $+$ for $c=c_\text{B}$ and $-$ for $c=c_\text{A}$; $\lambda$ is the CD weighting, $d$ a distance, and $\theta^- = \texttt{sg}(\mu \theta^- + (1-\mu)\theta)$ the stop-gradient EMA of $\theta$ with rate $\mu$.} See Algo.~\ref{alg:ibcd-vanilla} in App.~\ref{supp:algorithm}.

\rev{Sharing one conditional teacher across both domains halves the teacher memory and, since the student is initialized from it and inherits its embeddings, starts the student from a map that already sends $c$ to a clean image of domain $c$.} Methods in the literature~\citep{kim2023consistency, li2024bidirectional} extend CD in both directions or specify a target timestep, but do not integrate the domain condition into the framework itself.

\subsection{Objective Function}
\label{sec:ibcd2}

Vanilla IBCD lays a foundation for one-step bidirectional transport, but three issues remain. First, the consistency loss is a \textit{local} strategy (categorized by \citet{kim2023consistency}): it aligns only adjacent timesteps, so local errors accumulate into a growing discrepancy between $G_\theta(\mbx_t,t,c)$ and the true boundary $\mbx_{\epsilon(c)}$, an effect amplified by IBCD's doubled trajectory. Second, one student must fit two ODE trajectories whose targets differ by direction, which strains its capacity, as also observed in CD by \citet{li2024bidirectional}. Third, unlike EGSDE~\citep{zhao2022egsde}, which balances realism and fidelity by expert weighting, vanilla IBCD offers no such control.

\para{Distribution Matching for Consistency Distillation.}
To address these issues, we propose Distribution Matching for Consistency Distillation (DMCD), which brings the DMD loss into the CD framework: it minimizes the KL divergence between the student's output distribution and the target domain distribution over all timesteps of both directions, weighted by a distillation difficulty factor $\Hat{\mathcal{D}}(\cdot, \cdot)$ that focuses training on hard samples. \rev{Its gradient factorizes into four terms, each labelled in Fig.~\ref{fig:method_overview}(b):}
{\addc
\begin{align}
    \nabla_\theta & \mathcal{L}_{\text{DMCD}}(\theta; \phi, \psi) = \underset{\mathbf{t}_1, \mbx_{\mathbf{t}_1}, c, i, \mbx_{t_i}}{\mathbb{E}} \big[ \;
    \underbrace{w_{t_i}}_{\text{(i)}} \;
    \underbrace{\Hat{\mathcal{D}}(\texttt{sg}(\mbx_{\mathbf{t}_1}), c)}_{\text{(ii)}} \nonumber \\
    & \cdot \underbrace{(s_{\psi}(\mbx_{t_i},t_i,c)  - s_{\phi}(\mbx_{t_i}, t_i, c))}_{\text{(iii)}} \;
    \underbrace{\nabla_{\theta}G_{\theta}(\mbx_{\mathbf{t}_1},\mathbf{t}_1,c)}_{\text{(iv)}} \big], \label{eq:dmcd} \\
    & \text{where} \quad i \sim \mathcal{U}[0, N-1], \; \mbx_{t_i} \sim \gN(G_{\theta}(\mbx_{\mathbf{t}_{1}}, \mathbf{t}_1, c),t_i^2 \mathbf{I}), \nonumber
\end{align}}
and $\mathbf{t}_1, \mbx_{\mathbf{t}_1}, c$ are as in Eq.~(\ref{eq:ibcd}). \rev{Term (iii) is the distribution-matching direction: the gap between the score $s_{\psi}(\cdot,\cdot,c)$ of the student's output distribution and the score $s_{\phi}(\cdot,\cdot,c)$ of the real data of domain $c$, \ie the DMD estimate of the gradient of $D_{\text{KL}}(\mathcal{S}_{\text{A}\rightarrow\text{B}} \| \mathcal{S}_\text{B})$ marked in the DMCD block of Fig.~\ref{fig:method_overview}(b); $s_{\psi}$ is an auxiliary class-conditional ``fake'' diffusion model trained online by DSM on the student's outputs, and $s_{\phi}$ the frozen teacher. Term (iv) is the Jacobian of the one-step translator (the student block of the same figure), taken at the same input $\mbx_{\mathbf{t}_1}$ as the IBCD loss; $\mbx_{t_i}$ is its output re-noised to $t_i$. Terms (i) and (ii) are DMD's time weighting~\citep{yin2024one} and the distillation difficulty weight $\Hat{\mathcal{D}}$, defined next.}

Unlike DMD, DMCD acts as a regularizer rather than the primary objective. This matters in unpaired settings, where the DMD loss alone does not tie the domains together; recent work~\citep{rakitin2024regularized} therefore adds an explicit $L_2$ term between source and target. In IBCD the consistency loss already builds the trajectory that enforces the correspondence, so DMCD is left to supply distribution matching, and hence realism, without adversarial instability~\citep{zhu2017unpaired, parmar2024one, kim2023unpaired}.

\para{Distillation Difficulty Adaptive Weighting.}
\rev{Applied at the same strength everywhere, DMCD works against the objective it supports: its gradient pulls the student's output towards the target distribution and away from the teacher's estimate, at a cost in faithfulness. The pull is worth its price only where the consistency loss alone translates poorly, mostly on trajectories starting near a decision boundary of the source domain (App.~\ref{supp:meanpred}); elsewhere it buys realism that was not missing. To make DMCD act only where it is needed, we define the \textit{distillation difficulty} $\mathcal{D}([\mbx_{t_{-N+1}}, \cdots, \mbx_{t_{N-1}}], c):=d(G_\theta(\mbx_{\epsilon(c^\prime)}, \epsilon(c^\prime), c), \mbx_{\epsilon(c)})$, the end-to-end error the student makes on that trajectory, and weight DMCD by it.}

Evaluating $\mathcal{D}$ exactly is impractical: recovering $\mbx_{\epsilon(c)}$ and $\mbx_{\epsilon(c^\prime)}$ from a given $\mbx_{t}$ costs at least $N$ teacher NFEs per DMCD evaluation, so we use a one-step approximation,
\begin{equation}
     \Hat{\mathcal{D}}(\mbx_{\mathbf{t}_1}, c) =  g(d(G_{\theta}(\mathbf{x}_{\mathbf{t}_1}, {\mathbf{t}_1}, c), \; G_{\theta^{-}}(\Hat{\mbx}_{{\mathbf{t}_2}}, {\mathbf{t}_2}, c))) \label{eq:ddhat}
\end{equation}
where $\mathbf{t}_1, \mathbf{t}_2, \mbx_{\mathbf{t}_1}, \Hat{\mbx}_{\mathbf{t}_2}$ are those of Eqs.~(\ref{eq:ibcd}),~(\ref{eq:dmcd}) and $g$ is any monotone increasing function. \rev{Its argument is the IBCD residual, already computed in Eq.~(\ref{eq:ibcd}), so the weight costs no extra NFE.}
\rev{To justify this substitution, we prove a bound on the quantity it replaces:}

{\addc
\begin{proposition}[Residual bound on distillation difficulty]
\label{prop:dd_main}
Under the assumptions of App.~\ref{supp:ddtheory} (Lipschitz student, order-$p$
solver, EMA gap bounded by $\Delta_{\text{ema}}$, metric $d$), every exact bridge trajectory
$\mathbf{x}$ and target domain $c$ satisfy
\begin{equation*}
\mathcal{D}(\mathbf{x},c) \leq \textstyle\sum_{n} \delta_n(\mathbf{x},c) + (2N{-}2)\,\Delta_{\text{ema}} + \mathcal{O}((\Delta t)^{p}),
\end{equation*}
where $\delta_n$ is the argument of $g$ at step $n$, summed over the $2N{-}2$
steps of the bridge, $\Delta t$ is the largest noise-level step of the schedule,
and the floor is twice CD's.
\end{proposition}}

\rev{Here $\delta_n$ is the one-step consistency residual at step $n$: the quantity the IBCD
loss already computes, and the argument of $g$ in Eq.~(\ref{eq:ddhat}). The bound therefore
controls the end-to-end difficulty $\mathcal{D}$, which costs $N$ teacher evaluations to
measure, by residuals that are already available for free. A small residual means the
trajectory is already distilled, so weighting DMCD by $\Hat{\mathcal{D}}$ spends it only where
that is not the case; App.~\ref{supp:ddtheory} gives the proof.}

\para{Cycle Consistency Loss.}
Like DDIB, our framework performs cycle translation and must therefore satisfy cycle consistency, which we enforce with $\gL_{\text{cycle}}(\theta)$:
\begin{align}
\underset{c, \mbx_{\epsilon(c)}}{\mathbb{E}}[d(G_{\theta}(G_{\theta}(\mbx_{\epsilon(c)},\epsilon(c),c^\prime),\epsilon(c^\prime),c), \mbx_{\epsilon(c)})].
\end{align}

\para{Final Loss Functions.}
The final loss, weighted by $\lambda_{\text{DMCD}}$, $\lambda_{\text{cycle}}$, for training $G_{\theta}$ is given by:
\begin{align}
\theta^* = \arg \underset{\theta}{\min} \, \gL_{\text{IBCD}} + \lambda_{\text{DMCD}} \gL_{\text{DMCD}}
+\lambda_{\text{cycle}} \gL_{\text{cycle}}.
\end{align}
Training proceeds in two stages: it starts with the IBCD loss alone, and the DMCD and cycle losses are switched on once the student has approximately converged. \rev{The switch is not a free-running heuristic: each model's transition step was fixed in advance, where the student's FID stopped improving. App.~\ref{supp:ddtheory_switch} states the criterion this instantiates, and App.~\ref{supp:trainingcost} the per-model step and the resulting training cost.} The final algorithm is given in Algo.~\ref{alg:ibcd-final}, App.~\ref{supp:algorithm}.

\section{Experiments}

\subsection{Toy Data Experiment}

To evaluate the effectiveness of our framework in a controlled setting, we conducted bidirectional translation experiments on a two-dimensional synthetic toy dataset, where the domains $A$ and $B$ were represented by the S-curve and Swiss roll distributions, respectively.

\begin{figure*}[]
    \centering
    \includegraphics[width=0.60\textwidth]{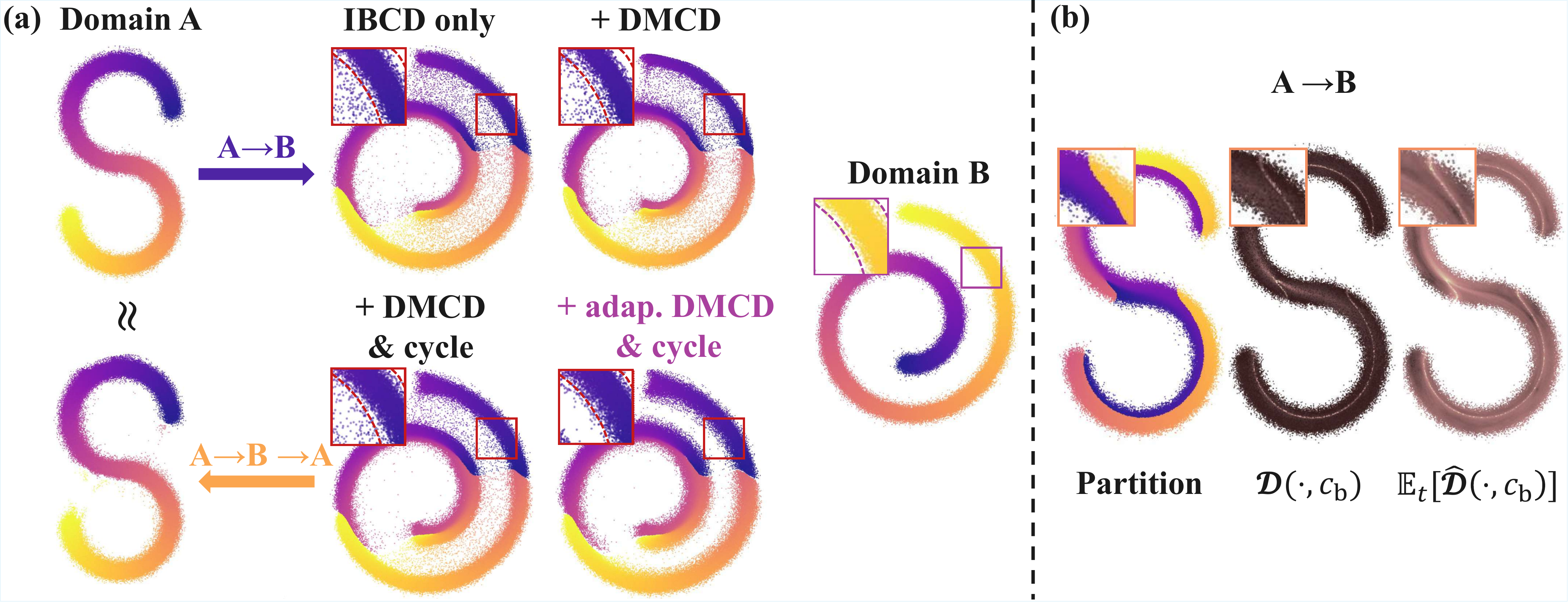}
    \caption{(a) Bidirectional translation results on a toy dataset, showing the contributions of each component. \rev{The boxes mark the inset region, the same portion of the outer branch of the target Swiss roll in every panel; the dotted curve is that branch's inner boundary, traced on the ground-truth Domain~B samples.} (b) Visualization of distillation difficulty $\mathcal{D}(\cdot, c_\text{B})$ and its one-step approximation $\mathbb{E}_t[\Hat{\mathcal{D}}(\cdot, c_\text{B})]$ for A$\rightarrow$B translation, with $g$ selected as a logarithm.}
    \label{fig:result_toy}
    \vspace{-0.5em}
\end{figure*}

\para{Validity of IBCD.} Fig.~\ref{fig:result_toy}(a) shows A$\rightarrow$B translation for the cumulative components. The IBCD loss alone translates, but maps some points, mostly originating at the source-domain decision boundary, into low-density regions of the target (Appendix~\ref{supp:meanpred}). DMCD drives more points into high-density regions but does not recover those already in low-density areas and reduces mode coverage; the cycle loss restores that coverage and sharpens the decision boundary; adaptive weighting then selectively corrects the drifted points. The complete cycle A$\rightarrow$B$\rightarrow$A of the final model recovers the source, demonstrating cycle consistency.

\para{Distillation Difficulty.} Fig.~\ref{fig:result_toy}(b) illustrates its role. The left panel shows the source-domain decision boundary induced by the DDIB teacher's B$\rightarrow$A translation; the middle and right panels show $\mathcal{D}([\mbx_{t_{-N+1}}, \cdots, \mbx_{t_{N-1}}], c_\text{B})$ and its expected one-step approximation $\mathbb{E}_{t\sim\mathcal{U}[-N+1, N-2]} [\Hat{\mathcal{D}}(\mbx_t, c_\text{B})]$ for A$\rightarrow$B, plotted at the source location $\mbx_{\epsilon(c_\text{A})}$. The measure tracks the decision boundary, marking the regions that are hard for the student, and its one-step approximation reproduces the same structure, which is what makes it usable as a training-time weight.

\begin{table*}[!t]
\centering
\scriptsize
\caption{\textbf{Quantitative comparison of unpaired image-to-image translation tasks}. Most results are from the EGSDE paper, except those marked with *, which are obtained by training the model on our dataset using the default configuration, and the Density-Coverage metric~\citep{naeem2020reliable}. Marker $\dagger$ marks \rev{the same network at} a hyperparameter configuration that prioritizes realism over faithfulness. \rev{PSNR/SSIM are \emph{source}-referenced and measure similarity to the source, not fidelity.}}
\label{table:main}
\renewcommand{\arraystretch}{0.88}
\setlength{\tabcolsep}{25pt}
\resizebox{0.98\textwidth}{!}{
\begin{small}
\begin{tabular}{lllllll}
\toprule
\textbf{Method} & NFE $\downarrow$ & FID $\downarrow$ & PSNR $\uparrow$ & SSIM $\uparrow$ & {Density $\uparrow$} & {Coverage $\uparrow$}  \\
\midrule
\multicolumn{7}{c}{\textbf{Cat$\rightarrow$Dog}}\\
\midrule
StarGAN v2~\citep{choi2020stargan} & 1 & 54.88 & 10.63  & 0.270 & - & - \\
CUT~\citep{park2020contrastive} & 1 & 76.21 &  17.48 & 0.601 & {0.971} & {0.696}\\
\cmidrule(lr){1-7}
NOT$^*$~\citep{korotin2022neural} & 1 & 161.54 & 15.12 & 0.566 & 0.531 & 0.072 \\
DSBM$^*$~\citep{shi2023diffusion} & 100 & 100.08 & \textbf{21.24} & 0.532 & 0.750 & 0.396 \\
UNSB$^*$~\citep{kim2023unpaired} & 5 & 68.59 & 17.65 & 0.587 & {1.045} & {0.706} \\
ASBM$^*$~\citep{gushchin2024adversarial} & 4 & 91.40 & 17.71 & 0.463 & 0.871 & 0.478 \\
Eg-EOT$^*$~\citep{mokrov2024energyguided} & 100 & 53.29 & 15.93 & 0.349 & 1.085 & 0.626 \\
DIOTM$^*$~\citep{choi2025improving} & 1 & 75.70 & 12.03 & 0.363 & 1.215 & 0.590 \\
\midrule
\cmidrule(lr){1-7}
ILVR~\citep{choi2021ilvr} & {1000} & 74.37 & 17.77 & 0.363 & {1.019} & {0.566} \\
SDEdit~\citep{meng2022sdedit} & 1000 & 74.17 & 19.19 & 0.423 & {0.997} & {0.526}\\
EGSDE~\citep{zhao2022egsde} & 1000 & 65.82 & 19.31 & 0.415 & {1.258} & {0.634} \\
EGSDE$\dagger$~\citep{zhao2022egsde} & 1200 & 51.04 & 17.17 & 0.361 & {1.509} & {0.823} \\
CycleDiffusion~\citep{wu2023latent} & 1000(+100) & 58.63 & 18.36 & 0.537 & {0.905} & {0.767} \\
SDDM~\citep{sun2023sddm} & 100 & 62.29 & - & 0.422 & - & - \\
SDDM$\dagger$~\citep{sun2023sddm} & 120 & 49.43 & - &  0.361 & - & - \\
\midrule
GPT-Image-1~(Foundation)~\citep{openai_image_generation_api_2025} & $\gg$1 & {\color{gray} 77.81} & {\color{gray} 12.18} &  {\color{gray} 0.283} & {\color{gray} 0.947} &  {\color{gray} 0.586} \\
DDIB$^*$ (Teacher)~\citep{su2022dual} & {160} & {\color{gray} 38.91} & {\color{gray} 17.58} &  {\color{gray} 0.588} & {\color{gray} 1.528} &  {\color{gray} 0.934} \\
\textbf{IBCD (Ours)} & 1 & \textbf{47.44}  & \underline{19.50}  & \textbf{0.701} & {1.412} & {\textbf{0.940}}\\
\textbf{IBCD$\dagger$ (Ours)} & 1 & \textbf{44.77}  & 18.04 & \textbf{0.663} & {\textbf{1.542}} & {\textbf{0.935}} \\
\midrule
\multicolumn{7}{c}{\textbf{Wild$\rightarrow$Dog}}\\
\midrule

CUT~\citep{park2020contrastive} & 1 & 92.94 & 17.20 & 0.592 & - & - \\
UNSB$^*$~\citep{kim2023unpaired} & 5  & 70.03 & 16.86 &  0.573 & {1.035} & {0.704} \\
\cmidrule(lr){1-7}
ILVR~\citep{choi2021ilvr} & {1000} & 75.33 &  16.85 & 0.287 & {1.275} & {0.531} \\
SDEdit~\citep{meng2022sdedit} & 1000 & 68.51 &  17.98 & 0.343 & {1.292} & {0.636} \\
EGSDE~\citep{zhao2022egsde} & 1000 & 59.75 &  18.14 & 0.343 & {1.482} & {0.683} \\
EGSDE$\dagger$~\citep{zhao2022egsde} & 1200 & 50.43 & 16.40 & 0.300 & {\textbf{1.733}} & {0.782} \\
CycleDiffusion~\citep{wu2023latent} & 1000(+100) & 58.92 & 17.68 & 0.458 & {1.014} & {0.801} \\
SDDM~\citep{sun2023sddm} & 100 & 57.38 & - &  0.328 & - & - \\
\midrule
GPT-Image-1~(Foundation)~\citep{openai_image_generation_api_2025} & $\gg$1 & {\color{gray} 100.72} & {\color{gray} 12.38} &  {\color{gray} 0.230} & {\color{gray} 0.578} &  {\color{gray} 0.294} \\
DDIB$^*$ (Teacher)~\citep{su2022dual} & {160} & {\color{gray} 38.59} & {\color{gray} 17.03} &  {\color{gray} 0.552} & {\color{gray} 1.594} &  {\color{gray} 0.924}\\
\textbf{IBCD (Ours)} & 1 &  \textbf{48.60}  &  \textbf{18.25} &  \textbf{0.653} & {1.539} & {\textbf{0.921}} \\
\textbf{IBCD$\dagger$ (Ours)} & 1 &  \textbf{46.06} & 16.78  & \textbf{0.612} & {1.583} & \textbf{0.919}\\

\midrule
\multicolumn{7}{c}{\textbf{Male$\rightarrow$Female}}\\
\midrule
CUT~\citep{park2020contrastive} & 1 & 31.94 &  19.87 & 0.74 & - & - \\
UNSB$^*$~\citep{kim2023unpaired} &  5  & 28.62 & 19.55 & 0.687 & {0.576} & {0.635} \\
\cmidrule(lr){1-7}
ILVR~\citep{choi2021ilvr} & {1000} & 46.12 &  18.59 & 0.510 & - & - \\
SDEdit~\citep{meng2022sdedit} & 1000 & 49.43 & 20.03 & 0.572 & {0.782} & {0.380 }\\
EGSDE~\citep{zhao2022egsde} & 1000 & 41.93 & 20.35 & 0.574& {0.875} & {0.437} \\
EGSDE$\dagger$~\citep{zhao2022egsde} & 1200 & 30.61 & 18.32 & 0.510 & {0.955} & {0.621} \\
SDDM~\citep{sun2023sddm} & 100 & 44.37 & - &  0.526 & - & - \\
\midrule
GPT-Image-1~(Foundation)~\citep{openai_image_generation_api_2025} & $\gg$1 & {\color{gray} 60.50} & {\color{gray} 13.11} &  {\color{gray} 0.381} & {\color{gray} 0.503} &  {\color{gray} 0.269} \\
DDIB$^*$ (Teacher)~\citep{su2022dual} & {160} & {\color{gray} 23.69} & {\color{gray} 18.70} &  {\color{gray} 0.664} & {\color{gray} 0.969} &  {\color{gray} 0.808} \\
\textbf{IBCD (Ours)} & 1 &  \textbf{24.93}  & \textbf{20.51} &  \textbf{0.749} & {\textbf{1.160}} & {\textbf{0.814}} \\
\textbf{IBCD$\dagger$ (Ours)} & 1 &  \textbf{24.70} &  20.11 & \textbf{0.744}  & \textbf{1.145} & \textbf{0.815} \\

\bottomrule
\end{tabular}
\end{small}
}
\vspace{-1.4em}
\end{table*}

\begin{figure*}[!t]
    \centering
    \includegraphics[width=0.70\textwidth]{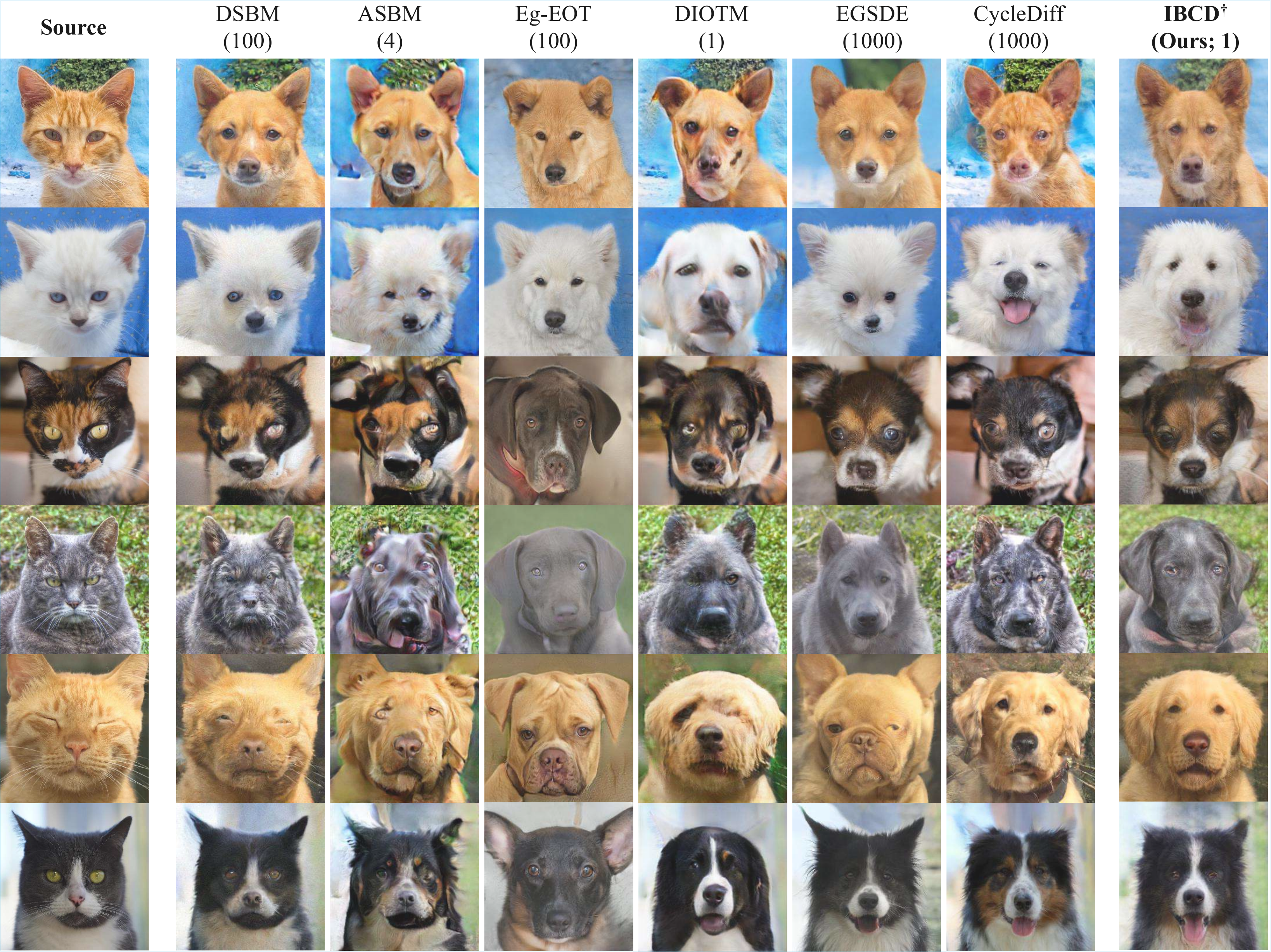}
    \caption{\textbf{Qualitative comparison of unpaired image-to-image translation tasks (Cat$\rightarrow$Dog)}. Compared to other baselines, our model achieves more realistic and source-faithful translations in a single step. The numbers in parentheses represent inference NFE. \rev{The teacher is omitted for space (App.~\ref{supp:further_exp}); sources are a fixed-seed draw, not a curated one (App.~\ref{supp:failure}).}}
    \label{fig:result_main}
    \vspace{-0.2em}  
\end{figure*}

\subsection{Unpaired Image-to-Image Translation}

\para{Evaluation.} Following EGSDE's protocol, we tested Cat$\rightarrow$Dog and Wild$\rightarrow$Dog (AFHQ)~\citep{choi2020stargan} and Male$\rightarrow$Female (CelebA-HQ)~\citep{karras2017progressive}; the Cat$\leftrightarrow$Dog and Wild$\leftrightarrow$Dog students were distilled from an AFHQ teacher DM and Male$\leftrightarrow$Female from a CelebA-HQ teacher DM. We report FID~\citep{heusel2017gans} and Density-Coverage~\citep{naeem2020reliable} for realism, PSNR, SSIM~\citep{wang2004image}, LPIPS~\citep{zhang2018unreasonable} and CLIP score~\citep{hessel2021clipscore} for faithfulness, and a user study.

\para{Comparison results.} Fig.~\ref{fig:intro-tradeoff}, Fig.~\ref{fig:result_main} and Tab.~\ref{table:main} compare IBCD with the baselines. \rev{IBCD and IBCD$^\dagger$ are the \emph{same} network at two points of one trade-off curve, not two methods: IBCD$^\dagger$ weakens the cycle loss relative to DMCD and distils for more steps, buying realism with source faithfulness (settings in Tab.~\ref{table:hyperparam}).} At one NFE, IBCD attains lower FID than every non-teacher diffusion- and bridge-based baseline on Cat$\rightarrow$Dog and Wild$\rightarrow$Dog, the highest SSIM, and the highest target coverage among non-teacher methods on all three tasks; the user study and perceptual metrics confirm the gain in human preference \rev{(twenty blinded raters; all numbers in App.~\ref{supp:userstudy})}.

\rev{\para{The cost of distillation.} Single-step distillation costs distributional realism, and our numbers are an instance of it: on all three tasks IBCD's FID trails that of the 160-NFE DDIB teacher (Tab.~\ref{table:main}), and IBCD$^\dagger$ narrows every gap without closing it. The loss is genuine, plausibly from collapsing a 160-step trajectory into one evaluation, and it is the price of the $120\times$ speed-up of Tab.~\ref{table:inference}, reached with roughly a third of the diffusion baselines' parameters. One case runs the other way: on the reverse Dog$\rightarrow$Cat task IBCD beats the teacher on FID (App.~\ref{supp:further_exp}), and blinded raters prefer IBCD to the teacher on realism.}

\rev{\para{Comparison with single-step adversarial models.} Feed-forward adversarial generators are the fastest translators at inference (Tab.~\ref{table:inference}); Tab.~\ref{table:main} carries two of them, StarGAN~v2 and CUT. On every task where they are reported, their FID is far worse than ours. IBCD also differs in form: one non-adversarial network, distilled from a diffusion teacher, that gives both directions and their composition at one NFE and carries over without adversarial tuning to brain MRI (Tab.~\ref{table:mri}).}

\rev{\para{What PSNR measures.}\label{sec:psnr_caveat} Similarity to the source measures how little an image changed; translation fidelity measures whether what changed was what the task required. With no ground-truth translation, the PSNR and SSIM columns of Tab.~\ref{table:main} compare a translation with its \emph{source}, so they report the first and not the second, and a model that translates less scores higher, as \citet{torbunov2023uvcganv2} also note. Three other lines of evidence carry the faithfulness claim. The BraTS data are \emph{paired}, so PSNR and SSIM there are computed against the true target contrast, where under-translation would score worse and IBCD still leads every baseline and its teacher (Tab.~\ref{table:mri}). Blinded raters preferred IBCD to every baseline on realism and overall, and to ILVR, EGSDE and the teacher on faithfulness (App.~\ref{supp:userstudy}). And under-translation would leave poor FID and low target coverage, whereas IBCD attains the highest coverage of all non-teacher methods on every task.}

\begin{table}[!t]
\addc
\centering
\caption{\textbf{Brain MRI contrast translation} (BraTS2021~\citep{baid2021rsna}, T1$\rightarrow$T2), scored against paired ground truth. Setup in App.~\ref{supp:mri}.}
\label{table:mri}
\renewcommand{\arraystretch}{0.9}
\setlength{\tabcolsep}{26pt}
\resizebox{1.00\linewidth}{!}{
\begin{tabular}{lcc}
\toprule
 & \multicolumn{2}{c}{\textit{vs.\ ground-truth T2}} \\
\textbf{Method}  & PSNR $\uparrow$ & SSIM $\uparrow$ \\
\midrule
SDEdit & 29.85 & 0.849 \\
EGSDE & 23.34 & 0.842 \\
CycleDiffusion & 30.52 & 0.825 \\
\midrule
GPT-Image-1 (Foundation) & \multicolumn{2}{c}{\color{gray} \textit{No valid translation}} \\
DDIB$^*$ (Teacher) & {\color{gray} 30.24} & {\color{gray} 0.825}  \\
\textbf{IBCD (Ours)}  & \textbf{33.28} & \textbf{0.855}   \\
\bottomrule
\end{tabular}
}
\vspace{0.0em}
\end{table}

\begin{figure}[!t]
    \centering
    \includegraphics[width=1.0\columnwidth]{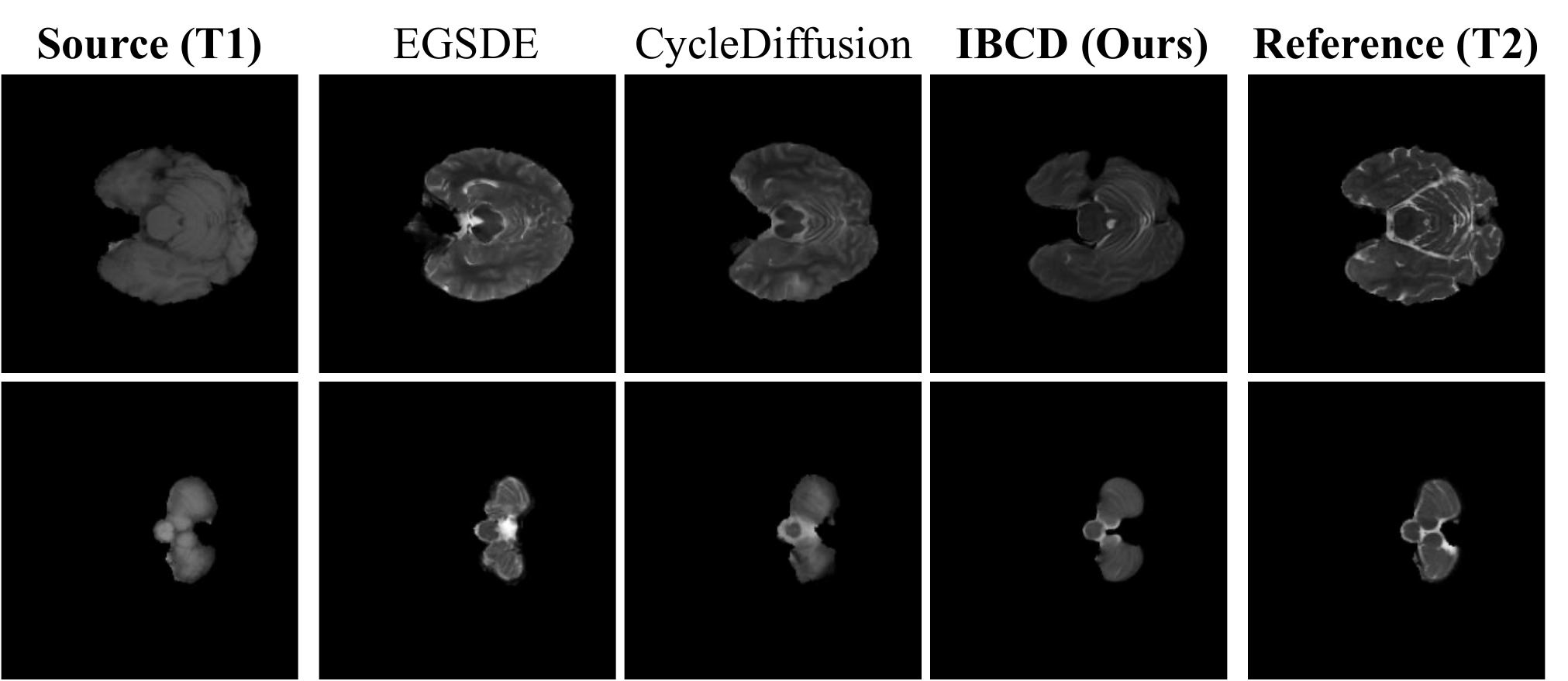}
    \caption{\rev{\textbf{Brain MRI contrast translation} (BraTS2021, T1$\rightarrow$T2), with the paired T2 reference. More in App.~\ref{supp:mri}.}}
    \label{fig:mri_main}
    \vspace{0.0em}
\end{figure}

\begin{table}[!t]
\centering
\small
\caption{\rev{\textbf{Inference cost on the Cat$\rightarrow$Dog task}, on a single NVIDIA RTX 4090 at batch size 1 (except GPT-Image-1). One representative method per family; the full 15-row table is in App.~\ref{supp:further_exp}.}}
\label{table:inference}
\renewcommand{\arraystretch}{0.9}
\resizebox{1.00\linewidth}{!}{
\begin{tabular}{lcccc}
\toprule
\textbf{Method} & \textbf{Params [M]} & \textbf{NFE} $\downarrow$ & \textbf{Time [$s/img$]} $\downarrow$ & \textbf{Rel.} $\downarrow$\\
\midrule
CUT~\citep{park2020contrastive} & 11.39 & 1 & 0.0070 & 0.58 \\
UNSB~\citep{kim2023unpaired} & 14.69 & 5 & 0.077 & 6.42 \\
EGSDE~\citep{zhao2022egsde} & 147.14 & 1000 & 15.89 & 1324.16 \\
CycleDiffusion~\citep{wu2023latent} & 187.12 & 1000(+100) & 26.03 & 2169.17 \\
\midrule
GPT-Image-1~\citep{openai_image_generation_api_2025} & $\gg1$ & $\gg1$ & 30.32 & 2526.67 \\
DDIB (Teacher)~\citep{su2022dual} & 32.95 & 160 & 1.45 & 120.83 \\
\textbf{IBCD (Ours)} & 32.95 & 1 & 0.012 & 1 \\
\bottomrule
\end{tabular}
}
\vspace{-0.2em}
\end{table}

\rev{\para{Inference cost.} IBCD translates an image in 0.012\,s (Tab.~\ref{table:inference}): $120\times$ faster than the teacher it distils and three orders of magnitude faster than the 1000-step diffusion baselines, with 32.95M parameters against those baselines' 93.56M DDPM backbone (Tab.~\ref{table:supp-inference-time}). Only feed-forward GAN generators are faster, and one network serves both directions.}

\begin{table}[!t]
\centering
\caption{\textbf{Ablation on Cat$\rightarrow$Dog.} Every row is the ablated model at \textit{its own lowest-FID setting}, not the deployed one. \rev{PSNR is reported against the teacher's output, which isolates distillation error, alongside the \emph{source}-referenced PSNR of Tab.~\ref{table:main}.}}
\label{table:result_abl}
\renewcommand{\arraystretch}{0.9}
\resizebox{1.00\linewidth}{!}{
\begin{tabular}{l c c c}
\toprule
\textbf{Component}  & \textbf{FID$\downarrow$} & \multicolumn{2}{c}{\textbf{PSNR $\uparrow$}} \\
 & & \rev{vs.\ teacher output} & \rev{vs.\ source image} \\
\cmidrule{1-1} \cmidrule{2-4}
\rev{IBCD loss only} & 48.12 & 18.27 & \rev{19.02}  \\
+ DMCD & 44.40 & 17.95 & \rev{16.80}   \\
+ DMCD \& Cycle & 44.31 & 18.22 & \rev{17.19}  \\
\textbf{+ adaptive DMCD \& Cycle}  & 44.69 & 18.97 & \rev{18.04} \\
\bottomrule
\end{tabular}
}
\vspace{0.2em}  
\end{table}

\para{Ablation Study.}
We ablated each component on Cat$\rightarrow$Dog, sequentially adding the DMCD loss, the cycle loss and adaptive DMCD to the IBCD-loss-only baseline; Tab.~\ref{table:result_abl} and Fig.~\ref{fig:result_abl} report each ablated model at \textit{its own lowest-FID setting}. Combining all components reduces FID well beyond vanilla IBCD while raising PSNR against the teacher, improving the task's inherent trade-off: DMCD enhances realism through target distribution matching, the cycle loss improves source faithfulness (Fig.~\ref{fig:supp_abl}), and adaptive DMCD trades a little FID for a large PSNR gain by weakening DMCD on the trajectories that the consistency loss alone already handles well.

Common distance functions for the consistency loss, compared at a smaller resolution over 50K iterations, gave similar source faithfulness, but LPIPS yielded noticeably lower FID, \rev{which is why IBCD adopts it (Tab.~\ref{table:supp-abl-loss})}; see App.~\ref{supp:further_exp}.

\section{Conclusion}

In this work, we introduced IBCD, an unpaired bidirectional single-step image translation framework. \rev{By distilling the diffusion implicit bridge with a consistency-distillation formulation on a signed time axis, whose sign selects the target domain,} we achieved bidirectional translation without paired data or adversarial training. Our approach overcomes traditional CD limitations with DMCD and distillation difficulty adaptive weighting. Evaluations on toy, natural and medical images demonstrate its effectiveness, and we believe IBCD advances single-step translation wherever paired data is limited and latency matters.

\section*{Acknowledgements}
\label{sec:ack}

This work was partly supported by the Institute of Information \& Communications Technology Planning \& Evaluation (IITP) grant funded by the Korea government (MSIT) (\mbox{IITP-2027-RS-2026-25614465}, AI Star Fellowship Support Program), and partly by the IITP grant funded by the Korea government (MSIT) (\mbox{No.~2019-0-01842}, Artificial Intelligence Graduate School Program (GIST)). This work was also supported by the National Research Foundation of Korea under Grant \mbox{RS-2024-00336454}.

{
    \small
    \bibliographystyle{ieeenat_fullname}
    \bibliography{citation,new_refs}

@String(ICCV= {Int. Conf. Comput. Vis.})

@String(ECCV= {Eur. Conf. Comput. Vis.})

@String(ICLR = {Int. Conf. Learn. Represent.})

@String(ICCV  = {ICCV})

@String(ECCV  = {ECCV})

@String(ICLR  = {ICLR})

@article{schrodinger1932theorie,
  title={Sur la th{\'e}orie relativiste de l'{\'e}lectron et l'interpr{\'e}tation de la m{\'e}canique quantique},
  author={Schr{\"o}dinger, Erwin},
  journal={Annales de l'institut Henri Poincar{\'e}},
  volume={2},
  number={4},
  pages={269--310},
  year={1932}
}

@article{kim2021noise2score,
  title={{{Noise2Score}}: {{Tweedie}}’s approach to self-supervised image denoising without clean images},
  author={Kim, Kwanyoung and Ye, Jong Chul},
  journal={Advances in Neural Information Processing Systems},
  volume={34},
  pages={864--874},
  year={2021}
}

@article{efron2011tweedie,
  title={{{Tweedie}}’s formula and selection bias},
  author={Efron, Bradley},
  journal={Journal of the American Statistical Association},
  volume={106},
  number={496},
  pages={1602--1614},
  year={2011},
  publisher={Taylor \& Francis}
}

@inproceedings{pix2pix,
	title={Image-to-image translation with conditional adversarial networks},
	author={Isola, Phillip and Zhu, Jun-Yan and Zhou, Tinghui and Efros, Alexei A},
	booktitle={Proceedings of the IEEE conference on computer vision and pattern recognition},
	pages={1125--1134},
	year={2017}
}

@article{fid,
	title={Gans trained by a two time-scale update rule converge to a local nash equilibrium},
	author={Heusel, Martin and Ramsauer, Hubert and Unterthiner, Thomas and Nessler, Bernhard and Hochreiter, Sepp},
	journal={Advances in neural information processing systems},
	volume={30},
	year={2017}
}

@InProceedings{not,
author = {Alexander Korotin and Daniil Selikhanovych and Evgeny Burnaev},
title = {Neural Optimal Transport},
booktitle = {ICLR},
year = {2023}
}

@inproceedings{
  song2021scorebased,
  title={Score-Based Generative Modeling through Stochastic Differential Equations},
  author={Yang Song and Jascha Sohl-Dickstein and Diederik P Kingma and Abhishek Kumar and Stefano Ermon and Ben Poole},
  booktitle={International Conference on Learning Representations},
  year={2021},
  url={https://openreview.net/forum?id=PxTIG12RRHS}
}

@article{ho2020denoising,
  title={Denoising diffusion probabilistic models},
  author={Ho, Jonathan and Jain, Ajay and Abbeel, Pieter},
  journal={Advances in neural information processing systems},
  volume={33},
  pages={6840--6851},
  year={2020}
}

@article{vincent2011connection,
  title={A connection between score matching and denoising autoencoders},
  author={Vincent, Pascal},
  journal={Neural computation},
  volume={23},
  number={7},
  pages={1661--1674},
  year={2011},
  publisher={MIT Press}
}

@article{lu2022dpm,
  title={{{DPM-Solver}}: A fast {{ODE}} solver for diffusion probabilistic model sampling in around 10 steps},
  author={Lu, Cheng and Zhou, Yuhao and Bao, Fan and Chen, Jianfei and Li, Chongxuan and Zhu, Jun},
  journal={Advances in Neural Information Processing Systems},
  volume={35},
  pages={5775--5787},
  year={2022}
}

@article{karras2022elucidating,
  title={Elucidating the design space of diffusion-based generative models},
  author={Karras, Tero and Aittala, Miika and Aila, Timo and Laine, Samuli},
  journal={Advances in neural information processing systems},
  volume={35},
  pages={26565--26577},
  year={2022}
}

@inproceedings{
      su2022dual,
      title={Dual Diffusion Implicit Bridges for Image-to-Image Translation},
      author={Su, Xuan and Song, Jiaming and Meng, Chenlin and Ermon, Stefano},
      booktitle={International Conference on Learning Representations},
      year={2023},
}

@inproceedings{zhu2017unpaired,
  title={Unpaired image-to-image translation using cycle-consistent adversarial networks},
  author={Zhu, Jun-Yan and Park, Taesung and Isola, Phillip and Efros, Alexei A},
  booktitle={Proceedings of the IEEE international conference on computer vision},
  pages={2223--2232},
  year={2017}
}

@inproceedings{choi2018stargan,
  title={{{StarGAN}}: Unified generative adversarial networks for multi-domain image-to-image translation},
  author={Choi, Yunjey and Choi, Minje and Kim, Munyoung and Ha, Jung-Woo and Kim, Sunghun and Choo, Jaegul},
  booktitle={Proceedings of the IEEE conference on computer vision and pattern recognition},
  pages={8789--8797},
  year={2018}
}

@inproceedings{song2023consistency,
  title={Consistency Models},
  author={Song, Yang and Dhariwal, Prafulla and Chen, Mark and Sutskever, Ilya},
  booktitle={International Conference on Machine Learning},
  pages={32211--32252},
  year={2023},
  organization={PMLR}
}

@inproceedings{yin2024one,
  title={One-step diffusion with distribution matching distillation},
  author={Yin, Tianwei and Gharbi, Micha{\"e}l and Zhang, Richard and Shechtman, Eli and Durand, Fredo and Freeman, William T and Park, Taesung},
  booktitle={Proceedings of the IEEE/CVF Conference on Computer Vision and Pattern Recognition},
  pages={6613--6623},
  year={2024}
}

@article{goodfellow2014generative,
  title={Generative adversarial nets},
  author={Goodfellow, Ian and Pouget-Abadie, Jean and Mirza, Mehdi and Xu, Bing and Warde-Farley, David and Ozair, Sherjil and Courville, Aaron and Bengio, Yoshua},
  journal={Advances in neural information processing systems},
  volume={27},
  year={2014}
}

@inproceedings{kim2023unpaired,
  title={Unpaired image-to-image translation via neural {{Schr{\"o}dinger}} bridge},
  author={Kim, Beomsu and Kwon, Gihyun and Kim, Kwanyoung and Ye, Jong Chul},
  booktitle={International Conference on Learning Representations},
  year={2024}
}

@article{kim2023consistency,
  title={Consistency trajectory models: Learning probability flow {{ODE}} trajectory of diffusion},
  author={Kim, Dongjun and Lai, Chieh-Hsin and Liao, Wei-Hsiang and Murata, Naoki and Takida, Yuhta and Uesaka, Toshimitsu and He, Yutong and Mitsufuji, Yuki and Ermon, Stefano},
  journal={International Conference on Learning Representations},
  year={2024}
}

@inproceedings{blau2018perception,
  title={The perception-distortion tradeoff},
  author={Blau, Yochai and Michaeli, Tomer},
  booktitle={Proceedings of the IEEE conference on computer vision and pattern recognition},
  pages={6228--6237},
  year={2018}
}

@inproceedings{zhang2018unreasonable,
  title={The unreasonable effectiveness of deep features as a perceptual metric},
  author={Zhang, Richard and Isola, Phillip and Efros, Alexei A and Shechtman, Eli and Wang, Oliver},
  booktitle={Proceedings of the IEEE conference on computer vision and pattern recognition},
  pages={586--595},
  year={2018}
}

@article{parmar2024one,
  title={One-step image translation with text-to-image models},
  author={Parmar, Gaurav and Park, Taesung and Narasimhan, Srinivasa and Zhu, Jun-Yan},
  journal={arXiv preprint arXiv:2403.12036},
  year={2024}
}

@article{li2024bidirectional,
  title={Bidirectional Consistency Models},
  author={Li, Liangchen and He, Jiajun},
  journal={arXiv preprint arXiv:2403.18035},
  year={2024}
}

@article{zhao2022egsde,
  title={{{EGSDE}}: Unpaired image-to-image translation via energy-guided stochastic differential equations},
  author={Zhao, Min and Bao, Fan and Li, Chongxuan and Zhu, Jun},
  journal={Advances in Neural Information Processing Systems},
  volume={35},
  pages={3609--3623},
  year={2022}
}

@article{heusel2017gans,
  title={{{GANs}} trained by a two time-scale update rule converge to a local {{Nash}} equilibrium},
  author={Heusel, Martin and Ramsauer, Hubert and Unterthiner, Thomas and Nessler, Bernhard and Hochreiter, Sepp},
  journal={Advances in neural information processing systems},
  volume={30},
  year={2017}
}

@article{wang2004image,
  title={Image quality assessment: from error visibility to structural similarity},
  author={Wang, Zhou and Bovik, Alan C and Sheikh, Hamid R and Simoncelli, Eero P},
  journal={IEEE transactions on image processing},
  volume={13},
  number={4},
  pages={600--612},
  year={2004},
  publisher={IEEE}
}

@inproceedings{karras2017progressive,
  title={Progressive growing of {{GANs}} for improved quality, stability, and variation},
  author={Karras, Tero and Aila, Timo and Laine, Samuli and Lehtinen, Jaakko},
  booktitle={International Conference on Learning Representations},
  year={2018}
}

@inproceedings{choi2020stargan,
  title={{{StarGAN}} v2: Diverse image synthesis for multiple domains},
  author={Choi, Yunjey and Uh, Youngjung and Yoo, Jaejun and Ha, Jung-Woo},
  booktitle={Proceedings of the IEEE/CVF conference on computer vision and pattern recognition},
  pages={8188--8197},
  year={2020}
}

@book{ascher1998computer,
  title={Computer methods for ordinary differential equations and differential-algebraic equations},
  author={Ascher, Uri M and Petzold, Linda R},
  year={1998},
  publisher={SIAM}
}

@inproceedings{
      meng2022sdedit,
      title={{{SDE}}dit: Guided image synthesis and editing with stochastic differential equations},
      author={Chenlin Meng and Yutong He and Yang Song and Jiaming Song and Jiajun Wu and Jun-Yan Zhu and Stefano Ermon},
      booktitle={International Conference on Learning Representations},
      year={2022},
}

@inproceedings{wu2023latent,
  title={A latent space of stochastic diffusion models for zero-shot image editing and guidance},
  author={Wu, Chen Henry and De la Torre, Fernando},
  booktitle={Proceedings of the IEEE/CVF International Conference on Computer Vision},
  pages={7378--7387},
  year={2023}
}

@inproceedings{sun2023sddm,
  title={{{SDDM}}: Score-decomposed diffusion models on manifolds for unpaired image-to-image translation},
  author={Sun, Shikun and Wei, Longhui and Xing, Junliang and Jia, Jia and Tian, Qi},
  booktitle={International Conference on Machine Learning},
  pages={33115--33134},
  year={2023},
  organization={PMLR}
}

@article{benaim2017one,
  title={One-sided unsupervised domain mapping},
  author={Benaim, Sagie and Wolf, Lior},
  journal={Advances in neural information processing systems},
  volume={30},
  year={2017}
}

@inproceedings{fu2019geometry,
  title={Geometry-consistent generative adversarial networks for one-sided unsupervised domain mapping},
  author={Fu, Huan and Gong, Mingming and Wang, Chaohui and Batmanghelich, Kayhan and Zhang, Kun and Tao, Dacheng},
  booktitle={Proceedings of the IEEE/CVF conference on computer vision and pattern recognition},
  pages={2427--2436},
  year={2019}
}

@inproceedings{zheng2021spatially,
  title={The spatially-correlative loss for various image translation tasks},
  author={Zheng, Chuanxia and Cham, Tat-Jen and Cai, Jianfei},
  booktitle={Proceedings of the IEEE/CVF conference on computer vision and pattern recognition},
  pages={16407--16417},
  year={2021}
}

@inproceedings{park2020contrastive,
  title={Contrastive learning for unpaired image-to-image translation},
  author={Park, Taesung and Efros, Alexei A and Zhang, Richard and Zhu, Jun-Yan},
  booktitle={Computer Vision -- ECCV 2020},
  pages={319--345},
  year={2020},
  organization={Springer}
}

@inproceedings{choi2021ilvr,
  title={{{ILVR}}: Conditioning method for denoising diffusion probabilistic models},
  author={Choi, Jooyoung and Kim, Sungwon and Jeong, Yonghyun and Gwon, Youngjune and Yoon, Sungroh},
  booktitle={Proceedings of the IEEE/CVF International Conference on Computer Vision (ICCV)},
  year={2021}
}

@inproceedings{naeem2020reliable,
  title={Reliable fidelity and diversity metrics for generative models},
  author={Naeem, Muhammad Ferjad and Oh, Seong Joon and Uh, Youngjung and Choi, Yunjey and Yoo, Jaejun},
  booktitle={International Conference on Machine Learning},
  pages={7176--7185},
  year={2020},
  organization={PMLR}
}

@inproceedings{simonyan2014very,
  title={Very deep convolutional networks for large-scale image recognition},
  author={Simonyan, Karen and Zisserman, Andrew},
  booktitle={International Conference on Learning Representations},
  year={2015}
}

@inproceedings{hertz2023delta,
  title={Delta denoising score},
  author={Hertz, Amir and Aberman, Kfir and Cohen-Or, Daniel},
  booktitle={Proceedings of the IEEE/CVF International Conference on Computer Vision},
  pages={2328--2337},
  year={2023}
}

@article{wang2023prolificdreamer,
  title={{{ProlificDreamer}}: High-fidelity and diverse text-to-{{3D}} generation with variational score distillation},
  author={Wang, Zhengyi and Lu, Cheng and Wang, Yikai and Bao, Fan and Li, Chongxuan and Su, Hang and Zhu, Jun},
  journal={Advances in Neural Information Processing Systems},
  volume={36},
  pages={8406--8441},
  year={2023}
}

@article{kaji2019overview,
  title={Overview of image-to-image translation by use of deep neural networks: denoising, super-resolution, modality conversion, and reconstruction in medical imaging},
  author={Kaji, Shizuo and Kida, Satoshi},
  journal={Radiological physics and technology},
  volume={12},
  number={3},
  pages={235--248},
  year={2019},
  publisher={Springer}
}

@article{chen2023deep,
  title={Deep learning based unpaired image-to-image translation applications for medical physics: a systematic review},
  author={Chen, Junhua and Chen, Shenlun and Wee, Leonard and Dekker, Andre and Bermejo, Inigo},
  journal={Physics in Medicine \& Biology},
  volume={68},
  number={5},
  pages={05TR01},
  year={2023},
  publisher={IOP Publishing}
}

@article{safayani2026unpaired,
  title={Unpaired image-to-image translation with content preserving perspective: A review},
  author={Safayani, Mehran and Mirzapour, Behnaz and Aghaebrahimiyan, Hanieh and Salehi, Nasrin and Ravaee, Hamid},
  journal={Multimedia Tools and Applications},
  volume={85},
  number={6},
  pages={527},
  year={2026},
  publisher={Springer}
}

@article{zheng2022ittr,
  title={{{ITTR}}: Unpaired image-to-image translation with transformers},
  author={Zheng, Wanfeng and Li, Qiang and Zhang, Guoxin and Wan, Pengfei and Wang, Zhongyuan},
  journal={arXiv preprint arXiv:2203.16015},
  year={2022}
}

@article{saad2024survey,
  title={A survey on training challenges in generative adversarial networks for biomedical image analysis},
  author={Saad, Muhammad Muneeb and O’Reilly, Ruairi and Rehmani, Mubashir Husain},
  journal={Artificial Intelligence Review},
  volume={57},
  number={2},
  pages={19},
  year={2024},
  publisher={Springer}
}

@inproceedings{parmar2023zero,
  title={Zero-shot image-to-image translation},
  author={Parmar, Gaurav and Kumar Singh, Krishna and Zhang, Richard and Li, Yijun and Lu, Jingwan and Zhu, Jun-Yan},
  booktitle={ACM SIGGRAPH 2023 conference proceedings},
  pages={1--11},
  year={2023}
}

@inproceedings{hertz2022prompt,
  title={Prompt-to-prompt image editing with cross-attention control},
  author={Hertz, Amir and Mokady, Ron and Tenenbaum, Jay and Aberman, Kfir and Pritch, Yael and Cohen-Or, Daniel},
  booktitle={International Conference on Learning Representations},
  year={2023}
}

@inproceedings{hessel2021clipscore,
  title={{{CLIPScore}}: A reference-free evaluation metric for image captioning},
  author={Hessel, Jack and Holtzman, Ari and Forbes, Maxwell and Bras, Ronan Le and Choi, Yejin},
  booktitle={Proceedings of the 2021 Conference on Empirical Methods in Natural Language Processing (EMNLP)},
  year={2021}
}

@misc{openai_image_generation_api_2025,
  author       = {{OpenAI}},
  title        = {Image Generation {{API}}},
  howpublished = {Online},
  year         = {2025},
  month        = {April},
  note         = {Accessed July 21, 2025},
  url          = {https://openai.com/index/image-generation-api/}
}

@misc{openai_gpt4_1_2025,
  author       = {OpenAI},
  title        = {Introducing {{GPT-4.1}} in the {{API}}},
  howpublished = {Online},
  year         = {2025},
  month        = {April},
  note         = {Accessed July 21, 2025},
  url          = {https://openai.com/index/gpt-4-1/}
}

@misc{qwen_vlo_2025,
  author       = {Qwen Team},
  title        = {{{Qwen VLo}}: From ``Understanding'' the World to ``Depicting'' It},
  howpublished = {Online},
  year         = {2025},
  month        = {June},
  note         = {Accessed July 28, 2025},
  url          = {https://qwenlm.github.io/blog/qwen-vlo/}
}

@article{baid2021rsna,
  title={The {{RSNA-ASNR-MICCAI}} {{BraTS}} 2021 benchmark on brain tumor segmentation and radiogenomic classification},
  author={Baid, Ujjwal and Ghodasara, Satyam and Mohan, Suyash and Bilello, Michel and Calabrese, Evan and Colak, Errol and Farahani, Keyvan and Kalpathy-Cramer, Jayashree and Kitamura, Felipe C and Pati, Sarthak and others},
  journal={arXiv preprint arXiv:2107.02314},
  year={2021}
}

@article{rakitin2024regularized,
  title={Regularized distribution matching distillation for one-step unpaired image-to-image translation},
  author={Rakitin, Denis and Shchekotov, Ivan and Vetrov, Dmitry},
  journal={arXiv preprint arXiv:2406.14762},
  year={2024}
}

@inproceedings{
korotin2022neural,
title={Neural optimal transport},
author={Korotin, Alexander and Selikhanovych, Daniil and Burnaev, Evgeny},
booktitle={The Eleventh International Conference on Learning Representations },
year={2023}
}

@article{gushchin2024adversarial,
  title={Adversarial schr{\"o}dinger bridge matching},
  author={Gushchin, Nikita and Selikhanovych, Daniil and Kholkin, Sergei and Burnaev, Evgeny and Korotin, Alexander},
  journal={Advances in Neural Information Processing Systems},
  volume={37},
  pages={89612--89651},
  year={2024}
}

@article{shi2023diffusion,
  title={Diffusion {{Schr{\"o}dinger}} bridge matching},
  author={Shi, Yuyang and De Bortoli, Valentin and Campbell, Andrew and Doucet, Arnaud},
  journal={Advances in Neural Information Processing Systems},
  volume={36},
  pages={62183--62223},
  year={2023}
}

@inproceedings{
      mokrov2024energyguided,
      title={Energy-guided Entropic Neural Optimal Transport},
      author={Petr Mokrov and Alexander Korotin and Alexander Kolesov and Nikita Gushchin and Evgeny Burnaev},
      booktitle={The Twelfth International Conference on Learning Representations},
      year={2024},
      url={https://openreview.net/forum?id=d6tUsZeVs7}
}

@inproceedings{
      choi2025improving,
      title={Improving neural optimal transport via displacement interpolation},
      author={Choi, Jaemoo and Chen, Yongxin and Choi, Jaewoong},
      booktitle={The Thirteenth International Conference on Learning Representations},
      year={2025},
}

@article{peyre2019computational,
  title={Computational optimal transport: With applications to data science},
  author={Peyr{\'e}, Gabriel and Cuturi, Marco},
  journal={Foundations and Trends{\textregistered} in Machine Learning},
  volume={11},
  number={5-6},
  pages={355--607},
  year={2019},
  publisher={Now Publishers, Inc.}
}

@inproceedings{dou2024theory,
  title     = {Theory of Consistency Diffusion Models: Distribution Estimation Meets Fast Sampling},
  author    = {Dou, Zehao and Chen, Minshuo and Wang, Mengdi and Yang, Zhuoran},
  booktitle = {Proceedings of the 41st International Conference on Machine Learning},
  series    = {Proceedings of Machine Learning Research},
  volume    = {235},
  year      = {2024},
  publisher = {PMLR},
}

@article{torbunov2023uvcganv2,
  title         = {{{UVCGAN}} v2: An improved cycle-consistent {{GAN}} for unpaired image-to-image translation},
  author        = {Torbunov, Dmitrii and Huang, Yi and Tseng, Huan-Hsin and Yu, Haiwang and Huang, Jin and Yoo, Shinjae and Lin, Meifeng and Viren, Brett and Ren, Yihui},
  journal       = {arXiv preprint arXiv:2303.16280},
  year          = {2023},
  eprint        = {2303.16280},
  archivePrefix = {arXiv},
  primaryClass  = {cs.CV},
}

@inproceedings{zheng2025revisiting,
  title     = {Revisiting Diffusion Models: From Generative Pre-training to One-Step Generation},
  author    = {Zheng, Bowen and Yang, Tianming},
  booktitle = {Proceedings of the 42nd International Conference on Machine Learning},
  series    = {Proceedings of Machine Learning Research},
  year      = {2025},
  publisher = {PMLR},
  note      = {arXiv:2506.09376},
}
}

\clearpage
\onecolumn
\begin{center}
  {\Large\textbf{\thetitle}}\\
  \vspace{0.5em}{\Large Supplementary Material}\\
  \vspace{1.0em}
\end{center}
\appendix
\numberwithin{theorem}{section}
\setcounter{figure}{0}\setcounter{table}{0}\setcounter{equation}{0}
\renewcommand{\thefigure}{S\arabic{figure}}
\renewcommand{\thetable}{S\arabic{table}}
\renewcommand{\theequation}{S\arabic{equation}}
\renewcommand{\thealgocf}{S\arabic{algocf}}

{
\section*{\rev{Contents}}
\addc

\vspace{0.5em}
{\centering
\small
\begin{tabular}{@{}p{0.05\textwidth}p{0.325\textwidth}p{0.535\textwidth}@{}}
\toprule
\textbf{App.} & \textbf{Title} & \textbf{What it contains} \\
\midrule
\ref{supp:algorithm}   & Algorithms & Pseudocode for vanilla and final IBCD, loss by loss. \\
\ref{supp:ddtheory}    & Theoretical Analysis of Distillation Difficulty & Assumptions, the restated Proposition~\ref{prop:dd_main} and its proof, the caveats it comes with, and a switch criterion for the staged schedule. \\
\ref{supp:extend-edm}  & Model Parametrization & EDM/CD preconditioning and timestep rescaling extended to signed time. \\
\ref{sec:ibcd_theory}  & Theoretical Robustness of IBCD & Lemma~\ref{lem:piecewise_consistency}: CD's error bound survives the junction at which the domain embedding switches. \\
\ref{supp:implementation} & Implementation Details & Architectures, teacher and distillation schedules, hyperparameters, evaluation protocol, baselines and reproductions. \\
\ref{supp:trainingcost} & Training Cost, Memory Footprint, and the Staged Schedule & What is held in memory, how long each phase runs, a reproducible switch criterion, and the training budget. \\
\midrule
\ref{supp:further_exp} & Further Experimental Results & \\
\ref{supp:meanpred}    & \quad Distillation Error in Vanilla IBCD & Where the one-step map fails on the toy data, and why. \\
\ref{supp:toy_topology} & \quad Topology of the Toy Manifolds & Why the toy translation follows ambient distance rather than position along the manifolds. \\
\ref{supp:userstudy}   & \quad User Study and Perceptual Metrics & Twenty-participant pairwise study, with LPIPS and CLIPScore. \\
\ref{supp:abl_weights} & \quad Effect of the Auxiliary Loss Weights & How the DMCD and cycle weights trade realism against faithfulness. \\
\ref{supp:abl_loss}    & \quad Distance Functions for the IBCD Loss & Full ablation over LPIPS, L1, L2 and pseudo-Huber. \\
\ref{supp:ddw_i2i}     & \quad Approximated Distillation Difficulty & The hardest and easiest real trajectories, translated without selection. \\
\ref{supp:stability}   & \quad Training Stability & Loss curves for a run with no adversarial objective. \\
\ref{supp:same_teacher} & \quad Comparison Under a Shared Teacher & Diffusion baselines re-run on our own teacher, so only the method differs. \\
\ref{supp:inference}   & \quad Model Inference Efficiency & Per-image time, NFE and parameter count for every baseline. \\
\ref{supp:gan_baseline} & \quad Adversarial Distillation of DDIB & The natural GAN-distillation baseline: what we ran, why it diverged, and what remains. \\
\ref{supp:failure}     & \quad Failure Cases & Incomplete translations and broken cycles, shown without selection. \\
\ref{supp:bidirectional} & \quad Bidirectional Translations & Opposite and cycle translation, quantitative and qualitative. \\
\ref{supp:mri}         & \quad Unpaired MRI Contrast Translation & BraTS2021 T1$\rightarrow$T2, scored against paired ground truth. \\
\ref{supp:foundation}  & \quad Limitations of Foundational Models & Where zero-shot editing models fall short on domain-specific tasks. \\
\ref{supp:morequal}    & \quad More Qualitative Results & Additional cycle-translation samples for all three domain pairs. \\
\bottomrule
\end{tabular}
\par}
}        

\clearpage

\section{Algorithms} \label{supp:algorithm}

In this section, we present the vanilla implicit bridge consistency distillation algorithm (Algo.~\ref{alg:ibcd-vanilla}), which utilizes only the IBCD losses. Additionally, we introduce the final implicit bridge consistency distillation algorithm (Algo.~\ref{alg:ibcd-final}), which incorporates all the losses discussed in the text, including DMCD, adaptive weighting strategies, and cycle translation loss, to enhance performance and address the limitations identified in the vanilla version.

\begin{algorithm}[!ht]
\caption{(Vanilla) Implicit Bridge Consistency Distillation (IBCD)}
\label{alg:ibcd-vanilla}

\KwIn{Teacher diffusion model \(\phi\), datasets \(\mathcal{S}_\text{A}\) and \(\mathcal{S}_\text{B}\), class conditions \(c_\text{A}\) and \(c_\text{B}\).}
$k \leftarrow 0, \theta \leftarrow \phi$, $\theta^- \leftarrow \phi$ \\
\Repeat{$\gL_{\text{IBCD}}$ convergence}{
    $c \leftarrow$ \textbf{if} ($k \% 2 == 0$ \textbf{then} $c_\text{A}$ \textbf{else} $c_\text{B}$) \\
    Sample $\mbx^\text{A} \sim \mathcal{S}_\text{A}$, $\mbx^\text{B} \sim \mathcal{S}_\text{B}$ \\
    \If{$c == c_\text{B}$}
    {
        Sample $i \sim \mathcal{U}[-N+1,-1]$, ${j} \sim \mathcal{U}[0,N-2]$
    }
    \Else{
        Sample ${i} \sim \mathcal{U}[-N+2,0]$, ${j} \sim \mathcal{U}[1,N-1]$
    }
    Sample $\mbx_{t_{{i}}} \sim \mathcal{N}(\mbx^\text{A},t_{{i}}^2 \bf{I})$, $\mbx_{t_{{j}}} \sim \mathcal{N}(\mbx^\text{B},t_{{j}}^2 \bf{I})$  \\
    \If{$c == c_\text{B}$}
    {
    Estimate $\hat \mbx_{t_{{i}+1}} , \hat \mbx_{t_{{j}+1}}$ with Eq.~(\ref{eq:forward})
    }
    \Else{
    Estimate $\hat \mbx_{t_{{i}-1}} , \hat \mbx_{t_{{j}-1}}$ with Eq.~(\ref{eq:backward})
    }
    $\mathbf{t}_1 \leftarrow [{t_{{i}}}; {t_{{j}}}], \, {\mathbf{t}_{2}} = [{t_{{i}\pm1}}; {t_{{j}\pm1}}]$ \\
    $\mbx_{\mathbf{t}_1} \leftarrow [\mbx_{t_{{i}}}; \mbx_{t_{{j}}}], \, \hat \mbx_{\mathbf{t}_2} \leftarrow [\hat \mbx_{t_{{i}\pm1}}; \hat \mbx_{t_{{j}\pm 1}}]$ \\
    $\mathcal{L}_{\text{IBCD}} \leftarrow [\lambda(\mathbf{t}_2) d_{\text{IBCD}}({\color{blue} G_{\theta}(\mathbf{x}_{\mathbf{t}_1}, \mathbf{t}_1, c)}, {\color{magenta} G_{\theta^{-}}(\hat{\mbx}_{\mathbf{t}_2}, \mathbf{t}_2, c)})]$ \\
    $\theta \leftarrow \theta - \zeta_{\theta} \nabla_{\theta}\gL_{\text{IBCD}}$ \\
    $\theta^- \leftarrow \texttt{sg}(\mu \theta^- + (1-\mu) \theta)$ \\
    $k \leftarrow k + 1$ \\
}
\KwOut{Unified single-step model \(G_{\theta}\) for bidirectional image translation.}
\end{algorithm}

\begin{algorithm}[!ht]
\caption{(Final) Implicit Bridge Consistency Distillation (IBCD)}
\label{alg:ibcd-final}

\KwIn{Teacher diffusion model \(\phi\), datasets \(\mathcal{S}_\text{A}\) and \(\mathcal{S}_\text{B}\), class conditions \(c_\text{A}\) and \(c_\text{B}\).}
$k \leftarrow 0, \theta \leftarrow \phi$, $\theta^- \leftarrow \phi$, $\psi \leftarrow \phi$ \\
\Repeat{$\gL_{\text{total}}$ convergence}{
    $c \leftarrow$ \textbf{if} ($k \% 2 == 0$ \textbf{then} $c_\text{A}$ \textbf{else} $c_\text{B}$) \\
    Sample $\mbx^\text{A} \sim \mathcal{S}_\text{A}, \; \mbx^\text{B} \sim \mathcal{S}_\text{B}$ \\
    \tcp{}
    \tcp{IBCD loss}
    \If{$c == c_\text{B}$}
    {
        Sample ${i} \sim \mathcal{U}[-N+1,-1], \; {j} \sim \mathcal{U}[0,N-2]$
    }
    \Else{
        Sample ${i} \sim \mathcal{U}[-N+2,0], \; {j} \sim \mathcal{U}[1,N-1]$
    }
    Sample $\mbx_{t_{{i}}} \sim \mathcal{N}(\mbx^\text{A},t_{{i}}^2 \bf{I})$$, \; \mbx_{t_{{j}}} \sim \mathcal{N}(\mbx^\text{B}, t_{{j}}^2 \bf{I})$  \\
    \If{$c == c_\text{B}$}
    {
    Estimate $\hat \mbx_{t_{{i}+1}}, \; \hat \mbx_{t_{{j}+1}}$ with Eq.~(\ref{eq:forward})
    }
    \Else{
    Estimate $\hat \mbx_{t_{{i}-1}}, \; \hat \mbx_{t_{{j}-1}}$ with Eq.~(\ref{eq:backward})
    }
    $\mathbf{t}_1 \leftarrow [{t_{{i}}}; {t_{{j}}}], \; {\mathbf{t}_{2}} = [{t_{{i}\pm1}}; {t_{{j}\pm1}}]$ \\
    $\mbx_{\mathbf{t}_1} \leftarrow [\mbx_{t_{{i}}}; \mbx_{t_{{j}}}], \; \hat \mbx_{\mathbf{t}_2} \leftarrow [\hat \mbx_{t_{{i}\pm1}}; \hat \mbx_{t_{{j}\pm 1}}]$ \\
    $\mathcal{L}_{\text{IBCD}} \leftarrow [\lambda(\mathbf{t}_2) d_{\text{IBCD}}({\color{blue} G_{\theta}(\mathbf{x}_{\mathbf{t}_1}, \mathbf{t}_1, c)}, {\color{magenta} G_{\theta^{-}}(\hat{\mbx}_{\mathbf{t}_2}, \mathbf{t}_2, c)})]$ \\
    \tcp{}
    \tcp{DMCD loss}
    Sample $i \sim \mathcal{U}[0, N-1]$ \\ 
    Sample $\mbx_{t_i} \sim \gN({\color{blue} G_{\theta}(\mbx_{\mathbf{t}_{1}}, \mathbf{t}_1, c)}, t_i^2 \mathbf{I})$ \\
    $\Hat{\mathcal{D}} \leftarrow \texttt{sg}( g(d_{\text{DMCD}}({\color{blue} G_{\theta}(\mathbf{x}_{\mathbf{t}_1}, \mathbf{t}_1, c)}, {\color{magenta} G_{\theta^{-}}(\hat{\mbx}_{\mathbf{t}_2}, \mathbf{t}_2, c)}))) $ \\
    $\nabla_\theta \mathcal{L}_{\text{DMCD}} \leftarrow w_{t_i} 
 \Hat{\mathcal{D}} \cdot (s_{\psi}(\mbx_{t_i},t_i,c) - s_{\phi}(\mbx_{t_i}, t_i, c)) \nabla_{\theta} {\color{blue} G_{\theta}(\mbx_{\mathbf{t}_1},\mathbf{t}_1,c)}$ \\
    \tcp{}
    \tcp{Cycle loss}
    Sample $\mbx_{{\epsilon(c_\text{A})}} \sim \mathcal{N}(\mbx^\text{A},\sigma_\text{min}^2 \bf{I})$$,\; \mbx_{{\epsilon(c_\text{B})}} \sim \mathcal{N}(\mbx^\text{B},\sigma_\text{min}^2 \bf{I})$ \\
    $\mathbf{t}_3 \leftarrow [\epsilon(c_\text{A}); \epsilon(c_\text{B})], \; \mathbf{t}_4 \leftarrow [\epsilon(c_\text{B}); \epsilon(c_\text{A})]$ \\
    $\mathbf{c}_3 \leftarrow [c_\text{B}; c_\text{A}], \; \mathbf{c}_4 \leftarrow [c_\text{A}; c_\text{B}]$\\
    $\mbx_{\mathbf{t}_3} \leftarrow [\mbx_{{\epsilon(c_\text{A})}}; \mbx_{{\epsilon(c_\text{B})}}]$ \\
    $\mathcal{L}_\text{cycle} \leftarrow d_{\text{cycle}}({\color{brown} G_\theta(}{\color{teal} G_\theta(\mbx_{\mathbf{t}_3}, \mathbf{t}_3, \mathbf{c}_3)}{\color{brown}, \mathbf{t}_4, \mathbf{c}_4)}, \mbx_{\mathbf{t}_3})$\\
    \tcp{}
    \tcp{Optimize the student}
    $\nabla_\theta \mathcal{L}_\text{total} \leftarrow \nabla_\theta \mathcal{L}_\text{IBCD} + \lambda_\text{DMCD} \nabla_\theta \mathcal{L}_\text{DMCD} + \lambda_\text{cycle} \nabla_\theta \mathcal{L}_\text{cycle}$\\
    $\theta \leftarrow \theta - \zeta_{\theta} \nabla_{\theta}\gL_{\text{total}}$ \\
    $\theta^- \leftarrow \texttt{sg}(\mu \theta^- + (1-\mu) \theta)$ \\
    \tcp{}
    \tcp{Optimize the fake DM}
    $\mathcal{L}_{DSM} \leftarrow$ DSM loss of EDM with sample ${\color{blue} G_{\theta}(\mathbf{x}_{\mathbf{t}_1}, \mathbf{t}_1, c)}$, class condition $c$, and fake DM \rev{$\psi$} \\
    \rev{$\psi \leftarrow \psi - \zeta_{\psi} \nabla_{\psi}\gL_{\text{DSM}}$} \\
    $k \leftarrow k + 1$ \\
}
\KwOut{Unified single-step model \(G_{\theta}\) for bidirectional image translation.}
\end{algorithm}

\clearpage
%

{\addc   
\section{\texorpdfstring{\rev{Theoretical Analysis of Distillation Difficulty}}{Theoretical Analysis of Distillation Difficulty}}
\label{supp:ddtheory}

This appendix restates and proves Proposition~\ref{prop:dd_main} of the main paper. Sec.~\ref{supp:ddtheory_setup} fixes the setup and the assumptions, Sec.~\ref{supp:ddtheory_bound} states the proposition in full and proves it, Sec.~\ref{supp:ddtheory_why} draws the consequence for the DMCD weight and Sec.~\ref{supp:ddtheory_switch} for the staged schedule.

Throughout, all statements are relative to the \emph{empirical} PF-ODE, \ie the ODE obtained by substituting the trained teacher into Eq.~(\ref{eq:pf-ode}), exactly as in \citet{song2023consistency}. Consequently $\mathcal{D}$ measures faithfulness to the DDIB teacher, not to any ground-truth correspondence between the two domains; no such correspondence exists in the unpaired setting.

\subsection{Setup and Assumptions}
\label{supp:ddtheory_setup}

\textbf{Signed-time grid.} We use the discretization of Eq.~(\ref{eq:t_i}), $t_i = \text{sign}(i)\,\sigma_{|i|}$ with $\sigma_0 = \sigma_\text{max} > \sigma_1 > \cdots > \sigma_{N-1} = \sigma_\text{min}$ and $\sigma_N = 0$ (schedule in App.~\ref{supp:implementation}). It is convenient to index a translation by its target condition: for $c \in \{c_\text{A}, c_\text{B}\}$ put
\begin{equation}
s(c) = \begin{cases} +1 & c = c_\text{B} \\ -1 & c = c_\text{A}\end{cases},
\qquad
\mathcal{I}_{c_\text{B}} = \{-N{+}1, \dots, N{-}2\},
\qquad
\mathcal{I}_{c_\text{A}} = \{-N{+}2, \dots, N{-}1\},
\end{equation}
so that $|\mathcal{I}_c| = 2N-2$ and the pair generation of Eqs.~(\ref{eq:forward}),~(\ref{eq:backward}) reads uniformly: for $n \in \mathcal{I}_c$ the source index is $n$ and the distillation target index is $n + s(c)$, one step closer to the target boundary $\epsilon(c)$, whose index we write $n_\epsilon(c)$, with $n_\epsilon(c_\text{B}) = N-1$ and $n_\epsilon(c_\text{A}) = -N+1$. Note that $n_\epsilon(c') \in \mathcal{I}_c$: the opposite boundary is reached only after all $2N-2$ steps.

\begin{samepage}
\begin{definition}[Exact bridge trajectory]
\label{def:traj}
Given a source sample $\mbx_{\epsilon(c')} \in \gX_{c'}$, the \emph{exact bridge trajectory} $\mathbf{x} = (\mbx_{t_{-N+1}}, \dots, \mbx_{t_{N-1}})$ is obtained by integrating the empirical PF-ODE exactly from noise level $\sigma_{N-1}$ up to $\sigma_0$ under $c'$, then from $\sigma_0$ back down to $\sigma_{N-1}$ under $c$, and reading off the grid points. This is the DDIB trajectory of Eq.~(\ref{ddib}) restricted to the grid: the object the student is asked to collapse.
\end{definition}
\end{samepage}

We write $G(\cdot,\cdot;\phi,c)$ for the \emph{bridge consistency function} of direction $c$: the map sending any point of an exact bridge trajectory to that trajectory's endpoint in the target domain, so $G(\mbx_{t_n}, t_n; \phi, c) = \mbx_{\epsilon(c)}$ for every $n$. Lemma~\ref{lem:piecewise_consistency} of App.~\ref{sec:ibcd_theory} establishes that it is well defined and inherits the regularity \citet{song2023consistency} require, because no integration step straddles the junction index $i=0$ at which the domain embedding switches. Assumptions~\ref{ass:lip} and~\ref{ass:lte} are the IBCD counterparts of the two assumptions of Theorem~1 of \citet{song2023consistency}; Assumptions~\ref{ass:ema} and~\ref{ass:metric} are additional, and are what let us dispense with an exactly zero distillation loss.

\begin{samepage}
\begin{assumption}[Lipschitz student]
\label{ass:lip}
There is $L > 0$ with $\| G_\theta(\mbx, t, c) - G_\theta(\mathbf{y}, t, c) \|_2 \leq L \|\mbx - \mathbf{y}\|_2$ for all $\mbx, \mathbf{y}$, all non-zero grid times $t$, and both $c \in \{c_\text{\normalfont A}, c_\text{\normalfont B}\}$.
\end{assumption}
\end{samepage}

\begin{samepage}
\begin{assumption}[Solver order]
\label{ass:lte}
The solver call $\texttt{\normalfont Solve}(\cdot\,;\phi, \cdot, |t_n|, |t_{n+s(c)}|)$ has local error uniformly bounded by $C\,\big|\sigma_{|n|} - \sigma_{|n+s(c)|}\big|^{p+1}$ with $p \geq 1$ and $C$ independent of $n$. We write
$\Delta t := \max_{0 \leq k \leq N-2} (\sigma_{k} - \sigma_{k+1})$
for the largest \emph{noise-level} step of the grid.
\end{assumption}
\end{samepage}

\begin{samepage}
\begin{assumption}[Bounded EMA gap]
\label{ass:ema}
$\Delta_\text{\normalfont ema} := \sup_{\mbx, t, c} \| G_{\theta^-}(\mbx, t, c) - G_{\theta}(\mbx, t, c) \|_2 < \infty$, where $\theta^- = \texttt{\normalfont sg}(\mu\theta^- + (1-\mu)\theta)$ is the EMA/stop-gradient teacher of Eq.~(\ref{eq:ibcd}).
\end{assumption}
\end{samepage}

\goodbreak
\begin{samepage}
\begin{assumption}[Metric form of $d$]
\label{ass:metric}
$d(\mbx, \mathbf{y}) = \| \Phi(\mbx) - \Phi(\mathbf{y}) \|_2$ for some map $\Phi$, and Assumptions~\ref{ass:lip} and~\ref{ass:ema} hold with $G_\theta$ replaced by $\Phi \circ G_\theta$. Taking $\Phi = \mathrm{id}$ recovers $d = \ell_2$.
\end{assumption}
\end{samepage}

Assumption~\ref{ass:lte} is stated on noise levels because the solver is always called on \emph{magnitudes}. The junction step $n = -1 \rightarrow 0$ is therefore an ordinary step of length $\sigma_0 - \sigma_1$ rather than one of size $\mathcal{O}(\sigma_\text{max})$, which keeps $\Delta t$ comparable to that of ordinary CD (App.~\ref{sec:ibcd_theory}).

\begin{samepage}
\begin{definition}[Distillation difficulty]
\label{def:dd}
For an exact bridge trajectory $\mathbf{x}$, a target condition $c$, and an index $n \in \mathcal{I}_c \cup \{n_\epsilon(c)\}$, the \emph{distillation difficulty at $t_n$} is
\begin{equation}
\mathcal{D}_n(\mathbf{x}, c) := d\big( G_\theta(\mbx_{t_n}, t_n, c),\; \mbx_{\epsilon(c)} \big)
= d\big( G_\theta(\mbx_{t_n}, t_n, c),\; G(\mbx_{t_n}, t_n; \phi, c) \big).
\label{eq:dd_n}
\end{equation}
The \emph{distillation difficulty of the trajectory} is its value at the opposite boundary,
\begin{equation}
\mathcal{D}(\mathbf{x}, c) := \mathcal{D}_{n_\epsilon(c')}(\mathbf{x}, c) = d\big( G_\theta(\mbx_{\epsilon(c')}, \epsilon(c'), c),\; \mbx_{\epsilon(c)} \big),
\label{eq:dd}
\end{equation}
which is the definition used in Sec.~\ref{sec:ibcd2}. We also write $\mathcal{D}_{\max}(\mathbf{x},c) := \max_n \mathcal{D}_n(\mathbf{x}, c)$.
\end{definition}
\end{samepage}

\goodbreak
\begin{samepage}
\begin{definition}[One-step residual and its surrogate]
\label{def:dd_hat}
Write $\varsigma_c(n) := c_\text{\normalfont A}$ if $\min\{n,\, n{+}s(c)\} < 0$ and $\varsigma_c(n) := c_\text{\normalfont B}$ otherwise, for the domain embedding governing the half of the bridge that the step from $n$ to $n+s(c)$ lies in; the junction index $0$ is therefore taken under $c_\text{\normalfont A}$ when the step descends into $\gX_\text{\normalfont A}$ and under $c_\text{\normalfont B}$ when it descends into $\gX_\text{\normalfont B}$, as in Eqs.~(\ref{eq:forward}),~(\ref{eq:backward}). For $n \in \mathcal{I}_c$, let $\Hat{\mbx}_{t_{n+s(c)}} = \texttt{\normalfont Solve}(\mbx_{t_n}; \phi, \varsigma_c(n), |t_n|, |t_{n+s(c)}|)$ be the one-step target of those equations. The \emph{one-step consistency residual} is
\begin{equation}
\delta_n(\mathbf{x}, c) := d\big( G_\theta(\mbx_{t_n}, t_n, c),\; G_{\theta^-}(\Hat{\mbx}_{t_{n+s(c)}}, t_{n+s(c)}, c) \big),
\label{eq:residual}
\end{equation}
\ie exactly the per-sample integrand of $\mathcal{L}_\text{\normalfont IBCD}$ in Eq.~(\ref{eq:ibcd}), and the surrogate of Sec.~\ref{sec:ibcd2} is $\Hat{\mathcal{D}}(\mbx_{t_n}, c) = g\big(\delta_n(\mathbf{x}, c)\big)$ for a monotone increasing $g$. We write $\bar{\delta}(\mathbf{x},c) := \frac{1}{2N-2}\sum_{n \in \mathcal{I}_c} \delta_n(\mathbf{x},c)$ for the mean residual along the trajectory.
\end{definition}
\end{samepage}

\subsection{The Surrogate Controls the True Distillation Difficulty}
\label{supp:ddtheory_bound}

The main paper states Proposition~\ref{prop:dd_main} without its assumptions and without the partial converse Eq.~(\ref{eq:dd_lower}); both are supplied here.

\begin{samepage}
\begin{proprestate}
\label{prop:dd_bound}
Let Assumptions~\ref{ass:lip}--\ref{ass:metric} hold (Lipschitz student, order-$p$ solver, bounded EMA gap, and metric form of $d$). Then for every exact bridge trajectory $\mathbf{x}$ and every target condition $c \in \{c_\text{\normalfont A}, c_\text{\normalfont B}\}$, the distillation difficulty of Definition~\ref{def:dd} obeys
\begin{equation}
\mathcal{D}(\mathbf{x}, c) \;\leq\; \mathcal{D}_{\max}(\mathbf{x}, c) \;\leq\;
\underbrace{(2N-2)\,\bar{\delta}(\mathbf{x}, c)}_{\text{\normalfont accumulated residual}}
\;+\; (2N-2)\,\Delta_\text{\normalfont ema}
\;+\; \underbrace{2\,C L \,(\sigma_\text{\normalfont max} - \sigma_\text{\normalfont min})\,(\Delta t)^{p}}_{=:\;\mathcal{F},\;\text{\normalfont discretization floor}},
\label{eq:dd_upper}
\end{equation}
and conversely
\begin{equation}
\mathcal{D}_{\max}(\mathbf{x}, c) \;\geq\; \tfrac{1}{2}\Big( \max_{n \in \mathcal{I}_c} \delta_n(\mathbf{x}, c) \;-\; \Delta_\text{\normalfont ema} \;-\; C L (\Delta t)^{p+1} \Big).
\label{eq:dd_lower}
\end{equation}
\end{proprestate}
\end{samepage}

\begin{proof}
Fix $\mathbf{x}$ and $c$, write $s = s(c)$, and set $\mbe_n := \Phi\big(G_\theta(\mbx_{t_n}, t_n, c)\big) - \Phi\big(\mbx_{\epsilon(c)}\big)$, so that $\mathcal{D}_n = \|\mbe_n\|_2$ by Assumption~\ref{ass:metric} and Eq.~(\ref{eq:dd_n}).

\noindent\textbf{Base case.} At the target boundary the direction-dependent boundary condition of Sec.~\ref{sec:ibcd} gives $G_\theta(\mbx_{\epsilon(c)}, \epsilon(c), c) = \mbx_{\epsilon(c)}$, so $\mbe_{n_\epsilon(c)} = \mathbf{0}$.

\noindent\textbf{One-step recursion.} Let $n \in \mathcal{I}_c$. Since $\mbx_{t_n}$ and $\mbx_{t_{n+s}}$ lie on the same trajectory they share the target value $\mbx_{\epsilon(c)}$, so
\begin{align}
\mbe_{n} = \;&
\underbrace{\Big[\Phi\big(G_\theta(\mbx_{t_n}, t_n, c)\big) - \Phi\big(G_{\theta^-}(\Hat{\mbx}_{t_{n+s}}, t_{n+s}, c)\big)\Big]}_{\text{(i) consistency residual}}
+ \underbrace{\Big[\Phi\big(G_{\theta^-}(\Hat{\mbx}_{t_{n+s}}, t_{n+s}, c)\big) - \Phi\big(G_{\theta}(\Hat{\mbx}_{t_{n+s}}, t_{n+s}, c)\big)\Big]}_{\text{(ii) EMA gap}} \notag \\
& + \underbrace{\Big[\Phi\big(G_{\theta}(\Hat{\mbx}_{t_{n+s}}, t_{n+s}, c)\big) - \Phi\big(G_{\theta}({\mbx}_{t_{n+s}}, t_{n+s}, c)\big)\Big]}_{\text{(iii) solver error}}
\;+\; \mbe_{n+s}.
\label{eq:decomp}
\end{align}
The norm of (i) is $\delta_n(\mathbf{x},c)$ by Definition~\ref{def:dd_hat} and that of (ii) is at most $\Delta_\text{ema}$ by Assumption~\ref{ass:ema}; since $\mbx_{t_{n+s}}$ is by Definition~\ref{def:traj} the exact flow of $\mbx_{t_n}$ and $\Hat{\mbx}_{t_{n+s}}$ its one-step numerical approximation, Assumptions~\ref{ass:lip} and~\ref{ass:lte} bound (iii) by $L\|\Hat{\mbx}_{t_{n+s}} - {\mbx}_{t_{n+s}}\|_2 \leq C L \big|\sigma_{|n|} - \sigma_{|n+s|}\big|^{p+1}$. The triangle inequality then gives
\begin{equation}
\|\mbe_n\|_2 \;\leq\; \|\mbe_{n+s}\|_2 \;+\; \delta_n(\mathbf{x},c) \;+\; \Delta_\text{ema} \;+\; C L \big|\sigma_{|n|} - \sigma_{|n+s|}\big|^{p+1}.
\label{eq:recursion}
\end{equation}

\noindent\textbf{Induction.} Unrolling Eq.~(\ref{eq:recursion}) outward from $n_\epsilon(c)$ visits each index of $\mathcal{I}_c$ at most once, and all $2N-2$ of them when $n = n_\epsilon(c')$; since every term dropped is non-negative, for every $n$,
\begin{equation}
\|\mbe_{n}\|_2 \;\leq\; \sum_{m \in \mathcal{I}_c} \delta_m(\mathbf{x},c) \;+\; (2N-2)\,\Delta_\text{ema} \;+\; C L \sum_{m \in \mathcal{I}_c}\big|\sigma_{|m|} - \sigma_{|m+s|}\big|^{p+1}.
\end{equation}
In the last sum, bound $p$ of the $p+1$ factors by $\Delta t$ and accumulate the remaining one along the trajectory: $\sigma_{|m|}$ rises monotonically from $\sigma_{N-1}$ to $\sigma_0$ over the source half of $\mathcal{I}_c$ and falls back over the target half, so each half contributes $\sigma_0 - \sigma_{N-1}$ and
\begin{equation}
\sum_{m \in \mathcal{I}_c}\big|\sigma_{|m|} - \sigma_{|m+s|}\big| \;=\; 2\,(\sigma_\text{max} - \sigma_\text{min}) \;=:\; \Lambda .
\label{eq:total_variation}
\end{equation}
Hence $C L \sum_m |\sigma_{|m|} - \sigma_{|m+s|}|^{p+1} \leq C L \Lambda (\Delta t)^p = \mathcal{F}$. Taking the maximum over $n$ and substituting $\sum_m \delta_m = (2N-2)\bar\delta$ yields Eq.~(\ref{eq:dd_upper}); $\mathcal{D} \leq \mathcal{D}_{\max}$ is immediate from Definition~\ref{def:dd}.

\noindent\textbf{Converse.} Reading Eq.~(\ref{eq:decomp}) as an expression for term (i) and applying the triangle inequality in the other direction gives, for every $n \in \mathcal{I}_c$,
\begin{equation}
\delta_n(\mathbf{x},c) \;\leq\; \|\mbe_n\|_2 + \|\mbe_{n+s}\|_2 + \Delta_\text{ema} + C L (\Delta t)^{p+1} \;\leq\; 2\,\mathcal{D}_{\max}(\mathbf{x},c) + \Delta_\text{ema} + C L (\Delta t)^{p+1}.
\end{equation}
Taking the maximum over $n$ and rearranging gives Eq.~(\ref{eq:dd_lower}).
\end{proof}

\begin{samepage}
\begin{corollary}[Rate]
\label{cor:rate}
If the student is $\varepsilon$-optimal along the trajectory, in the sense that $\bar{\delta}(\mathbf{x},c) \leq \varepsilon$, then
$\mathcal{D}(\mathbf{x},c) \leq (2N-2)(\varepsilon + \Delta_\text{\normalfont ema}) + \mathcal{O}\big((\Delta t)^{p}\big)$.
In particular, if $\mathcal{L}_\text{\normalfont IBCD} = 0$ and $\theta^- = \theta$, then $\sup_{\mathbf{x}, c} \mathcal{D}(\mathbf{x},c) = \mathcal{O}\big((\Delta t)^{p}\big)$: the conclusion of Theorem~1 of \citet{song2023consistency}, transported to the signed-time bridge, with a constant twice as large.
\end{corollary}
\end{samepage}

\begin{samepage}
\begin{remark}[Where the doubled trajectory enters]
\label{rem:doubling}
The bridge is twice as long as a standard CD trajectory in two senses, which enter differently: the number of accumulation steps doubles, $N-1 \rightarrow 2N-2$, doubling the \emph{residual} and \emph{EMA} terms, and the total variation of the noise level doubles, $\sigma_\text{max}-\sigma_\text{min} \rightarrow \Lambda$, doubling the \emph{discretization floor}. Neither changes the exponent: the rate remains $\mathcal{O}((\Delta t)^p)$, so the bridge costs a factor of two in the constants and not an order.
\end{remark}
\end{samepage}

\begin{samepage}
\begin{remark}[Why the $\varepsilon$-version, and not the exactly-zero-loss version]
\label{rem:epsilon}
The cleaner form ``if $\mathcal{L}_\text{\normalfont IBCD} = 0$ then $\sup_{\mathbf{x}}|\Hat{\mathcal{D}} - \mathcal{D}| = \mathcal{O}((\Delta t)^p)$'' is true for $g = \mathrm{id}$ once $\theta^- = \theta$, as in Cor.~\ref{cor:rate}, and useless: at zero loss every residual vanishes, so $\Hat{\mathcal{D}} \equiv g(0)$ is constant across trajectories and the claim says nothing about whether $\Hat{\mathcal{D}}$ \emph{discriminates} hard trajectories from easy ones, the only property the adaptive weighting uses; it is also empty in the regime where the weighting is applied (training, with $\mathcal{L}_\text{\normalfont IBCD} > 0$), and fails outright for monotone $g$ with $g(0) \neq 0$. We therefore state the proposition for a non-zero residual, and Cor.~\ref{cor:rate} should be read as a consistency check on the constants rather than as the content of the result.
\end{remark}
\end{samepage}

\begin{samepage}
\begin{remark}[Direction of the guarantee]
\label{rem:direction}
Eq.~(\ref{eq:dd_upper}) is a one-sided guarantee on $\mathcal{D}$: a \emph{small} accumulated residual certifies a small distillation difficulty, but a large residual does not by itself prove $\mathcal{D}$ is large. Eq.~(\ref{eq:dd_lower}) repairs this only partially: a single large observed residual does certify that the consistency error is large \emph{somewhere} on the trajectory, \ie that $\mathcal{D}_{\max}$ is large, but it does not localize that error at the opposite boundary, so no matching lower bound on $\mathcal{D}$ itself follows and we claim none. The empirical counterpart of the missing direction is Fig.~\ref{fig:supp_ddw_vis_i2i}: among real Male$\rightarrow$Female trajectories, those with the largest $\mathbb{E}_t[\Hat{\mathcal{D}}]$ are visibly the ones the student translates worst.
\end{remark}
\end{samepage}

\begin{samepage}
\begin{remark}[Role of $g$]
\label{rem:g}
The bound of Sec.~\ref{supp:ddtheory_bound} aggregates residuals \emph{linearly}, so the transparent choice is $g = \mathrm{id}$, for which the weight is proportional to the certified bound on $\mathcal{D}$. Any strictly increasing $g$ preserves the ordering of trajectories by that bound, which is all the argument of Sec.~\ref{supp:ddtheory_why} needs, but changes the relative magnitudes: the concave choice used in our experiments (Tab.~\ref{table:hyperparam}) compresses the dynamic range and limits the influence of outlying trajectories, and $g \equiv 1$ recovers unweighted DMCD. The theory thus justifies the \emph{ranking} induced by $\Hat{\mathcal{D}}$, not a particular $g$; selecting among monotone $g$ remains an empirical matter, and is treated as such in Sec.~\ref{sec:ibcd2}.
\end{remark}
\end{samepage}

\begin{samepage}
\begin{remark}[What is estimated during training]
\label{rem:estimator}
The restated proposition concerns residuals along one fixed exact bridge trajectory. During training $\Hat{\mathcal{D}}$ is instead a single-sample estimate: one index $n$ is drawn uniformly and $\mbx_{t_n}$ by forward diffusion from a data point, so different draws lie on different trajectories through that point. The training-time weight therefore estimates the residual at the level of the marginal at $t_n$, and the per-trajectory $\bar{\delta}$ of Eq.~(\ref{eq:dd_upper}) is recovered only when the trajectory is held fixed and averaged over $n$, as in the diagnostic of Fig.~\ref{fig:supp_ddw_vis_i2i}. This is a real gap between the theory and the implementation: the implementation uses a cheaper, higher-variance estimator of the same quantity.
\end{remark}
\end{samepage}

\begin{samepage}
\begin{remark}[The distance function used in practice]
\label{rem:lpips}
Assumption~\ref{ass:metric} holds for $d = \ell_2$ with $\Phi = \mathrm{id}$, and also for $\sqrt{\text{LPIPS}}$, which is by construction the $\ell_2$ distance between concatenated, channel-normalized, per-layer-weighted deep features~\citep{zhang2018unreasonable}. Our implementation uses LPIPS itself, the square of such a distance, for which the triangle inequality does not hold and the restated proposition does not directly apply. Since $x \mapsto x^2$ is increasing on $\mathbb{R}_{\geq 0}$, the \emph{ordering} of trajectories, the property Sec.~\ref{supp:ddtheory_why} relies on, is unaffected, but the additive bound of Eq.~(\ref{eq:dd_upper}) should be read as a statement about the $\ell_2$ or $\sqrt{\text{LPIPS}}$ instantiation.
\end{remark}
\end{samepage}

\subsection{Why the Surrogate Is the Principled Weight for DMCD}
\label{supp:ddtheory_why}

The bound of Sec.~\ref{supp:ddtheory_bound} says on which trajectories DMCD's price in faithfulness (Fig.~\ref{fig:supp_abl}) is worth paying.

\textbf{(i) A small $\Hat{\mathcal{D}}$ is a certificate, so the weight should vanish there.} By Eq.~(\ref{eq:dd_upper}), a trajectory whose residuals are small has $\mathcal{D}$ within the floor $\mathcal{F}$ of zero: the student's single step already reproduces the teacher's $N$-step DDIB translation as closely as any consistency-based student can, so a distributional push there can only move the output away from a target it already matches.

\textbf{(ii) A large $\Hat{\mathcal{D}}$ marks where the objective has no control.} There Eq.~(\ref{eq:dd_upper}) degrades to a vacuous bound and Eq.~(\ref{eq:dd_lower}) certifies that the consistency error is in fact large somewhere along the bridge. The student's boundary value is then pinned down only to within a ball of radius $\sim (2N-2)\bar{\delta}$, and the consistency objective says nothing about which point of that ball to pick. A distribution-matching term is the only ingredient of Eq.~(\ref{eq:dmcd}) that can: it selects the point of high target density.

\textbf{(iii) The residual and the failure mode coincide.} The ball is not centred on a harmless point: App.~\ref{supp:meanpred} shows that the trajectories with large distillation error are those whose source samples sit near the source-domain decision boundary, and that the student's output on them collapses towards the mean of the two target modes, which in high dimension lies in a low-density region~\citep{blau2018perception}. Weighting DMCD by $\Hat{\mathcal{D}}$ therefore corrects density exactly where the failure is a density failure.

The argument establishes the right \emph{shape} of the weight. It does not show that DMCD is the optimal correction on the uncertified set, nor that a weight linear in $\Hat{\mathcal{D}}$ is optimal among monotone choices (Rem.~\ref{rem:g}).

\subsection{A Switch Criterion for the Staged Schedule}
\label{supp:ddtheory_switch}

The staged schedule of Sec.~\ref{sec:ibcd2} (train with $\mathcal{L}_\text{IBCD}$ alone, then introduce $\mathcal{L}_\text{DMCD}$ and $\mathcal{L}_\text{cycle}$) follows from the same decomposition. Eq.~(\ref{eq:dd_upper}) splits the certified distillation difficulty into a \emph{trainable} part, $(2N-2)(\bar{\delta} + \Delta_\text{ema})$, and an \emph{irreducible} floor $\mathcal{F}$ fixed by the solver order $p$, the grid, and the Lipschitz constants. While $(2N-2)\bar{\delta} \gg \mathcal{F}$, reducing $\mathcal{L}_\text{IBCD}$ reduces the guaranteed deviation from the teacher roughly proportionally, and adding DMCD would only leave $s_\psi$ chasing a student distribution far from converged. Once $(2N-2)\bar{\delta} \lesssim \mathcal{F}$ the floor dominates: further reduction of the residual cannot reduce the certified error, and whatever gap to the target distribution remains is exactly the residue the distribution-level and cycle terms exist to address. That is the moment to switch them on.

$C$, $L$ and hence $\mathcal{F}$ are not measurable, so the crossing point is not directly observable, but it is detectable as a \emph{plateau}, since $\bar{\delta}$ falls while the trainable term dominates and flattens once it reaches the floor. This yields a criterion stated entirely in quantities the training loop already computes.

\begin{samepage}
\begin{definition}[Residual-plateau switch criterion]
\label{def:switch}
Let $\widehat{\delta}_k$ denote the running mean of the one-step residual $\delta$ of Eq.~(\ref{eq:residual}) over the minibatches in a window of $W$ optimization steps ending at step $k$. Because our schedule uses $\lambda(t) \equiv 1$ (App.~\ref{supp:implementation}), $\widehat{\delta}_k$ is simply the running mean of $\mathcal{L}_\text{\normalfont IBCD}$ and requires no extra computation. Fix a relative-improvement tolerance $\tau > 0$ and introduce $\lambda_\text{\normalfont DMCD}$ and $\lambda_\text{\normalfont cycle}$ at
\begin{equation}
k^\star \;=\; \min\left\{ k \;\middle|\; \frac{\widehat{\delta}_{k-W} - \widehat{\delta}_{k}}{\widehat{\delta}_{k-W}} \;<\; \tau \right\}.
\label{eq:switch}
\end{equation}
\end{definition}
\end{samepage}

Here $W$ and $\tau$ are a smoothing window and a stopping tolerance, as in any plateau-based schedule. We state the criterion on $\delta$ rather than on $\Hat{\mathcal{D}} = g(\delta)$ deliberately: it is $\delta$ that Eq.~(\ref{eq:dd_upper}) aggregates linearly, and a running mean of $g(\delta)$ would differ from $g$ of the running mean for non-affine $g$.

\textbf{Relation to the schedule actually used.} Our reported models did not run Eq.~(\ref{eq:switch}) online; they used the fixed transition steps of Tab.~\ref{table:hyperparam}, set where the vanilla student's FID had stabilized (App.~\ref{supp:trainingcost_switch}). Def.~\ref{def:switch} is the criterion that fixed schedule \emph{instantiates}, not the procedure that produced it. We have not ablated one against the other, and we do not claim Eq.~(\ref{eq:switch}) would reproduce the reported transition steps exactly.

\begin{samepage}
\begin{remark}[Relation to other analyses of consistency models]
\label{rem:related_theory}
Theorem~1 of \citet{song2023consistency} bounds the consistency error under an exactly zero distillation loss; \citet{dou2024theory} relax the zero-loss requirement in a complementary direction, giving statistical rates in Wasserstein distance for the distribution produced by a consistency model. The bound restated in Sec.~\ref{supp:ddtheory_bound} is neither: it is a deterministic, per-trajectory bound whose purpose is not to certify the model globally but to relate two quantities computed on the \emph{same} trajectory, so that the cheaper can stand in for the expensive one inside a training weight. To our knowledge no comparable bound has been stated for a bridge-type consistency model: \citet{li2024bidirectional} and \citet{kim2023consistency} extend consistency distillation bidirectionally but do not analyze the resulting error accumulation.
\end{remark}
\end{samepage}

}        

\section{Model Parametrization} \label{supp:extend-edm}

The EDM~\citep{karras2022elucidating} parametrization for the student $G_\theta$ in consistency distillation~\citep{song2023consistency} is defined as follows for positive real-valued $t$ and the neural network $F_{\theta}$:
\begin{equation}
    G_{\theta}(\mathbf{x}_{t}, t) = c_{\text{skip}}(t) \mathbf{x}_{t} + c_{\text{out}}(t) F_{\theta}(c_{\text{in}}(t) \mathbf{x}_{t}, t^\prime(t)).
\end{equation}
In CD, authors choose
\begin{gather}
    c_{\text{skip}}(t) = \frac{\sigma_{\text{data}}^2}{(t-\epsilon)^2 + \sigma_{\text{data}}^2}, \qquad c_{\text{out}}(t) = \frac{\sigma_{\text{data}}(t-\epsilon)}{\sqrt{\sigma_{\text{data}}^2+t^2}}, \qquad c_{\text{in}}(t) = \frac{1}{\sqrt{ \sigma_{\text{data}}^2 + t^2}}, \\
    \qquad t^{\prime}(t) = 250 \cdot \ln (t + 10^{-44}) \notag
\end{gather}
to satisfy the boundary condition $G(\mbx_\epsilon, \epsilon)=\mbx_\epsilon$, and rescale the timestep.

For IBCD, we parametrize the student $G_\theta$ for non-zero real-valued $t$ and target domain condition $c$ as:
\begin{equation}
    G_\theta(\mbx_t, t, c) = c_{\text{skip}}(t, c) \mathbf{x}_{t} + c_{\text{out}}(t, c) F_{\theta}(c_{\text{in}}(t, c) \mathbf{x}_{t}, t^\prime(t)),
\end{equation}
which reflects the need for $c_{\text{skip}}$, $c_{\text{out}}$, and $c_{\text{in}}$ to depend on the target-domain condition $c$, ensuring that the appropriate boundary condition is applied at $t=\epsilon(c)$ for the selected target domain $c \in \{c_{\text{A}},c_{\text{B}}\}$.

\begin{figure}
    \centering
    \includegraphics[width=0.8\textwidth]{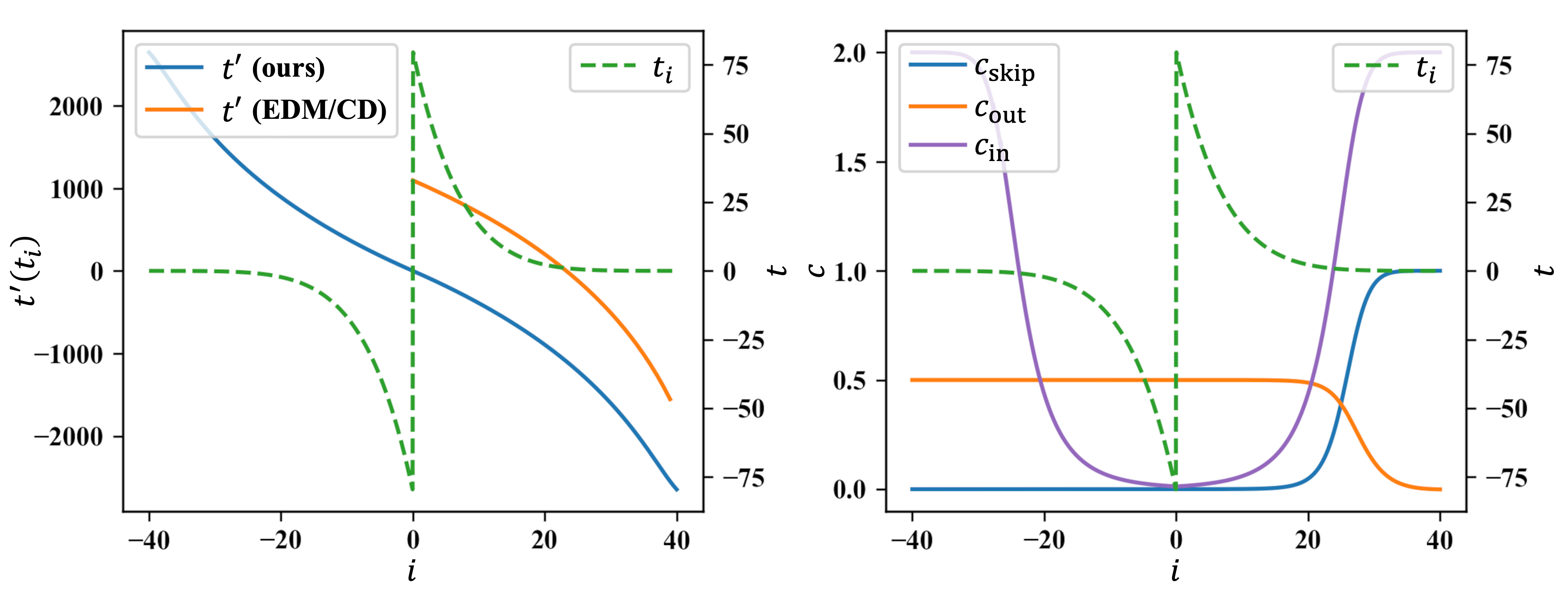}
    \caption{\textbf{Extension of EDM/CD model formulation for negative $t$ in IBCD student model.} $c_{\text{skip}}, c_{\text{out}},$ and $ c_{\text{in}}$ are shown for $c=c_\text{B}$ (the translation direction is $\gX_\text{A} \rightarrow \gX_\text{B}$).}
    \label{fig:supp-rescale-t}
\end{figure}

Although the student model is fully trained during the distillation process and does not theoretically need to be compatible with the teacher model, initializing it using the teacher model makes it advantageous to design the student to be as compatible as possible. We select $c_{\text{skip}}, c_{\text{out}},$ and $ c_{\text{in}}$ according to Eq.~(\ref{eq:new_c_skip}), (\ref{eq:new_c_out}), (\ref{eq:new_c_in}), ensuring continuity and compliance with the new boundary conditions while maintaining the definitions within the target domain regions ($t>0 \text{ for } c=c_\text{B}$, $t<0 \text{ for } c=c_\text{A}$).
\begin{equation} \label{eq:new_c_skip}
    c_{\text{skip}}(t, c) = 
    \begin{cases}
    \frac{1+\text{sign}(t)}{2} \frac{\sigma_{\text{data}}^2}{(t-\epsilon(c))^2 + \sigma_{\text{data}}^2} & \text{if } c=c_\text{B} \\
    \frac{1+\text{sign}(-t)}{2} \frac{\sigma_{\text{data}}^2}{(t-\epsilon(c))^2 + \sigma_{\text{data}}^2} & \text{if } c=c_\text{A}
    \end{cases}
\end{equation}
\begin{equation} \label{eq:new_c_out}
    c_{\text{out}}(t, c) = 
    \begin{cases}
    \frac{1+\text{sign}(t)}{2} \frac{\sigma_{\text{data}}(t-\epsilon(c))}{\sqrt{\sigma_{\text{data}}^2+t^2}} + \frac{1-\text{sign}(t)}{2} \sigma_{\text{data}} & \text{if } c=c_\text{B} \\
    - \frac{1+\text{sign}(-t)}{2} \frac{\sigma_{\text{data}}(t-\epsilon(c))}{\sqrt{\sigma_{\text{data}}^2+t^2}} + \frac{1-\text{sign}(-t)}{2} \sigma_{\text{data}} & \text{if } c=c_\text{A}
    \end{cases}
\end{equation}
\begin{equation} \label{eq:new_c_in}
   \qquad c_{\text{in}}(t, c) = \frac{1}{\sqrt{ \sigma_{\text{data}}^2 + t^2}}
\end{equation}
We also extend the timestep rescaler as Eq.~(\ref{eq:new_rescale_t}) to a symmetric and continuous form, ensuring shape compatibility with the original positive-bound domain. This symmetric design reflects the fact that the sign of the timestep separates the domains, while its absolute value represents the noise magnitude:
\begin{equation} \label{eq:new_rescale_t}
    t^{\prime}(t) = 250 \cdot \text{sign}(t) (\ln (|t| + 10^{-3}) - \ln (\sigma_{\text{max}}+ 10^{-44})).
\end{equation}
This approach preserves the structural integrity of the model and maintains consistent behavior across both domains. The parametrization extension of EDM/CD, as presented here, is visually illustrated in Fig.~\ref{fig:supp-rescale-t}.

\section{Theoretical Robustness of IBCD}
\label{sec:ibcd_theory}

A potential theoretical concern in IBCD is whether the \rev{switch} of the underlying PF-ODE vector field at the bridge center affects the convergence guarantees of consistency distillation. \rev{It is worth being precise about where that switch sits on the signed axis. Under the grid of Eq.~(\ref{eq:t_i}), $t_i = \text{sign}(i)\,\sigma_{|i|}$, the signed time is \emph{not} monotone in the index: $|t_i|$ is largest at $i=0$ and vanishes at the two ends $i=\pm N$. The bridge center (the shared Gaussian latent at which the governing domain embedding changes from $c_\text{A}$ to $c_\text{B}$) is therefore the \emph{index} $i=0$, at noise level $|t_0| = \sigma_\text{max}$, whereas $t=0$ is where the two \emph{clean-data} endpoints $\mbx_{t_{-N}}$ and $\mbx_{t_{N}}$ sit and no handover occurs at all. The lemma below is stated at the junction index.} In this section, we show that the assumptions underlying Theorem~1 of \citet{song2023consistency} remain satisfied under our signed-time bridge construction.

\begin{lemma}[Applicability of Consistency Distillation to IBCD]
\label{lem:piecewise_consistency}
The global error bound $O((\Delta t)^p)$ established in \citet{song2023consistency} remains applicable to the IBCD consistency mapping, provided that the signed-time discretization grid $\{t_i\}$ \rev{contains the junction index $i=0$, \ie the shared Gaussian latent at noise level $|t_0| = \sigma_\text{max}$ at which the governing domain embedding changes}.
\end{lemma}

\begin{proof}
\rev{We verify that the IBCD construction preserves the key assumptions in Theorem~1 of \citet{song2023consistency}: the Lipschitz continuity of the target consistency mapping and the student model, and the local truncation
error (LTE) order of the numerical solver.}

\noindent \textbf{Lipschitz Regularity of the Target Consistency Mapping.} Theorem~1 of \citet{song2023consistency} assumes that the ideal consistency mapping being approximated is Lipschitz continuous in $\mathbf{x}$ for each timestep. In IBCD, the target mapping is defined piecewise over the signed-time axis: \rev{on the indices $i<0$ (\ie $t_i<0$)}, it coincides with the standard diffusion consistency mapping associated with domain $\gX_\text{A}$, and \rev{on the indices $i \geq 0$ (\ie $t_i>0$)}, with that of domain $\gX_\text{B}$. Since the consistency mapping associated with each domain satisfies the regularity conditions of \citet{song2023consistency} under standard noise schedules, and the respective Lipschitz constants are uniformly bounded, the target mapping remains Lipschitz continuous in $\mathbf{x}$ at every discretized
timestep on the signed-time axis.

\noindent \textbf{Lipschitz Continuity of the Student Mapping.} The framework further requires the student model $G_\theta(\cdot,t)$ to belong to a Lipschitz function class in $\mathbf{x}$. IBCD adopts the standard architectural and regularization choices of consistency models, which are applied uniformly across the signed-time axis. Consequently, the student mapping $G_\theta(\cdot,t)$ satisfies the required Lipschitz condition on both sub-intervals $t<0$ and $t>0$, as in the original formulation.

\noindent \textbf{Local Truncation Error at the Junction.} \rev{The IBCD discretization places a grid point at the junction index $i=0$, and the solver is called only on noise-level \emph{magnitudes}, $\texttt{Solve}(\cdot\,;\phi,c,|t|,|t'|)$ (Eqs.~(\ref{eq:forward}),~(\ref{eq:backward})). Consequently every integration step is either an $\gX_\text{A}$-step among the indices $\{-N{+}1,\dots,0\}$, taken under the embedding $c_\text{A}$, or an $\gX_\text{B}$-step among $\{0,\dots,N{-}1\}$, taken under $c_\text{B}$; the junction index is always an \emph{endpoint} of such a step and never an interior point of one. In particular the step $i = -1 \rightarrow 0$ runs from noise level $\sigma_1$ to $\sigma_0 = \sigma_\text{max}$ entirely under $c_\text{A}$, and the step $i = 0 \rightarrow 1$ runs from $\sigma_0$ back to $\sigma_1$ entirely under $c_\text{B}$. No solver step therefore straddles the change of the governing vector field, and each step operates within a single smooth PF-ODE. Because the step size that enters the local error is the change in \emph{noise level} and not the jump in signed time, the junction step is an ordinary grid step of length $\sigma_0-\sigma_1$ rather than a step of size $\mathcal{O}(\sigma_\text{max})$.} As a result, the solver maintains the original $O((\Delta t)^{p+1})$ LTE order, which preserves the cumulative $O((\Delta t)^p)$ global error bound.

\noindent \textbf{Conclusion.} As both the regularity of the target consistency mapping and the solver error order required by Theorem~1 of  \citet{song2023consistency} are preserved under the proposed discretization, the consistency distillation argument applies to IBCD \rev{over the whole signed grid $\{t_i\}_{i=-N+1}^{N-1}$, \ie over both halves of the bridge,} without introducing additional theoretical assumptions. \rev{App.~\ref{supp:ddtheory} takes this further and turns the same decomposition into a quantitative bound on the distillation difficulty.}
\end{proof}

\section{Implementation Details} \label{supp:implementation}

\textbf{Model Architectures.} All models used in this study (the teacher $\phi$, the student $\theta$ and the fake DM $\psi$) employed the same model architecture as in EDM/CD~\citep{karras2022elucidating, song2023consistency}. The architecture configuration followed that of the LSUN-256 teacher EDM model introduced by~\citet{song2023consistency}. However, the student model was further modified with the model parametrization described in Appendix~\ref{supp:extend-edm}, while the teacher and fake DM maintained the original EDM parametrization.

\textbf{Teacher Model Training.} The teacher model was trained using the EDM implementation and the LSUN-256 model training configuration provided by \citet{song2023consistency}. The training setup included a log-normal schedule sampler and L2 loss, with a global batch size of 288, a learning rate of 1e-4, a dropout rate of 0.1, and an exponential moving average (EMA) of 0.9999. Mixed precision training was enabled, and weight decay was not applied. The teacher model was trained with class conditions on two types of AFHQ-256 models (cat, dog, and wild) and CelebA-HQ-256 models (female and male). The AFHQ and CelebA-HQ models were trained using their respective training sets from the AFHQ~\citep{choi2020stargan} and CelebA-HQ~\citep{karras2017progressive} datasets. Each model was trained for approximately 5 days, completing 800K steps on an NVIDIA A100 40GB eight-GPU setup.

\textbf{Implicit Bridge Consistency Distillation.} The discretization of DDIB trajectories is defined by extending the sampling discretization of EDM to satisfy Eq.~(\ref{eq:t_i}):
\begin{equation}
    t_i = \rev{\text{sign}(i)\,\sigma_{|i|}} =
    \begin{cases}
        \text{sign}(i)(\sigma_\text{max}^{1/\rho} + \frac{|i|}{N-1}(\sigma_\text{min}^{1/\rho} - \sigma_\text{max}^{1/\rho}))^{\rho} \quad & (\rev{-N<i<N}) \\
        0 \quad & (i = \pm N)
    \end{cases}
\end{equation}
\begin{equation*}
    \text{where} \quad 
    \text{sign}(x) =
    \begin{cases}
        +1 &\; (x \geq 0) \\
        -1 &\; (x < 0)
    \end{cases}
    , \sigma_{\text{min}}=0.002, \; \sigma_{\text{max}}=80, \; \sigma_{\text{data}}=0.5,\; N=40, \; \rho=7.0.
\end{equation*}

For the distance function $d$ in each loss, $d_\text{IBCD}$ and $d_\text{DMCD}$ were based on LPIPS~\citep{zhang2018unreasonable}, while $d_\text{cycle}$ used the L1 loss. The EMA parameter of the EMA model $\theta^-$ was 0.95, and an additional EMA with a separate parameter 0.9999432189950708 was applied to the student model $\theta$ and used during inference. The global batch size was 256, with the student learning rate of 4e-5 and the fake DM learning rate of  1e-4. Dropout and weight decay were not used, and mixed precision learning was employed. 

The ODE solver used was the 2nd order Heun solver~\citep{ascher1998computer}, consistent with EDM/CD. The weight scheduler for the IBCD loss employed $\lambda(t)=1$, while the  DMCD loss used the weight scheduler $w_{t}$ as suggested in \citet{yin2024one}. For the three tasks, Cat$\leftrightarrow$Dog, Wild$\leftrightarrow$Dog models were distilled using the AFHQ teacher model and its corresponding training dataset. The Male$\leftrightarrow$Female models were distilled using the CelebA-HQ teacher model and its training dataset. 

The distillation process began with only the IBCD loss and transitioned to using the full loss set once the FID~\citep{heusel2017gans} evaluation metrics stabilized (\textit{i.e.} transition step). Distillation was conducted on the same eight NVIDIA A100 40GB GPUs used to train the teacher model. Additional hyperparameters for each model and configuration are detailed in Tab.~\ref{table:hyperparam}.

\textbf{Evaluation.} We followed the evaluation methodology and tasks outlined in EGSDE~\citep{zhao2022egsde}. The publicly available evaluation code\footnote{https://github.com/ML-GSAI/EGSDE} was used without modification. Validation sets from the AFHQ and CelebA-HQ datasets were used as the evaluation datasets. All images in each validation set were translated using the respective task-specific models. For each image pair (source domain and translated target domain), PSNR, SSIM~\citep{wang2004image}, LPIPS~\citep{zhang2018unreasonable}, and CLIPScore~\citep{hessel2021clipscore} were computed, and the average values across all pairs were reported.

FID~\citep{heusel2017gans} was calculated using the \texttt{pytorch-fid}\footnote{https://github.com/mseitzer/pytorch-fid} library to measure the distance between the real target domain image distribution and the translated target image distribution. Following the methodology of \citet{choi2020stargan} and \citet{zhao2022egsde}, images from the CelebA-HQ dataset were resized and normalized before FID calculation, while images for other tasks were evaluated without additional preprocessing. L2 distance measurement was not included in this evaluation.

Density-coverage~\citep{naeem2020reliable} was computed using \texttt{prdc-cli}\footnote{https://github.com/Mahmood-Hussain/generative-evaluation-prdc} between the distribution of real target domain images and the distribution of images translated into the target domain, similar to the FID measurement. The measurement mode was \texttt{T4096} (features of the \texttt{fc2} layer of the ImageNet pre-trained VGG16~\citep{simonyan2014very} model). The metric was computed for the entire dataset at once, without using mini-batches. Unlike FID, no specific transformation was applied for the CelebA-HQ dataset.

In the user study, twenty participants were recruited to perform a blinded, pairwise comparison between our method and each baseline. For each model and task, two identical source images were randomly selected, and their corresponding outputs, generated by each model, were presented side by side for evaluation. Participants were asked to make three separate selections for each comparison: one based on realism, one based on fidelity to the source image, and one based on overall preference. \rev{Results are reported as the average ratio of participants who preferred each baseline over our model for each criterion, averaged across all tasks; values below 0.5 therefore indicate that our model was preferred (Tab.~\ref{table:userstudy}).}

\textbf{Baselines.} As baselines, we compare our method against several GAN-based methods, including CycleGAN~\citep{zhu2017unpaired}, Self-Distance~\citep{benaim2017one}, GcGAN~\citep{fu2019geometry}, LeSeSIM~\citep{zheng2021spatially}, StarGAN v2~\citep{choi2020stargan}, and CUT~\citep{park2020contrastive}. We also benchmark against diffusion model (DM)-based methods such as ILVR~\citep{choi2021ilvr}, SDEdit~\citep{meng2022sdedit}, EGSDE~\citep{zhao2022egsde}, CycleDiffusion~\citep{wu2023latent}, and SDDM~\citep{sun2023sddm}. Furthermore, we evaluate our approach against Optimal Transport (OT)-based methods, including NOT~\cite{korotin2022neural}, Eg-EOT~\cite{mokrov2024energyguided}, and DIOTM~\cite{choi2025improving}. We also compare it with Schrödinger Bridge (SB)-based methods such as DSBM~\citep{shi2023diffusion}, UNSB~\citep{kim2023unpaired}, and ASBM~\cite{gushchin2024adversarial}, as well as the teacher model, DDIB~\citep{su2022dual}. Most of the comparison results are sourced from \citet{zhao2022egsde} except for the density-coverage, while the results for NOT, DSBM, UNSB, ASBM, Eg-EOT, DIOTM, and DDIB are based on our re-implementations. Finally, we also include results from GPT-Image-1~\citep{openai_image_generation_api_2025}, a multi-turn image generation and editing foundation model powered by GPT-4.1~\citep{openai_gpt4_1_2025}, OpenAI’s multimodal large language model.

\textbf{Reproductions.}
To evaluate our method, we replicated NOT, DSBM, UNSB, ASBM, Eg-EOT, DIOTM, and DDIB, which have not been previously evaluated on our benchmark datasets. 
For UNSB\footnote{https://github.com/cyclomon/UNSB}, we used the default configuration for the Horse$\rightarrow$Zebra task and trained the model for 400 epochs. Sampling was performed with 5 steps. 
For NOT\footnote{https://github.com/iamalexkorotin/NeuralOptimalTransport}, we used the default configuration for the Handbag$\rightarrow$Shoes task with strong costs. 
For DSBM\footnote{https://github.com/yuyang-shi/dsbm-pytorch}, we adopted the configuration for the Wild$\rightarrow$Cat task in DSBM-IMF mode, except that the number of training steps was set to 10K. 
For ASBM\footnote{https://github.com/Daniil-Selikhanovych/ASBM}, we used the default configuration for CelebA translation. Specifically, for one-sided pretraining, we used a batch size of 8 and trained for 120 epochs, while for D-IMF training, we used a batch size of 4. 
For Eg-EOT\footnote{https://github.com/PetrMokrov/Energy-guided-Entropic-OT}, we first trained a StyleGAN2-ADA\footnote{https://github.com/NVlabs/stylegan2-ada-pytorch} model on AFHQ-dog using the paper256 configuration. The main model was then trained using the default configuration for the Cat$\rightarrow$Dog task. 
For DIOTM\footnote{https://openreview.net/attachment?id=CfZPzH7ftt\&name=supplementary\_material}, we used the configuration for the Male$\rightarrow$Female-128 task, except that the batch size was set to 32. 
For DDIB, we implemented the method within our framework. Specifically, DDIB was executed by first solving the ODE backward from the source domain and then solving it forward to the target domain using the EDM model trained for IBCD. The ODE solver was implemented in the same manner as the EDM sampler, utilizing the same sampling hyperparameters as those used for EDM/IBCD. This setup ensured consistency in the evaluation and enabled a direct comparison across methods.

We also re-sampled the result from models (CUT, ILVR, SDEdit, EGSDE, CycleDiffusion) for which the density-coverage~\citep{naeem2020reliable}, LPIPS~\citep{zhang2018unreasonable} and CLIPScore~\citep{hessel2021clipscore}, and user study metrics were not originally reported. These metrics were measured for these models using the method described above, and we included results in Tab.~\ref{table:main} and Tab.~\ref{table:userstudy}. The target models for this evaluation were limited to baseline models that met the following criteria: 1) Open-source code and checkpoints were available. 2) FID, PSNR, and SSIM values reported by the authors could be reproduced using the reported sampling strategy. This ensured that all metrics in Tab.~\ref{table:main} and Tab.~\ref{table:userstudy} were measured on consistent samples.

The results for GPT-Image-1 were obtained by providing an input image to the \texttt{gpt-4.1-2025-04-14} model via the OpenAI API, instructing it to transform the image into the target domain using the \texttt{gpt-image-1-2025-04-23} tool. The input image was resized to 256×256 pixels, consistent with all other experiments, and the output image (originally 1024×1024) was downsampled to 256×256. The prompts used for generation are provided in Tab.~\ref{table:gpt-prompt}. This experiment incurred a total cost of approximately \$100.

\begin{table}

\centering
\small
\caption{\textbf{Specific hyperparameters employed by different models and configurations.}}
\label{table:hyperparam}
\resizebox{0.6\textwidth}{!}{
\begin{tabular}{ccccccc}
\toprule

\textbf{Model} & \multicolumn{2}{c}{\textbf{Cat$\leftrightarrow$Dog}} & \multicolumn{2}{c}{\textbf{Wild$\leftrightarrow$Dog}} & \multicolumn{2}{c}{\textbf{Male$\leftrightarrow$Female}} \\
Configuration & IBCD          & IBCD$^\dagger$          & IBCD          & IBCD$^\dagger$          & IBCD          & IBCD$^\dagger$          \\
\midrule
$\lambda_\text{DMCD}$   &  1     & 0.18   &    0.2  &   0.2     &       0.02     &       0.02      \\
$\lambda_\text{cycle}$  &  0.03  & 0.003  &   0.001       &   0.0003   &         0.00001    &     0.00003        \\
$g(\cdot)$              &  1     & \multicolumn{5}{c}{$\min(0, \log(\cdot)+10)$} \\
transition step         &  200K  & 200K &   200K & 200K  &       500K & 500K  \\
total distillation step &  210K  & 230K  &  210K &   230K  &   510K   &    520K     \\
\bottomrule
\end{tabular}
}
\end{table}
\begin{table}[!t]
\caption{\textbf{Prompts used in the GPT-Image-1 translation experiment for each task.} These prompts were provided to the \texttt{gpt-4.1-2025-04-14} model via the OpenAI API using the \texttt{gpt-image-1-2025-04-23} tool. Each prompt was designed to preserve pose, lighting, background, and composition while achieving realistic domain translation.}
\centering
\label{table:gpt-prompt}
\setlength{\tabcolsep}{0.5em}
\resizebox{0.95\textwidth}{!}{
\begin{tabular}{@{}lp{0.95\linewidth}@{}}
\toprule
\textbf{Task} & \textbf{Prompt} \\
\midrule
Cat$\rightarrow$Dog & Transform the image of the cat into a realistic dog, preserving the same pose, lighting, background, and overall composition. Ensure the dog appears natural and lifelike, matching the fur color, orientation, and proportions of the original cat as closely as possible. The final image should look photorealistic and faithful to the original scene, as if the dog were actually photographed in place of the cat. \\
\midrule
Wild$\rightarrow$Dog & Transform the image of the wild animal into a realistic dog, preserving the same pose, lighting, background, and overall composition. Ensure the dog appears natural and lifelike, matching the fur color, orientation, and proportions of the original animal as closely as possible. The final image should look photorealistic and faithful to the original scene, as if the dog were actually photographed in place of the wild animal. \\
\midrule
Male$\rightarrow$Female & Transform the image of the man into a realistic woman, preserving the same pose, lighting, background, and overall composition. Ensure the woman appears natural and lifelike, matching the skin tone, orientation, and proportions of the original man as closely as possible. The final image should look photorealistic and faithful to the original scene, as if the woman were actually photographed in place of the man. \\
\bottomrule
\end{tabular}
}
\end{table}

{\addc   
\section{\texorpdfstring{\rev{Training Cost, Memory Footprint, and the Staged Schedule}}{Training Cost, Memory Footprint, and the Staged Schedule}}
\label{supp:trainingcost}

The final IBCD objective combines four loss terms and, in its complete form, holds a frozen teacher, a student, two EMA copies of the student and an auxiliary fake score network in memory at once. This section accounts for that cost and gives a reproducible criterion for the staged schedule, together with the training budget in optimization steps and wall-clock time. Two facts frame the accounting. The auxiliary network and the auxiliary losses are active only \emph{after} the transition step, and that phase is a short tail of the schedule, 10K--30K steps out of a 210K--520K-step run (Tab.~\ref{table:hyperparam}), so for the great majority of training IBCD is plain consistency distillation against a frozen teacher, and costs accordingly. And this is a one-off training cost that buys the inference cost of Tab.~\ref{table:supp-inference-time}: a single 32.95M-parameter network at 1 NFE, against the 160 NFE of the DDIB teacher it is distilled from.

\subsection{What Is Resident During Training}
\label{supp:trainingcost_components}

All networks in IBCD share the EDM/CD architecture (App.~\ref{supp:implementation}), differing only in the student's reparametrization (App.~\ref{supp:extend-edm}), which changes the preconditioning coefficients and not the network; each therefore carries the 32.95M parameters reported in Tab.~\ref{table:supp-inference-time}. Tab.~\ref{table:supp-train-components} lists what is resident, what receives gradients and optimizer state, and which stage each component is needed in. Only the student $G_\theta$ and the fake score network $s_\psi$ are trained; the teacher $s_\phi$ is frozen throughout, and the two EMA copies are never differentiated. Two features of the construction bear on the total: a \emph{single} class-conditional teacher serves both domains, so one frozen teacher is resident where a literal DDIB setup would hold two; and $s_\psi$, instantiated at the transition step, is what turns a one-network training loop into a two-network one.

\textbf{Passes per optimization step.} A Stage~1 iteration (Algo.~\ref{alg:ibcd-vanilla}) is one Heun solver step of the frozen teacher per minibatch half (two function evaluations, taken without gradients~\citep{ascher1998computer}), one student forward with gradients, one $\theta^-$ forward without, and one backward through the student. Stage~2 (Algo.~\ref{alg:ibcd-final}) adds an $s_\psi$ and an $s_\phi$ forward at the re-noised student output for Eq.~(\ref{eq:dmcd}); two \emph{chained} student forwards with gradients for $\gL_\text{cycle}$, the deepest activation chain in the objective since backpropagation traverses the student twice; and one forward and one backward of $s_\psi$ for its DSM update. DMCD itself needs no extra student forward, reusing the output already computed for $\gL_\text{IBCD}$. A Stage~2 step is therefore substantially more expensive than a Stage~1 step in both compute and activation memory, which is precisely why the schedule confines Stage~2 to a short tail (Sec.~\ref{supp:trainingcost_schedule}).

\begin{table}[ht]
\addc
\centering
\small
\caption{\addc \textbf{Components held in memory during distillation, and the phase each is needed in.} All five networks share the EDM/CD architecture, so each carries the 32.95M parameters reported for IBCD in Tab.~\ref{table:supp-inference-time}. Only $\theta$ and $\psi$ receive gradients and carry optimizer state; $\theta^-$ and $\theta_\text{EMA}$ are exponential moving averages of $\theta$ and are never differentiated; $\phi$ is frozen for the whole run. Stage~1 is the phase before the transition step, Stage~2 the phase after it.}
\label{table:supp-train-components}
\resizebox{0.95\textwidth}{!}{
\begin{tabular}{llcccccc}
\toprule
\textbf{Component} & \textbf{Symbol} & \textbf{Params [M]} & \textbf{Role} & \textbf{Gradients} & \textbf{Update rule} & \textbf{Stage 1} & \textbf{Stage 2} \\
\midrule
Teacher (EDM)        & $s_\phi$              & 32.95 & PF-ODE targets, real score & \xmark & frozen & \cmark & \cmark \\
Student              & $G_\theta$            & 32.95 & the deployed translator & \cmark & gradient step on $\gL_\text{total}$, $\zeta_\theta = $ 4e-5 & \cmark & \cmark \\
CD target            & $G_{\theta^-}$        & 32.95 & target branch of $\gL_\text{IBCD}$ & \xmark & EMA of $\theta$, $\mu = 0.95$ & \cmark & \cmark \\
Inference copy       & $G_{\theta_\text{EMA}}$ & 32.95 & evaluation only & \xmark & EMA of $\theta$, $\mu = 0.9999432\ldots$ & \cmark & \cmark \\
Fake score network   & $s_\psi$              & 32.95 & student-distribution score in $\gL_\text{DMCD}$ & \cmark & DSM on student outputs, $\zeta_\psi = $ 1e-4 & \xmark & \cmark \\
\midrule
\multicolumn{2}{l}{\textit{Weight copies resident}} & \multicolumn{6}{l}{\textit{4 copies (131.80M parameters) in Stage 1; 5 copies (164.75M) in Stage 2}} \\
\multicolumn{2}{l}{\textit{Networks with optimizer state}} & \multicolumn{6}{l}{\textit{1 ($\theta$) in Stage 1; 2 ($\theta$, $\psi$) in Stage 2}} \\
\bottomrule
\end{tabular}
}
\end{table}

\subsection{The Two-Phase Schedule, With Numbers}
\label{supp:trainingcost_schedule}

Phase~1 optimizes $\gL_\text{IBCD}$ alone, with $\lambda_\text{DMCD} = \lambda_\text{cycle} = 0$ and no fake score network; at the \emph{transition step} both auxiliary losses are switched on at the weights of Tab.~\ref{table:hyperparam} and $s_\psi$ begins training. Taking transition steps and totals from that table, Phase~2 is between 2.0\% and 13.0\% of all optimization steps across the six reported configurations. For the remaining 87--98\% the run holds four weight copies, one of them trained, and evaluates the teacher only as a solver; the five-network, two-optimizer configuration of Tab.~\ref{table:supp-train-components} exists only in the short tail. Fig.~\ref{fig:supp_wandb} shows this directly: both auxiliary losses are identically zero until the transition.

The shape is deliberate. $\gL_\text{DMCD}$ measures the student's output distribution through a fake score network that must itself be fitted to the student's outputs, so running it against a student far from converged means fitting $s_\psi$ to a rapidly moving target. Deferring it turns the expensive part of the objective into a short fine-tuning phase on top of a cheap and stable one, which is also why the training cost is dominated by Phase~1.

\subsection{The Switch Criterion}
\label{supp:trainingcost_switch}

Describing the switch only as occurring ``once the FID evaluation metrics stabilized'' is not a criterion a reader can reproduce. App.~\ref{supp:ddtheory_switch} supplies one: the bound of App.~\ref{supp:ddtheory} splits the certified distillation difficulty into a trainable part, driven by the accumulated one-step consistency residual, and an irreducible discretization floor. While the trainable part dominates, further consistency distillation still reduces the guaranteed deviation from the teacher; once it reaches the floor it no longer can, and the remaining gap is exactly the residue the distribution-level and cycle terms exist to address. Def.~\ref{def:switch} turns that crossing into a plateau test on the running mean of the one-step residual, which, since our schedule uses $\lambda(t) \equiv 1$, is the running mean of $\gL_\text{IBCD}$ and is already logged at every step.

What the reported runs used is a fixed transition step per model, 200K for Cat$\leftrightarrow$Dog and Wild$\leftrightarrow$Dog and 500K for Male$\leftrightarrow$Female (Tab.~\ref{table:hyperparam}), set in advance at the step where periodic FID evaluation of the vanilla-IBCD student had stopped improving. Def.~\ref{def:switch} is the principle those schedules instantiate: the same plateau test, read off the residual the bound is written in rather than off a downstream FID evaluation, and cheaper, since $\widehat{\delta}$ is logged at every step while FID requires periodic sampling. Stating it in that form is what makes the schedule reproducible; running it online, and ablating it against the fixed schedule, is the natural next step.

\subsection{Training Budget and Memory Footprint}
\label{supp:trainingcost_wallclock}

Every run used the same eight-GPU NVIDIA A100 40GB node, at a global batch size of 256 for distillation and 288 for the teacher, with mixed precision, no weight decay and no dropout (App.~\ref{supp:implementation}). The teacher was trained on that node for approximately 5 days, completing 800K steps. The Cat$\leftrightarrow$Dog student was then distilled from it in a further 3 days, completing the 210K steps of Tab.~\ref{table:hyperparam}; the other five configurations in that table run 210K--520K steps on the same node. Because one class-conditional teacher serves both domains and both directions, the teacher cost is paid once per dataset pair and is also the cost of the DDIB teacher baseline itself, not an overhead specific to IBCD.

Teacher and student together therefore amount to 1.01M optimization steps for Cat$\leftrightarrow$Dog, comparable to the 1.0M steps of the 93.56M-parameter DDPM backbone on which ILVR, SDEdit, EGSDE and CycleDiffusion are built (Tab.~\ref{table:supp-teacher-backbone}), and IBCD reaches it with roughly a third of the parameters, 32.95M against 93.56M. What that comparable budget buys is a single-NFE translator: 0.012\,s per image, against relative inference costs of over $1000\times$ for ILVR, EGSDE and CycleDiffusion, and roughly $120\times$ for the 160-NFE DDIB teacher it is distilled from (Tab.~\ref{table:supp-inference-time}). The one-off training cost is of the same order as the baselines'; the recurring inference cost is three orders of magnitude smaller.

The memory footprint follows from the residency accounting of Tab.~\ref{table:supp-train-components}. Stage~1 holds four weight copies, 131.80M parameters in total, with a single optimizer state; Stage~2 holds five copies, 164.75M parameters, with two. Since Stage~2 occupies 2.0--13.0\% of the schedule (Sec.~\ref{supp:trainingcost_schedule}), the lighter four-network, one-optimizer configuration is what the run holds for 87--98\% of its life, and the heavier configuration fits on the same node at the same batch size throughout.
}        

\section{Further Experimental Results} \label{supp:further_exp}

\subsection{{Distillation Error} in Vanilla IBCD}
\label{supp:meanpred}

\begin{figure}[ht!]
    \centering
    \includegraphics[width=0.9\textwidth]{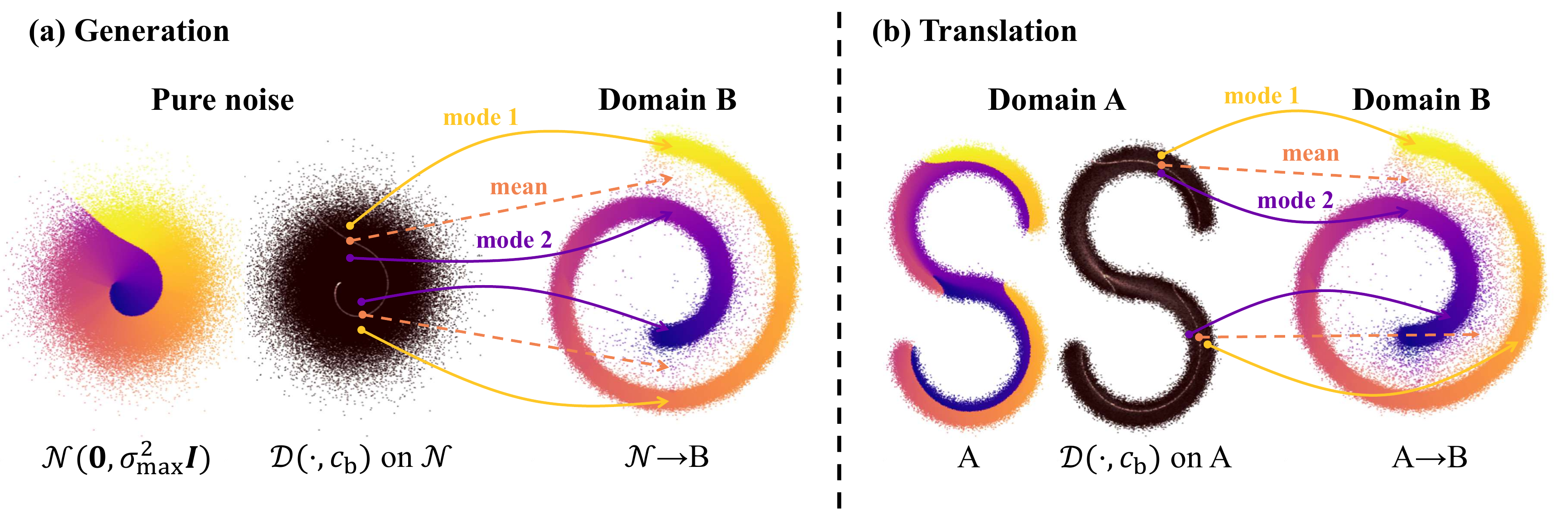}
    \caption{\textbf{Incorrect mapping to low-density regions due to distillation error.} (a) Generation and (b) translation results using vanilla IBCD. Samples with high distillation difficulty (\textit{i.e.}, large distillation errors), which originate near the source domain decision boundary, tend to be mapped to low-probability regions in the target domain.}
    \label{fig:supp_meanpred}
\end{figure}
Fig.~\ref{fig:supp_meanpred} illustrates the {distillation error} that arises when using only vanilla IBCD loss on the synthetic toy dataset. When generating samples from pure noise to domain $B$ (Fig.~\ref{fig:supp_meanpred}~(a)) or translating samples from domain $A$ to domain $B$ (Fig.~\ref{fig:supp_meanpred}~(b)) using only IBCD loss, the translated results often fall in the low-density region of the target distribution. These translated points primarily originate from the source domain decision boundary, which is the boundary separating the partition in the source domain that should be mapped to two different target domain modes. {Translation errors are more pronounced in longer neural jump paths, such as those involved in translations ($i=-N+1\to N-1$), compared to shorter paths in generation ($i=0\to N-1$).}

{\addc   
\subsection{\texorpdfstring{\rev{Topology of the Toy Manifolds and What a One-Step Map Can Represent}}{Topology of the Toy Manifolds and What a One-Step Map Can Represent}}
\label{supp:toy_topology}

Both toy distributions (the S-curve for $\gX_\text{A}$ and the Swiss roll for $\gX_\text{B}$) are noisy one-dimensional manifolds in the plane, and IBCD's translation in Fig.~\ref{fig:result_toy}(a) is governed by flat two-dimensional proximity rather than by position \emph{along} them. That reading is correct, and it describes from the outside what App.~\ref{supp:meanpred} describes from the inside.

\textbf{The observation and the decision boundary are the same set.} A translation respecting both topologies would be an arc-length correspondence between two curves whose windings differ, and any such map must somewhere either separate a region of $\gX_\text{A}$ that the ambient metric keeps connected or join two that it keeps apart. That locus is exactly what Sec.~\ref{sec:ibcd2} calls a \emph{decision boundary} of the source domain: the partition whose two sides the teacher's PF-ODE sends to different modes of $\gX_\text{B}$. Fig.~\ref{fig:result_toy}(b) shows the distillation difficulty concentrating precisely there, and App.~\ref{supp:meanpred} shows that the vanilla student's outputs on those points land in low-density regions of the target. ``The map ignores the topology'' and ``the student fails near the decision boundary and maps to low density'' are one observation seen from either side.

\textbf{Why a single step makes this hard.} Neither map is discontinuous: the teacher's is a composition of two PF-ODE flow maps and the student is a neural network. What a topology-mismatched correspondence forces is a region of arbitrarily large derivative, where nearby source points must be carried to targets separated by roughly the diameter of the target distribution. The teacher realises that steepness \emph{over its trajectory}, spreading it across many small solver steps, so the correspondence lives in the trajectories rather than in any single map. The student must reproduce the same endpoint map in one forward pass, within an architecture of finite Lipschitz constant, bounded by Assumption~\ref{ass:lip}, the very assumption the guarantee of App.~\ref{supp:ddtheory} rests on. A bounded-Lipschitz approximation of a steep map smooths the steep region, and the smoothed values fall between the two target modes: the empty space between the coils here, the blurred, averaged output of Fig.~\ref{fig:supp_meanpred} on images. The difficulty is intrinsic to collapsing a trajectory into a function, and it lives on a measure-zero set that the plots make visible because they show the whole plane rather than samples.

\textbf{What our components do and do not fix.} The adaptive weighting exists because $\Hat{\mathcal{D}}$ detects this set at no extra cost (Fig.~\ref{fig:result_toy}(b)), and DMCD plus the cycle loss move mis-mapped points off the low-density ridge and back onto the target manifold: the improvement visible from left to right in Fig.~\ref{fig:result_toy}(a). But that is a \emph{density} correction, not a topological one: it relocates points the one-step map placed between modes, and it does not make the correspondence arc-length faithful. We do not claim to have solved this problem. Doing so would mean distilling a teacher whose transport is defined under a manifold-aware cost, as in optimal-transport formulations of unpaired translation~\citep{korotin2022neural, mokrov2024energyguided, choi2025improving}, so that the trajectories already encode arc length; or relaxing ``one continuous map'' to a few steps, or to a student with an explicitly multi-modal output head. Both change the setting rather than the hyperparameters, and neither is evaluated here. The corresponding failures on real images, which we believe have the same origin, are shown without selection in Fig.~\ref{fig:supp_fail}.
}        

\subsection{User Study and Perceptual Metric Evaluations} \label{supp:userstudy}

The results of the user study and perceptual metric evaluations are summarized in Tab.~\ref{table:userstudy}, which complement the main quantitative comparisons presented in Tab.~\ref{table:main}. For the user study, we report the proportion of participants who preferred each baseline over our method (IBCD) in the pairwise comparisons. Values below 0.5 indicate that the majority preferred our method over the corresponding baseline. When multiple variants of a baseline exist, comparisons are made against the default version. Taken together with the results in Tab.~\ref{table:main}, our approach demonstrates consistently strong performance across both automated metrics and human evaluations. We note that perceptual consistency may be further improved by replacing the current L1-based cycle loss with a perceptual loss formulation.

\begin{table}[!t]
\centering
\caption{\textbf{User study and Perceptual Evaluation Results}. User study 1:1 win rates: the proportion of users preferring each baseline over our model (IBCD) in pairwise comparisons. Values below 0.5 indicate that the majority preferred our model. When two versions of a model exist, they are compared against the default version (without $\dagger$).}
\label{table:userstudy}
\resizebox{\columnwidth}{!}{
\begin{small}
\begin{tabular}{lccccccccccc}
\toprule
 & & \multicolumn{3}{c}{\textbf{User Study} (1:1 win rate vs. IBCD)} & \multicolumn{2}{c}{\textbf{Cat$\rightarrow$Dog}} & \multicolumn{2}{c}{\textbf{Wild$\rightarrow$Dog}} & \multicolumn{2}{c}{\textbf{Male$\rightarrow$Female}}\\
\textbf{Method} & \textbf{NFE} & Realism $\uparrow$ & Faithfulness $\uparrow$ & Preference $\uparrow$ & LPIPS $\downarrow$ & CLIP $\uparrow$ & LPIPS $\downarrow$ & CLIP $\uparrow$ & {LPIPS $\downarrow$} & {CLIP $\uparrow$} \\
\midrule
ILVR & 1000 & 0.25 & 0.49 & 0.25 & 0.454 & 72.65 & 0.486 & 67.22 & - & - \\
SDEdit & 1000 & 0.38 & \textbf{0.73} & 0.42 & 0.438 & 73.55 & 0.465 & 67.81 & 0.290 & 53.50 \\
EGSDE & 1000 & 0.37 & 0.28 & 0.43 & 0.433 & 73.98 & 0.467 & 67.34 & 0.284 & 53.12 \\
EGSDE$\dagger$ & 1200 & - & - & - & 0.497 & 73.53 & 0.526 & 66.25 & 0.343 & 51.11 \\
CycleDiffusion & 1000 & 0.25 & \textbf{0.62} & 0.25 & \textbf{0.381} & \textbf{74.77} & 0.417 & \textbf{67.84} & - & - \\
\midrule
DDIB$^*$ (Teacher) & 160 & 0.42 & 0.23 & 0.42 & 0.475 & 73.65 & 0.492 & 67.05 & 0.326 & 56.20 \\
\textbf{IBCD (Ours)} & \textbf{1} & \multicolumn{3}{c}{\textit{compared with each baseline}} & \underline{0.384} & \underline{74.00} & \textbf{0.404} & 66.86 & \textbf{0.261} & \textbf{56.57} \\
\textbf{IBCD$\dagger$ (Ours)} & \textbf{1} & - & - & - & \underline{0.406} & 73.69 & 0.423 & 66.99 & \textbf{0.263} & \textbf{56.53} \\
\bottomrule
\end{tabular}
\end{small}
}
\end{table}

\subsection{Effect of the Auxiliary Loss Weights}
\label{supp:abl_weights}

\begin{figure}[t]
    \centering
    \includegraphics[width=0.7\linewidth]{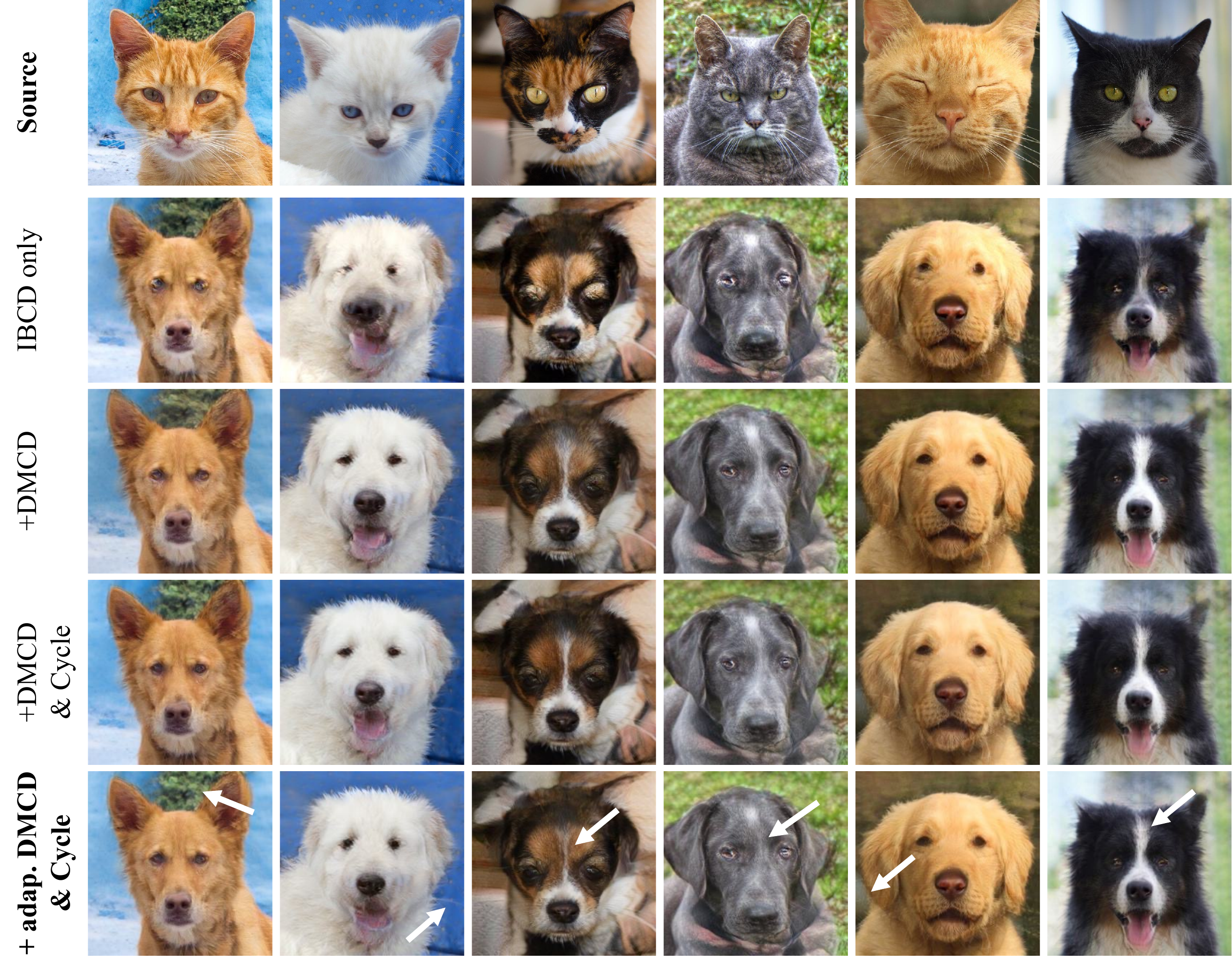}
    \caption{\textbf{Ablation study results on Cat$\rightarrow$Dog task.} The DMCD loss improves the realism of the generated results compared to the vanilla IBCD. Additionally, the cycle translation loss and adaptive DMCD loss enhance source fidelity (as indicated by the arrows). These findings confirm that the components of IBCD work synergistically to achieve a better balance between realism and faithfulness.}
    \label{fig:result_abl}
\end{figure}

\begin{figure}
    \centering
    \includegraphics[width=0.9\textwidth]{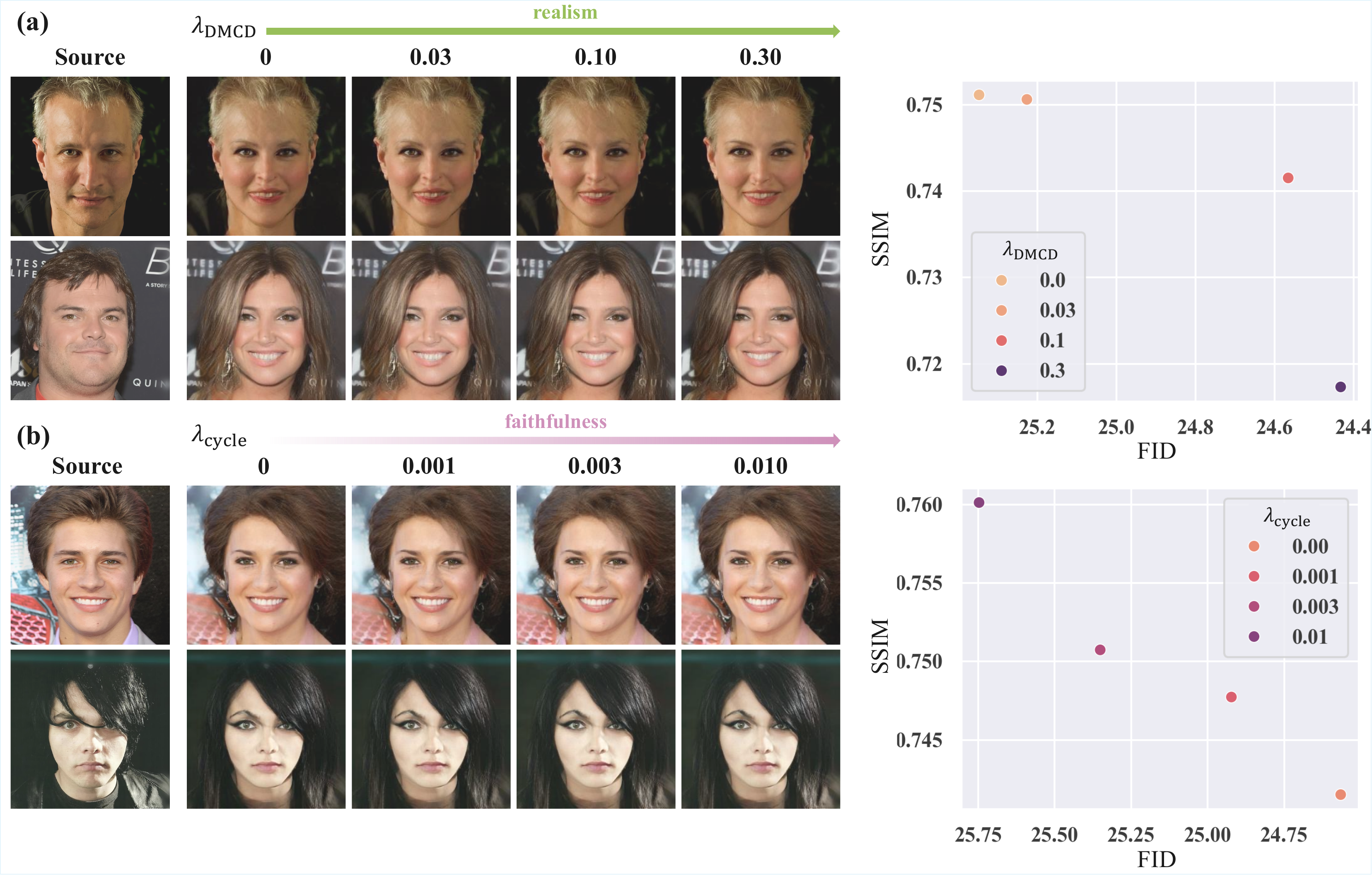}
    \caption{\textbf{Effect of the auxiliary loss weights ($\lambda_\text{DMCD}$, $\lambda_\text{cycle}$) for the Male$\rightarrow$Female task.} In (a) $\lambda_\text{cycle}$ was set to 0, and in (b) $\lambda_\text{DMCD}$ was set to 0.10. Distillation difficulty adaptive weighting was not applied.}
    \label{fig:supp_abl}
\end{figure}

Following the component ablation study of IBCD in the main text and Fig.~\ref{fig:result_abl}, we further investigated the influence of auxiliary loss weights on translation outcomes. Specifically, we varied the weight of the DMCD loss $\lambda_\text{DMCD}$ and the cycle loss $\lambda_\text{cycle}$ in the Male$\rightarrow$Female task (Fig.~\ref{fig:supp_abl}). During these experiments, the distillation difficulty adaptive weighting was not applied. The results aligned with expectations: as $\lambda_\text{DMCD}$ increases, the realism of the translation result improved, while increasing $\lambda_\text{cycle}$ enhanced the faithfulness of the translation.  Thus, in the realism-faithfulness trade-off curve, the DMCD loss emphasizes realism, whereas the cycle loss emphasizes faithfulness.

\subsection{Distance Functions for the IBCD Loss}
\label{supp:abl_loss}

\rev{The main text states this ablation's conclusion in one sentence; the full table is given here.} \rev{The IBCD consistency loss is defined up to a choice of distance function $d(\cdot,\cdot)$ between the student's predictions at adjacent points of the time axis.} Tab.~\ref{table:supp-abl-loss} compares the common choices on Cat$\rightarrow$Dog and Dog$\rightarrow$Cat, each model trained at a smaller resolution for 50K iterations. All four give comparable source faithfulness, but LPIPS attains substantially better FID in both directions, which is why it is the distance used throughout the paper.

\begin{table}[ht!]
\centering
\caption{\textbf{Quantitative ablation study on distance functions for the IBCD loss.} While different distance functions yield comparable source faithfulness, LPIPS achieves substantially better FID. \rev{Trained at a smaller resolution for 50K iterations, so the values are not comparable with Tab.~\ref{table:main}.}}
\label{table:supp-abl-loss}
\renewcommand{\arraystretch}{0.9}
\resizebox{0.62\columnwidth}{!}{
\begin{tabular}{l c c c c c c}
\toprule
& \multicolumn{3}{c}{\textbf{Cat$\rightarrow$Dog}} & \multicolumn{3}{c}{\textbf{Dog$\rightarrow$Cat}} \\
\textbf{Distance}  & \textbf{FID$\downarrow$} & \textbf{PSNR $\uparrow$} & \textbf{SSIM $\uparrow$} & \textbf{FID$\downarrow$} & \textbf{PSNR $\uparrow$} & \textbf{SSIM $\uparrow$}\\
\cmidrule{1-1} \cmidrule{2-7}
LPIPS & 49.13 & 19.70 & 0.690 & 27.59 & 19.30 & 0.666\\ 
L1 & 58.77 & 19.74 & 0.697 & 37.89 & 19.38 & 0.682 \\ 
L2 & 67.82 & 19.92 & 0.709 & 47.95 & 19.36 & 0.686 \\ 
Pseudo-Huber & 59.42 & 20.02 & 0.706 & 39.57 & 19.32 & 0.680 \\ 
\bottomrule
\end{tabular} 
}
\end{table}

\subsection{Approximated Distillation Difficulty in Image-to-image Translation}
\label{supp:ddw_i2i}

\begin{figure}
    \centering
    \includegraphics[width=1.00\textwidth]{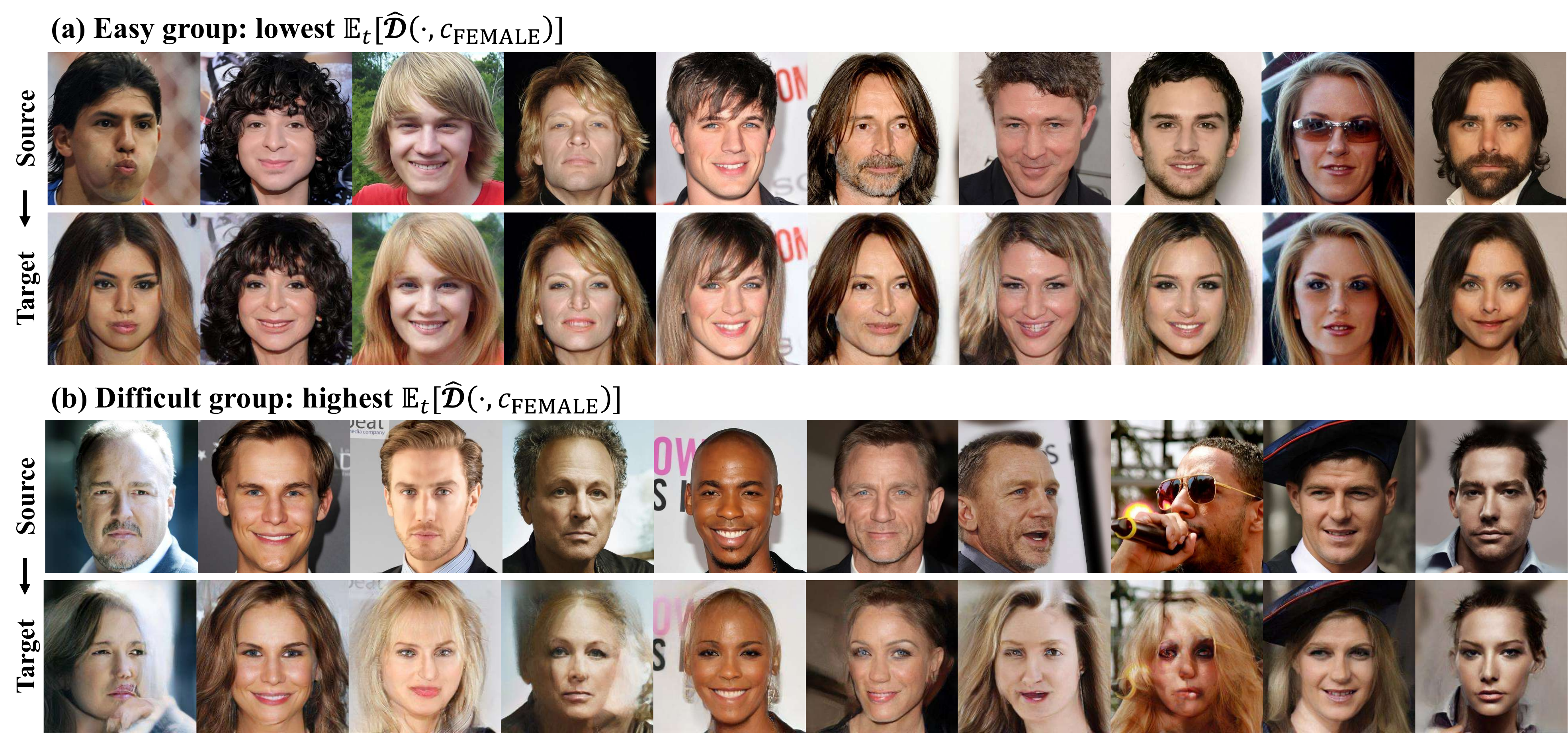}
    \caption{Relationship between self-assessed approximate distillation difficulty $\mathbb{E}_t [\Hat{\mathcal{D}}(\cdot, c_\text{FEMALE})]$ and the translations performed in the Male$\rightarrow$Female task.}
    \label{fig:supp_ddw_vis_i2i}
    \vspace{-1em}
\end{figure}

To explore the implications of the approximated distillation difficulty for real image-to-image translation tasks, we computed an expected approximated distillation difficulty $\mathbb{E}_{t\sim\mathcal{U}[-N+1,N-2]} [\Hat{\mathcal{D}}(\mbx_t, c_\text{FEMALE})]$ for all trajectories generated with the DDIB teacher in the Male$\rightarrow$Female task using the vanilla IBCD model. We then selected the trajectories with the top 10 and bottom 10 approximate distillation difficulties and performed Male$\rightarrow$Female translation using the vanilla IBCD model for these trajectories, as shown in Fig.~\ref{fig:supp_ddw_vis_i2i} without cherry-picking. The results indicate that the IBCD model struggles to effectively transform source images from trajectories with high approximate distillation difficulty into target images compared to those with low approximate distillation difficulty. Specifically, the translation results within the top 10 distillation difficulty groups exhibit relatively inferior image quality, highlighting the impact of distillation difficulty on translation performance.

\subsection{Training Stability}
\label{supp:stability}

\begin{figure}
    \centering
    \includegraphics[width=0.9\textwidth]{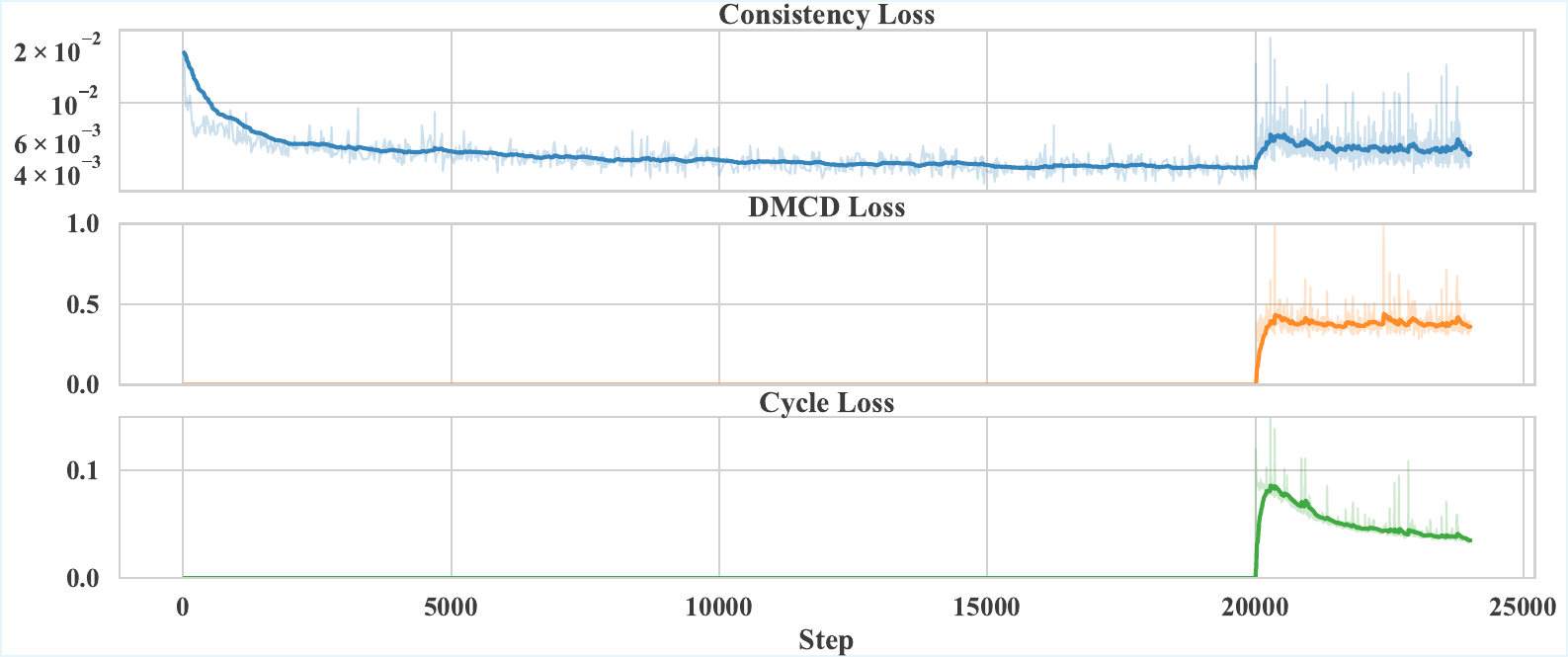}
    \caption{\textbf{Visualization of the training loss curve highlighting the stability of our framework on the Cat$\rightarrow$Dog task.} \rev{The DMCD and cycle losses are introduced at the transition step, which for this task is 200K optimization steps (Tab.~\ref{table:hyperparam}); the horizontal axis of the plot is the logging index of the run rather than the optimization step counter, so the switch appears at the corresponding logging index rather than at 200K. Before the switch both auxiliary losses are identically zero and the corresponding networks are not instantiated (App.~\ref{supp:trainingcost}).} }
    \label{fig:supp_wandb}
    \vspace{-1em}
\end{figure}

Our framework does not rely on adversarial losses, which are known to be unstable and difficult to tune, giving it a clear advantage in training stability. To demonstrate this, we present the training loss curves in Fig.~\ref{fig:supp_wandb}. As shown in the figure, both the consistency loss and the cycle loss steadily decrease throughout training, with a brief fluctuation only when the auxiliary losses are introduced. The DMCD loss also shows a stable plateau, as expected, since the fake score is computed by a jointly trained fake-score model. Overall, the loss trajectories indicate that our framework trains smoothly and remains stable throughout the process.

{\addc   
\subsection{\texorpdfstring{\rev{Comparison Under a Shared Teacher Diffusion Model}}{Comparison Under a Shared Teacher Diffusion Model}}
\label{supp:same_teacher}

Most baseline numbers in Tab.~\ref{table:main} come from EGSDE~\citep{zhao2022egsde}, where the diffusion-based methods use a 93.56M-parameter DDPM backbone, whereas IBCD and its DDIB teacher use a 32.95M-parameter EDM (Tab.~\ref{table:supp-inference-time}); the comparison could therefore be read as confounding the translation method with the backbone. Two measurements bear on this.

First, the two backbones are comparable as generators of the target domain despite the difference in size: Tab.~\ref{table:supp-teacher-backbone} gives essentially the same FID for both (40.91 against 41.88), ours with roughly a third of the parameters and fewer training steps. Our teacher is therefore not a stronger starting point that would explain the results on its own.

Second, Tab.~\ref{table:supp-same-teacher} re-runs the strongest diffusion baselines on top of \emph{our} EDM teacher for Cat$\rightarrow$Dog, so the rows share one pre-trained model and differ only in what is done with it. IBCD$^\dagger$ takes the best FID in the group at a single NFE, and both IBCD configurations are far more source-faithful than any iterative baseline. We note honestly that CycleDiffusion on the same teacher lands between our two configurations in FID, but at 160 NFE, with the lowest PSNR of any row and an SSIM well below either IBCD configuration.

\begin{table}[ht]
\addc
\centering
\small
\begin{minipage}{0.46\textwidth}
\addc
\centering
\captionof{table}{\addc \textbf{Generation quality of the two backbones (AFHQ-dog).} Our EDM teacher and the DDPM backbone used by the diffusion baselines of Tab.~\ref{table:main} are comparable as generators, at about a third of the parameters.}
\label{table:supp-teacher-backbone}
\resizebox{\linewidth}{!}{
\begin{tabular}{lrrr}
\toprule
\textbf{Diffusion Model} & \textbf{Parameters [M]} & \textbf{Training Steps [M]} & \textbf{FID} $\downarrow$ \\
\midrule
EDM (teacher of ours) & 32.95 & 0.8 & 40.91 \\
\makecell[l]{DDPM (ILVR, SDEdit, \\ EGSDE, CycleDiffusion)} & 93.56 & 1.0 & 41.88 \\
\bottomrule
\end{tabular}
}
\end{minipage}%
\hspace{0.04\textwidth}
\begin{minipage}{0.46\textwidth}
\addc
\centering
\captionof{table}{\addc \textbf{Cat$\rightarrow$Dog under the same teacher diffusion model.} Every row uses our EDM teacher, so the rows differ only in the translation method built on it.}
\label{table:supp-same-teacher}
\resizebox{\linewidth}{!}{
\begin{tabular}{lrrrr}
\toprule
\textbf{Model} & \textbf{NFE} $\downarrow$ & \textbf{FID} $\downarrow$ & \textbf{PSNR} $\uparrow$ & \textbf{SSIM} $\uparrow$ \\
\midrule
SDEdit (EDM)         & 80  & 59.60 & 17.36 & 0.378 \\
EGSDE (EDM)          & 80  & 56.84 & 17.77 & 0.370 \\
CycleDiffusion (EDM) & 160 & 45.97 & 14.43 & 0.558 \\
\midrule
\textbf{IBCD (ours)}               & \textbf{1} & 47.44          & \textbf{19.50} & \textbf{0.701} \\
\textbf{IBCD$^\dagger$ (ours)}     & \textbf{1} & \textbf{44.77} & 18.04 & 0.663 \\
\bottomrule
\end{tabular}
}
\end{minipage}
\end{table}
}        

\subsection{Model Inference Efficiency}
\label{supp:inference}

\begin{table}
\centering
\small
\caption{\textbf{Quantitative comparison of model inference times and parameter sizes}. Our method achieves higher translation quality with high parameter and inference-time efficiency.}
\label{table:supp-inference-time}
\resizebox{0.85\textwidth}{!}{
\begin{tabular}{lcccc}
\toprule
\textbf{Method} & \textbf{Parameters [M]} & \textbf{NFE} $\downarrow$ & \textbf{Time [$s/img$] $\downarrow$} & \textbf{Relative Time $\downarrow$}\\
\midrule
StarGAN v2~\citep{choi2020stargan}   & 64.45 & 1 & 0.0052 & 0.43 \\
CUT~\citep{park2020contrastive}  & 11.39 & 1 & 0.0070 & 0.58  \\
\midrule
NOT~\citep{korotin2022neural} & 9.72 & 1 & 0.0026 & 0.22 \\
DSBM~\citep{shi2023diffusion} & 131.02 & 100 & 4.36 & 363.33 \\
UNSB~\citep{kim2023unpaired} &  14.69 & 5 & 0.077 & 6.42   \\
ASBM~\citep{gushchin2024adversarial} & 79.58 & 4 & 0.069 & 5.71 \\
Eg-EOT~\citep{mokrov2024energyguided} & 26.21 & 100 & 9.87 & 822.51 \\
DIOTM~\citep{choi2025improving} & 39.65 & 1 & 0.013 & 1.08 \\
\midrule
ILVR~\citep{choi2021ilvr}    & 93.56 & 1000 & 13.40  & 1116.67  \\
SDEdit~\citep{meng2022sdedit}   & 93.56 & 1000 & 6.78 & 565.00   \\
EGSDE~\citep{zhao2022egsde} & 147.14 & 1000 & 15.89 & 1324.16 \\
CycleDiffusion~\citep{wu2023latent} & 187.12 & 1000(+100) & 26.03 & 2169.17 \\
\midrule
GPT-Image-1 (Foundation)~\citep{openai_image_generation_api_2025} & $\gg1$ & $\gg1$ & 30.32 & 2526.67 \\ 
DDIB (Teacher)~\citep{su2022dual} & 32.95 & 160 & 1.45 & 120.83 \\
\textbf{IBCD (Ours)} & 32.95 & 1 & 0.012 & 1 \\
\bottomrule
\end{tabular}
}
\end{table}

To reflect real-world constraints such as model size and inference algorithms, we conducted an inference speed comparison experiment. Instead of relying solely on NFE comparisons, we measured the actual inference time for a Cat$\rightarrow$Dog task on a single NVIDIA GeForce RTX 4090 GPU (except for GPT-Image-1) with a batch size of 1. Tab.~\ref{table:supp-inference-time} presents the average inference time per image, the relative time for each methodology, and the number of model parameters. The results demonstrate that our methodology is computationally efficient in real-world sampling scenarios, while also using substantially fewer parameters than diffusion-based baseline methods. Some faster methods, such as GAN-based approaches and NOT, exhibit translation quality that limits the relevance of direct efficiency comparisons. \rev{A condensed version of Tab.~\ref{table:supp-inference-time}, restricted to the representative baselines discussed there, is reported in the main paper; the complete comparison, including every method of Tab.~\ref{table:main}, is retained here and is the version to cite for the full picture. The corresponding \emph{training}-side accounting is given separately in App.~\ref{supp:trainingcost}.}

{\addc   
\subsection{\texorpdfstring{\rev{Adversarial Distillation of DDIB}}{Adversarial Distillation of DDIB}}
\label{supp:gan_baseline}

Distilling the same DDIB teacher into a one-step student with an adversarial objective, and comparing that student against IBCD, is the right comparison to ask for. \citet{zheng2025revisiting} show that an adversarial objective on a diffusion-pre-trained backbone yields a strong one-step generator after little fine-tuning, and \citet{pix2pix} is the standard conditional discriminator such a variant would use.

\textbf{What we ran.} We kept the IBCD consistency loss, removed $\gL_\text{DMCD}$, and put an adversarial term in its place, so that a discriminator supplies the distribution-level signal that the distribution-matching regularizer supplies in IBCD. This is our framework with the distribution-matching regularizer swapped for a discriminator: it is neither a GAN trained from scratch nor our objective with an adversarial term added on top of $\gL_\text{DMCD}$. The variant used the same teacher, the same schedule and the same number of steps as the runs reported in Tab.~\ref{table:main}, on Cat$\leftrightarrow$Dog.

\textbf{Outcome.} Training diverged. The consistency objective and the adversarial objective pull in different directions, and the run did not reach a usable student, so there is no number to report for it.

\textbf{What we take from it.} We read this as a statement about the discriminator rather than about the comparison. Making the variant work would require engineering on the discriminator itself, its capacity and its design, and we judge that out of scope for this paper; it is the natural next step for this line of work. We therefore make no claim that IBCD outperforms a well-tuned GAN-distilled DDIB. What we claim is narrower and is what Tab.~\ref{table:main} and Tab.~\ref{table:supp-same-teacher} support: one-step bidirectional translation, at the reported quality, with no adversarial objective anywhere in training.

A purely adversarial variant, dropping the consistency term as well, does not fit this framework naturally. DDIB is a deterministic ODE map, so running the teacher over the source data yields \emph{paired} endpoints that a discriminator plus a regression term could be trained on directly; but the intermediate states then disappear, and with them the distillation difficulty of Sec.~\ref{sec:ibcd2} that our weighting is defined on. Such a student would also need a conditional discriminator per target domain, and would not by itself give the \emph{shared} bidirectional generator that in IBCD follows from the two boundary conditions of one consistency function on a signed time axis.

\textbf{Evidence already in the paper.} Tab.~\ref{table:main} evaluates adversarially trained one- and few-step translators under the same protocol (StarGAN v2, CUT and DIOTM at 1 NFE, ASBM at 4, UNSB at 5), and their FIDs bound the range such methods currently reach on these benchmarks; but they are not distilled from our teacher, so they isolate neither the objective nor the backbone. App.~\ref{supp:same_teacher} controls the other variable: Tab.~\ref{table:supp-same-teacher} fixes the teacher and varies the method, where a converged adversarial student would fix the teacher and vary the \emph{objective}. Tab.~\ref{table:supp-inference-time} settles the deployment side in advance, since such a student would have the same parameter count at 1 NFE.
}        

\subsection{Failure Cases}
\label{supp:failure}
\begin{figure}
    \centering
    \includegraphics[width=0.75\textwidth]{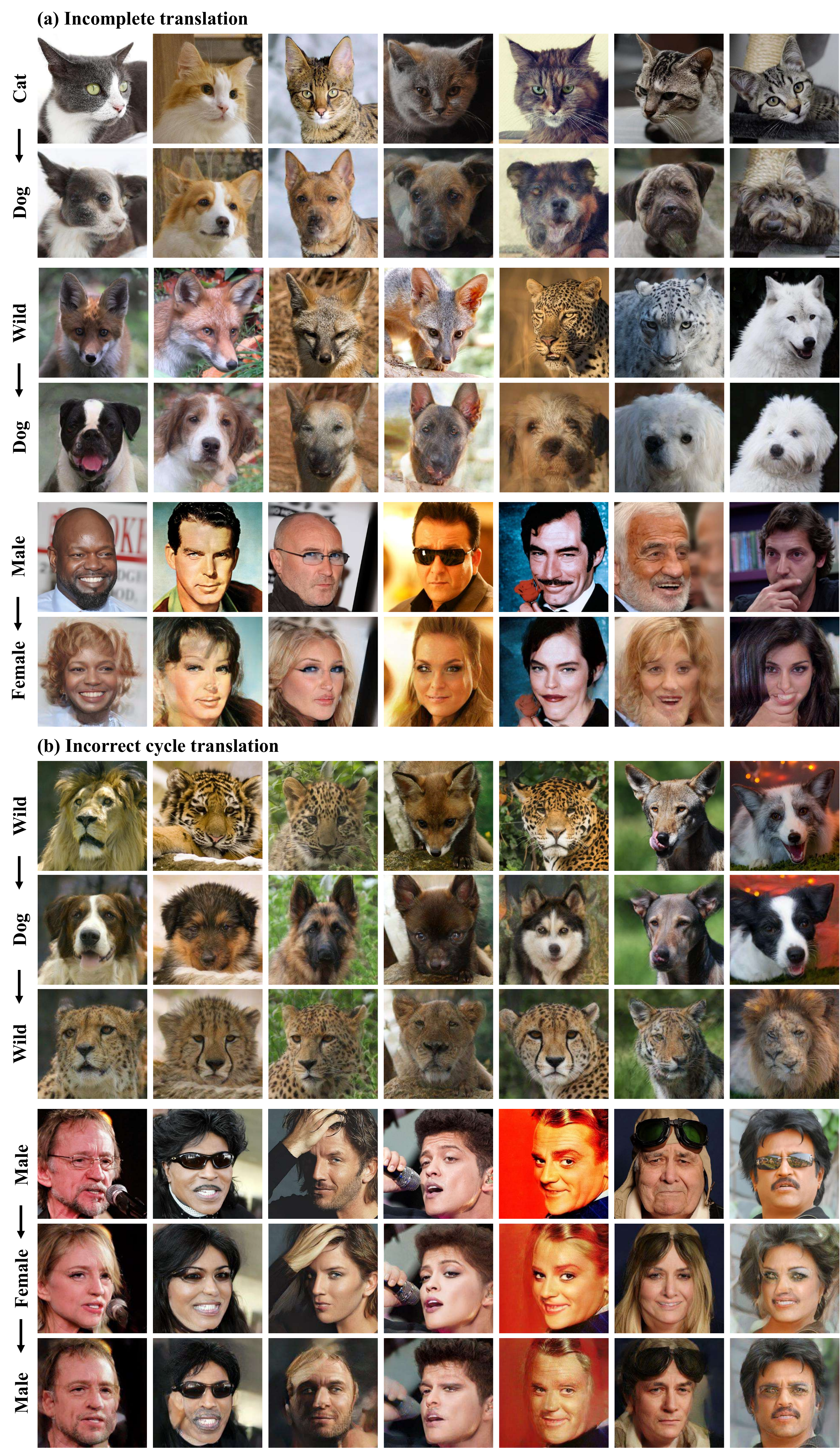}
    \caption{{Example of failure cases, which are (a) incomplete translation and (b) incorrect cycle translation.}}
    \label{fig:supp_fail}
\end{figure}

IBCD occasionally produces failure cases as illustrated in Fig.~\ref{fig:supp_fail}. The primary failures can be attributed to incomplete translations (Fig.~\ref{fig:supp_fail}(a)) and incorrect cycle translations (Fig.~\ref{fig:supp_fail}(b)), which are likely due to distillation errors and the side effects of auxiliary losses. Distillation errors from the CD, in particular, appear to be the primary reason. The DMCD and cycle translation loss can also contribute to these issues, with the former leading to incorrect cycle translations and the latter to incomplete translations. Minimizing distillation errors through improved distillation methods and advanced weighting strategies for auxiliary losses might address this issue.

\subsection{{Bidirectional Translations}}
\label{supp:bidirectional}

To evaluate IBCD's bidirectional translation capabilities, we compared it to baseline methods through two tasks: \textit{opposite translation} and \textit{cycle translation}. Opposite translation involves reversing the main translation task (Dog$\rightarrow$Cat, Dog$\rightarrow$Wild, Female$\rightarrow$Male), while cycle translation involves performing the reverse task after the main translation (Cat$\rightarrow$Dog$\rightarrow$Cat, Wild$\rightarrow$Dog$\rightarrow$Wild, Male$\rightarrow$Female$\rightarrow$Male). To ensure a fair comparison of bidirectional performance, we used the same model and sampling hyperparameters for each domain pair (Cat$\leftrightarrow$Dog, Wild$\leftrightarrow$Dog, Male$\leftrightarrow$Female) in both opposite and cycle translation tasks.

Given the limited number of models capable of bidirectional translation, we selected StarGAN v2~\citep{choi2020stargan}, CycleDiffusion~\citep{wu2023latent}, and DDIB (teacher)~\citep{su2022dual} as baselines. We measured FID for the final target domain for the cycle translation task. It's worth noting that StarGAN v2's inference process differs from its main translation task (Tab.~\ref{table:main}) performed by \citet{zhao2022egsde} for a fairer comparison. It inputs the same source image as both the source and reference images, enabling it to achieve both high realism and faithfulness.

\begin{figure}
    \centering
    \includegraphics[width=0.8\textwidth]{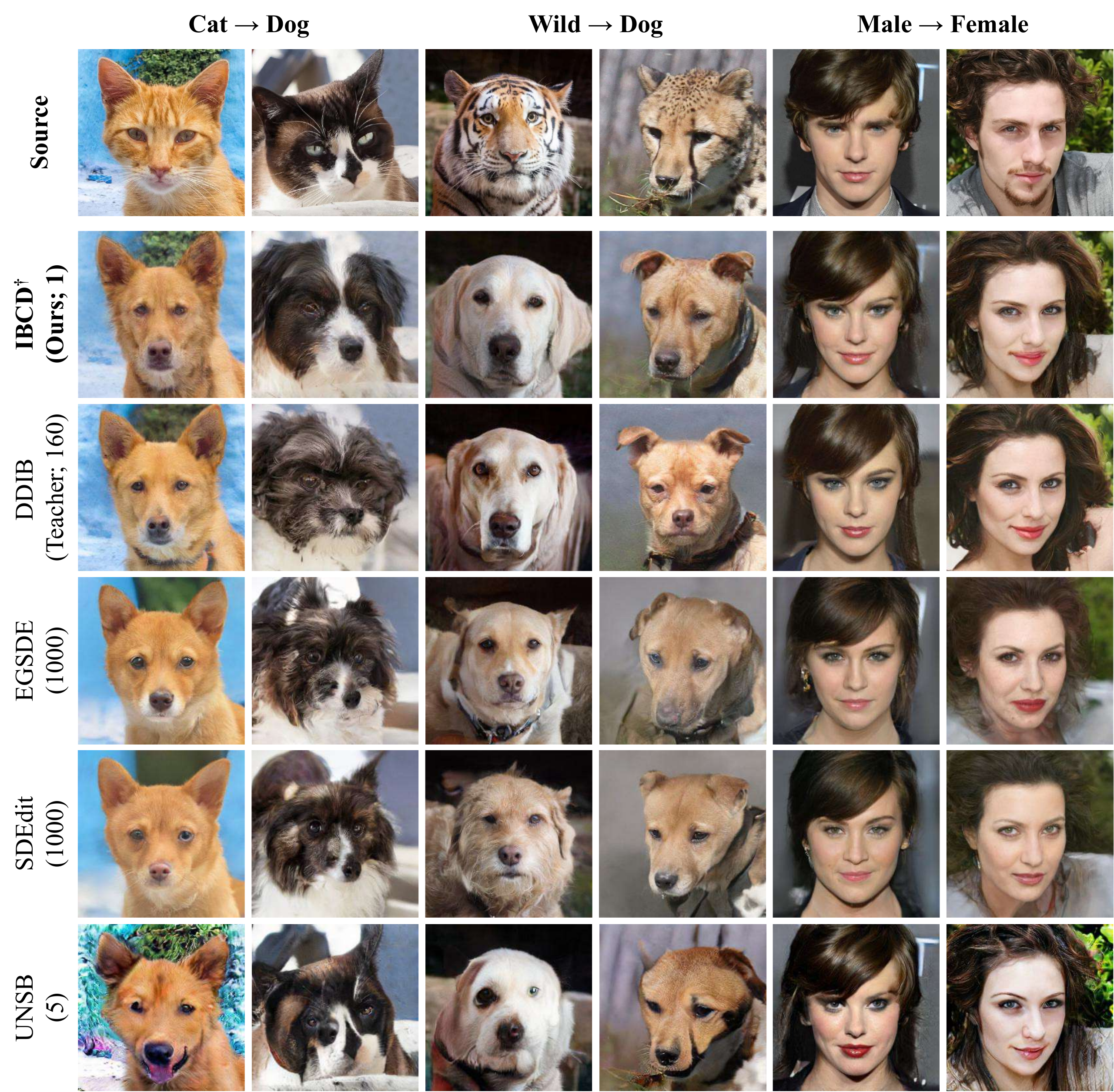}
    \caption{\textbf{Extra qualitative comparison of unpaired image-to-image translation tasks.} Compared to other baselines, our model achieves more realistic and source-faithful translations in a single step. The numbers in parentheses represent inference NFE.}
    \label{fig:supp_result_main}
\end{figure}

\begin{figure}
    \centering
    \begin{overpic}[width=0.8\textwidth]{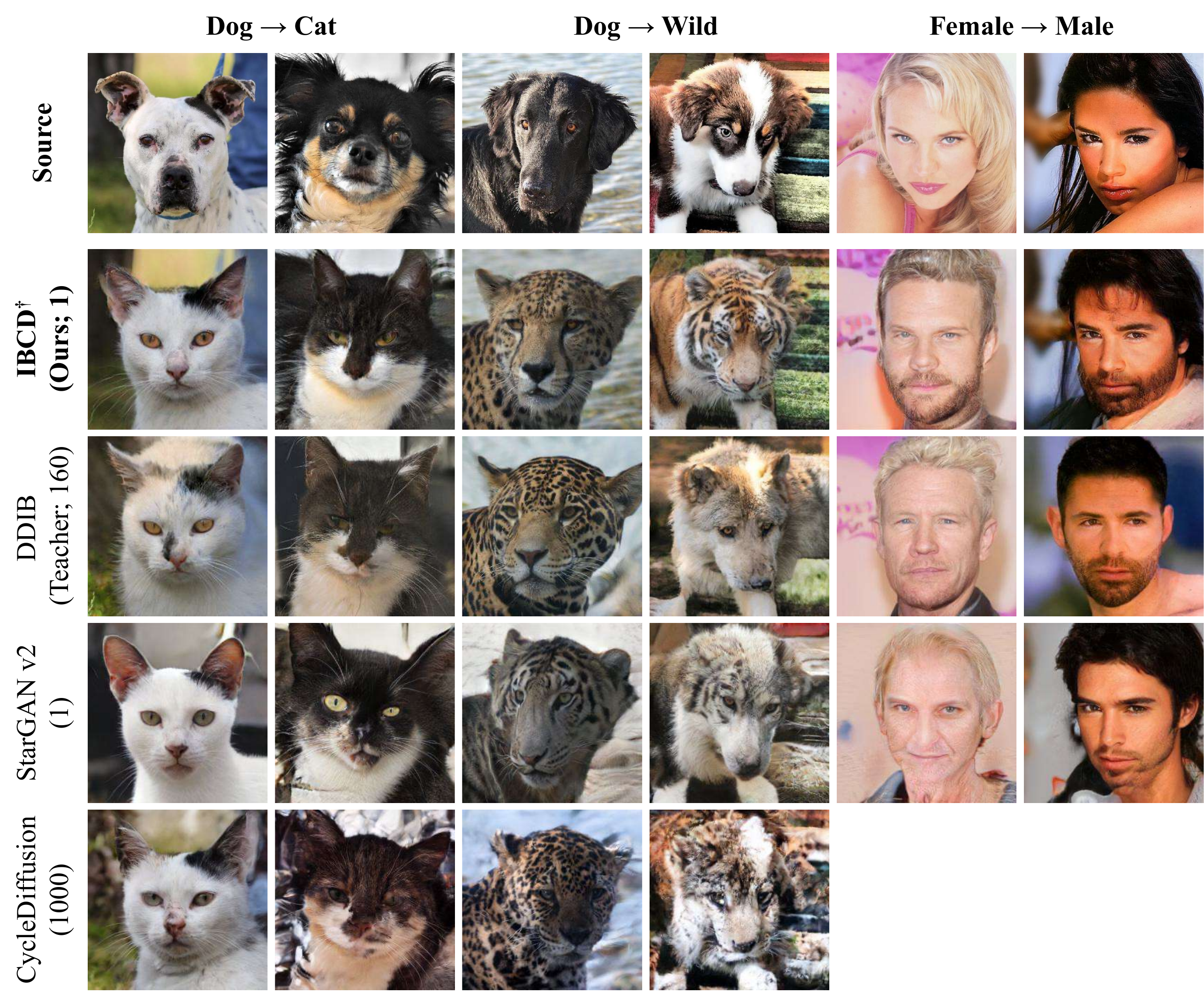}
        \put(77,8){\makebox(0,0){\footnotesize N/A}}
        \put(92.5,8){\makebox(0,0){\footnotesize N/A}}
    \end{overpic}
    \caption{\textbf{Qualitative comparison of unpaired image-to-image translation tasks (opposite translation)}. Compared to other baselines, our model achieves more realistic and source-faithful translations in a single step. The numbers in parentheses represent inference NFE.}
    \label{fig:supp_opposite}
\end{figure}

\begin{figure}
    \centering
    \begin{overpic}[width=0.8\textwidth]{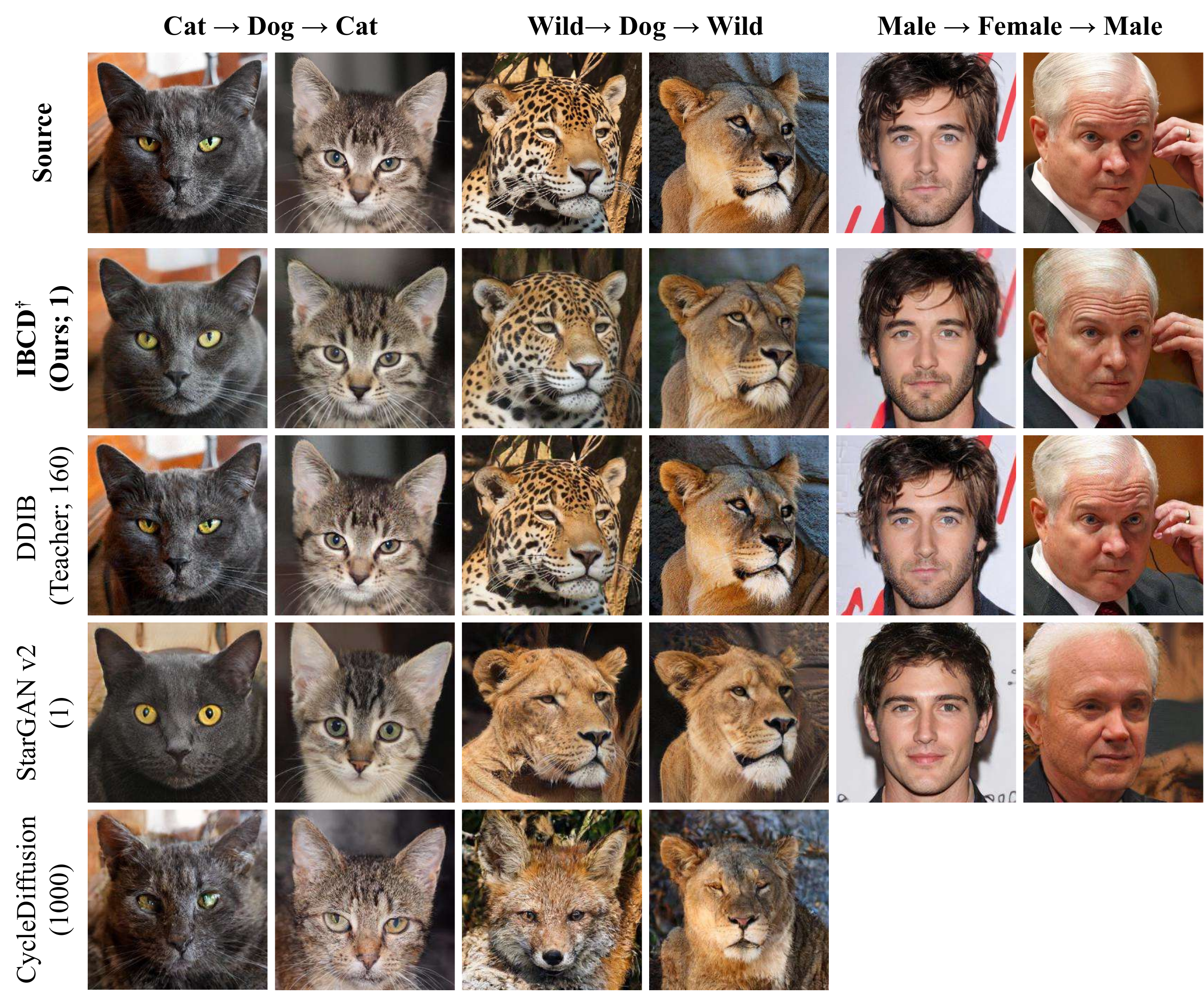}
    \put(77,8){\makebox(0,0){\footnotesize N/A}}
    \put(92.5,8){\makebox(0,0){\footnotesize N/A}}
    \end{overpic}
    \caption{{\textbf{Qualitative comparison of unpaired image-to-image translation tasks (cycle translation)}. Compared to other baselines, our model achieves consistent cycle translations in a single step. The numbers in parentheses represent inference NFE.}}
    \label{fig:supp_cycle}
\end{figure}

First, Fig.~\ref{fig:supp_result_main} provides additional qualitative comparisons for the main translation tasks across all datasets. Tab.~\ref{table:supp-opposite-cycle} and Fig.~\ref{fig:supp_opposite},~\ref{fig:supp_cycle} demonstrate that our model also excels in reverse and cycle translation tasks, exhibiting the best performance and high efficiency. This further supports its strong bidirectional translation capabilities.

\begin{table}
\centering
\scriptsize
\caption{\textbf{Quantitative comparison of unpaired image-to-image translation tasks (opposite \& cycle translation)}. The opposite-direction tasks used the same model and inference hyperparameters as the main direction task using bi-directionality.}
\resizebox{0.80\textwidth}{!}{
\begin{small}
\begin{tabular}{lcccccc}
\toprule
\textbf{Method} & NFE $\downarrow$ & FID $\downarrow$ & PSNR $\uparrow$ & SSIM $\uparrow$ & Density $\uparrow$ & Coverage $\uparrow$  \\
\midrule
\multicolumn{7}{c}{\textbf{Dog$\rightarrow$Cat}}\\
\midrule
StarGAN v2~\citep{choi2020stargan} & 1 & 37.73 & 16.02 & 0.399 & 1.336 & 0.778 \\
CycleDiffusion~\citep{wu2023latent} & 1000(+100) & 40.45 & 17.83 & 0.493 & 1.064 & 0.774 \\
DDIB (Teacher)~\citep{su2022dual} & {160} & {\color{gray} 30.28} & {\color{gray} 17.15} &  {\color{gray} 0.597} & {\color{gray} 2.071} &  {\color{gray} 0.902} \\
\textbf{IBCD (Ours)} & 1 & 28.99 & \textbf{19.10} & \textbf{0.695} & 1.699 & 0.894 \\
\textbf{IBCD$\dagger$ (Ours)} & 1 & \textbf{28.41} & 17.40 & 0.653 & \textbf{2.112} & \textbf{0.920} \\
\midrule
\multicolumn{7}{c}{\textbf{Dog$\rightarrow$Wild}}\\
\midrule

StarGAN v2~\citep{choi2020stargan} & 1 & 49.35 & 16.17 & 0.386 & 0.772 & 0.478 \\
CycleDiffusion~\citep{wu2023latent} & 1000(+100) & 27.01 & 16.99 & 0.421 & 0.816 & 0.752 \\
DDIB (Teacher)~\citep{su2022dual} & {160} & {\color{gray} 13.20} & {\color{gray} 16.80} &  {\color{gray} 0.583} & {\color{gray} 1.202} &  {\color{gray} 0.760} \\
\textbf{IBCD (Ours)} & 1 & 18.79 & \textbf{17.56} & \textbf{0.671} & 0.900 & \textbf{0.830 }\\
\textbf{IBCD$\dagger$ (Ours)} & 1 & \textbf{16.67} & 16.22 & 0.646 & \textbf{1.058} & 0.814 \\

\midrule
\multicolumn{7}{c}{\textbf{Female$\rightarrow$Male}}\\
\midrule
StarGAN v2~\citep{choi2020stargan} & 1 & 59.56 & 15.75 & 0.465 & 1.145 & 0.587 \\
DDIB (Teacher)~\citep{su2022dual} & {160} & {\color{gray} 26.98} & {\color{gray} 18.74} &  {\color{gray} 0.668} & {\color{gray} 1.154} &  {\color{gray} 0.858} \\
\textbf{IBCD (Ours)} & 1 & \textbf{31.28} & 19.93 & \textbf{0.733} & 1.300 & 0.808  \\
\textbf{IBCD$\dagger$ (Ours)} & 1 & 31.49 & \textbf{19.51} & 0.726 & \textbf{1.311} & \textbf{0.809} \\

\midrule
\multicolumn{7}{c}{\textbf{Cat$\rightarrow$Dog$\rightarrow$Cat}}\\
\midrule
StarGAN v2~\citep{choi2020stargan} & 1 & 30.53 & 16.30 & 0.382 & 1.717 & 0.890 \\
CycleDiffusion~\citep{wu2023latent} & 1000(+100) & 39.59 & 19.01 & 0.434 & 0.731 & 0.676 \\
DDIB (Teacher)~\citep{su2022dual} & {160} & {\color{gray} 16.56} & {\color{gray} 25.88} &  {\color{gray} 0.804} & {\color{gray} 1.330} &  {\color{gray} 0.990} \\
\textbf{IBCD (Ours)} & 1 & \textbf{22.42} & \textbf{22.35} & \textbf{0.767} & 1.322 & \textbf{0.992} \\
\textbf{IBCD$\dagger$ (Ours)} & 1 & 24.03 & 20.28 & 0.724 & \textbf{1.749} & 0.988 \\
\midrule
\multicolumn{7}{c}{\textbf{Wild$\rightarrow$Dog$\rightarrow$Wild}}\\
\midrule

StarGAN v2~\citep{choi2020stargan} & 1 & 37.76 & 15.30 & 0.285 & 1.102 & 0.566  \\
CycleDiffusion~\citep{wu2023latent} & 1000(+100) & 19.43 & 16.39 & 0.281 & 0.649 & 0.616 \\
DDIB (Teacher)~\citep{su2022dual} & {160} & {\color{gray} 6.75} & {\color{gray} 26.08} &  {\color{gray} 0.803} & {\color{gray} 1.118} &  {\color{gray} 0.974} \\
\textbf{IBCD (Ours)} & 1 & \textbf{9.89} & \textbf{20.56} & \textbf{0.739} & 1.118 & \textbf{0.972} \\
\textbf{IBCD$\dagger$ (Ours)} & 1 & 10.66 & 18.80 & 0.693 & \textbf{1.259} & 0.968 \\

\midrule
\multicolumn{7}{c}{\textbf{Male$\rightarrow$Female$\rightarrow$Male}}\\
\midrule
StarGAN v2~\citep{choi2020stargan} & 1 & 57.80 & 15.39 & 0.502 & 1.634 & 0.728 \\
DDIB (Teacher)~\citep{su2022dual} & {160} & {\color{gray} 28.29} & {\color{gray} 27.70} &  {\color{gray} 0.853} & {\color{gray} 0.821} &  {\color{gray} 0.993} \\
\textbf{IBCD (Ours)} & 1 & \textbf{39.84} & \textbf{22.22} & \textbf{0.790} & \textbf{1.341} & 0.979  \\
\textbf{IBCD$\dagger$ (Ours)} & 1 & 39.96 & 21.85 & 0.783 & 1.332 & \textbf{0.984} \\

\bottomrule
\end{tabular}
\end{small}
}

\label{table:supp-opposite-cycle}
\end{table}

\subsection{Unpaired MRI Contrast Translation}
\label{supp:mri}

We conducted experiments on the BraTS2021 dataset~\citep{baid2021rsna} to evaluate our model on the brain MRI contrast translation task, demonstrating its applicability in the medical imaging domain. We used T1- and T2-weighted brain MRI scans from the dataset (excluding other contrasts) and performed unpaired translation from T1- to T2-weighted images at a resolution of 256×256. Although the dataset provides paired images, we utilized them in an unpaired setting during training. The dataset was split into 1,126 volumes (174,530 images) for training and 10 volumes (1,550 images) for validation. Quantitative and qualitative comparisons with existing baselines (Tab.~\ref{table:supp-mri}, \rev{repeated as Tab.~\ref{table:mri} of the main paper,} and Fig.~\ref{figure:supp-mri}) demonstrate the effectiveness of our method, highlighting its potential for practical deployment in clinical workflows. \rev{Because the BraTS2021 volumes are paired, the PSNR and SSIM reported here are computed against the \emph{true} T2-weighted image rather than against the source, and therefore measure translation fidelity directly, unlike the source-referenced values of Tab.~\ref{table:main}.}

\setlength{\tabcolsep}{1pt}
{
\centering
\begin{minipage}{.33\columnwidth}
    \centering
    \captionof{table}{Quantitative comparison of unpaired MRI contrast translation tasks. \rev{PSNR/SSIM are measured against the paired ground-truth target-contrast image. Reproduced compactly as Tab.~\ref{table:mri} of the main paper.}}
    \label{table:supp-mri}
    \resizebox{\columnwidth}{!}{
\begin{small}
\begin{tabular}{lcc}
\toprule
 & \multicolumn{2}{c}{\textit{compared with G.T.}} \\
\textbf{Method}  & PSNR $\uparrow$ & SSIM $\uparrow$ \\
\midrule
SDEdit & 29.85 & 0.849 \\
EGSDE & 23.34 & 0.842 \\
CycleDiffusion & 30.52 & 0.825 \\
\midrule
GPT-Image-1 (Foundation) & \multicolumn{2}{c}{\color{gray} \textit{No valid translation}} \\ 
DDIB$^*$ (Teacher) & {\color{gray} 30.24} & {\color{gray} 0.825}  \\
\textbf{IBCD (Ours)}  & \textbf{33.28} & \textbf{0.855}   \\
\bottomrule
\end{tabular}
\end{small}
}
\end{minipage}%
\hspace{0.02\columnwidth}%
\begin{minipage}{0.64\columnwidth}
    \centering
    \includegraphics[width=\columnwidth]{figs/fig_supp_brats.pdf}
    \captionof{figure}{Qualitative comparison of unpaired MRI contrast translation tasks.}
    \label{figure:supp-mri}
\end{minipage}
}

\subsection{Limitations of Foundational Models in Domain-Specific Translation}
\label{supp:foundation}

The rise of foundational diffusion-based image editing methods~\citep{parmar2023zero, hertz2023delta, hertz2022prompt}, particularly those coupled with large language models, has enabled intuitive zero-shot multi-turn editing~\citep{openai_image_generation_api_2025, qwen_vlo_2025}. Despite their appeal, unpaired I2I translation remains essential, especially in domains where paired data is scarce, such as medical and scientific imaging~\citep{kaji2019overview, chen2023deep}.

\begin{figure}
    \centering
    \includegraphics[width=1.00\textwidth]{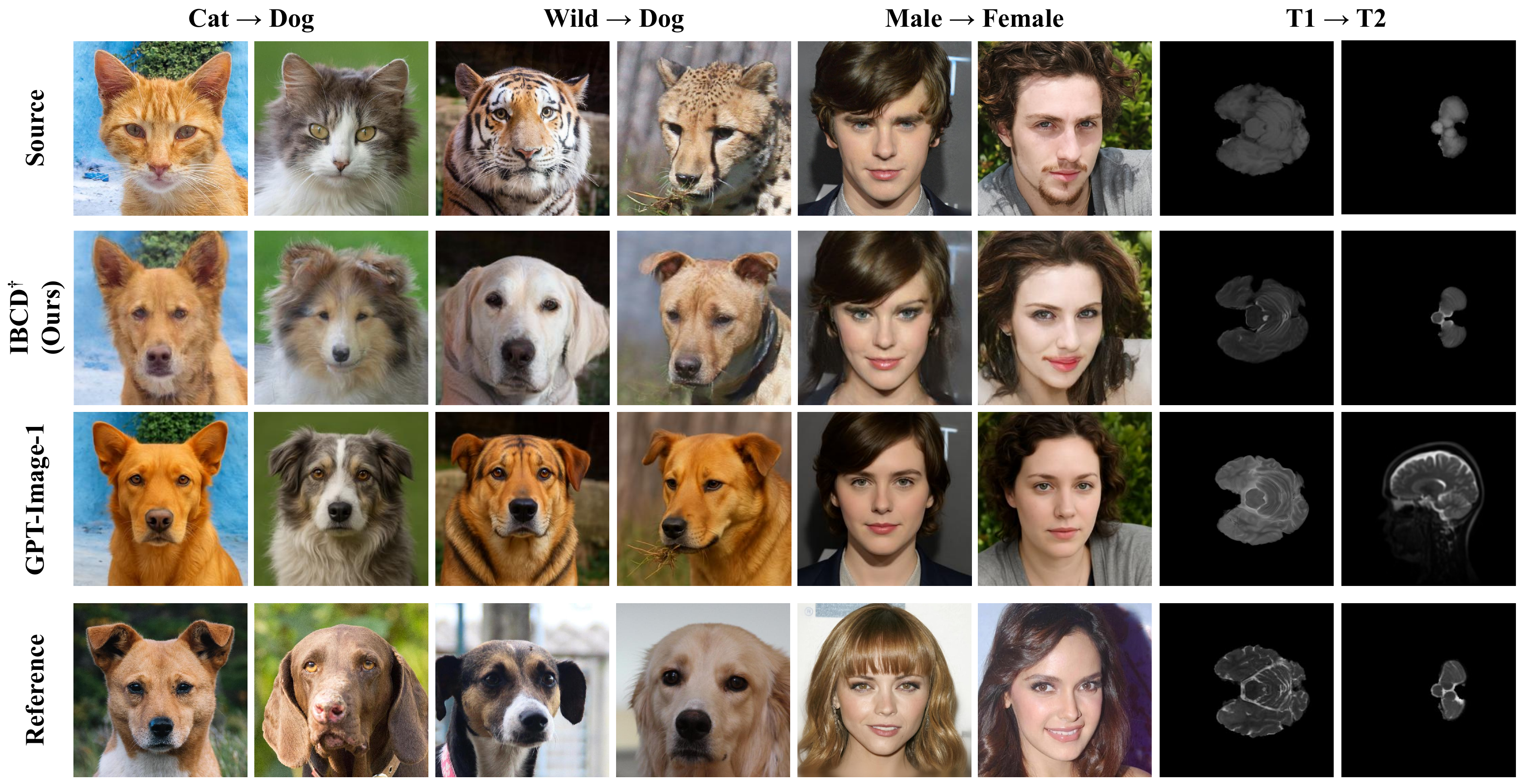}
    \caption{Limitations of foundational models in domain-specific translation.}
    \label{fig:supp_gpt}
\end{figure}

These tasks demand specific, accurate knowledge of both source and target domains, along with fine-grained detail preservation and low-latency performance. Foundational models, while flexible, lack such specialization. As shown in Fig.~\ref{fig:supp_gpt}, GPT-Image-1~\citep{openai_image_generation_api_2025} often generates outputs that deviate significantly from the target distribution under zero-shot conditions, producing results that may appear unrealistic, incomplete, or exaggerated. This tendency is especially pronounced in medical scenarios like MRI translation, where the model frequently fails to produce clinically meaningful outputs. These limitations are further supported by the quantitative results in Tab.~\ref{table:main} and Tab.~\ref{table:supp-mri}.

We do not claim our method outperforms foundational models across the board. Instead, we argue that unpaired I2I approaches remain indispensable in scenarios where zero-shot methods fall short due to their lack of domain-specific adaptation.

\subsection{More Qualitative Results}
\label{supp:morequal}
\begin{figure}
    \centering
    \includegraphics[width=0.8\textwidth]{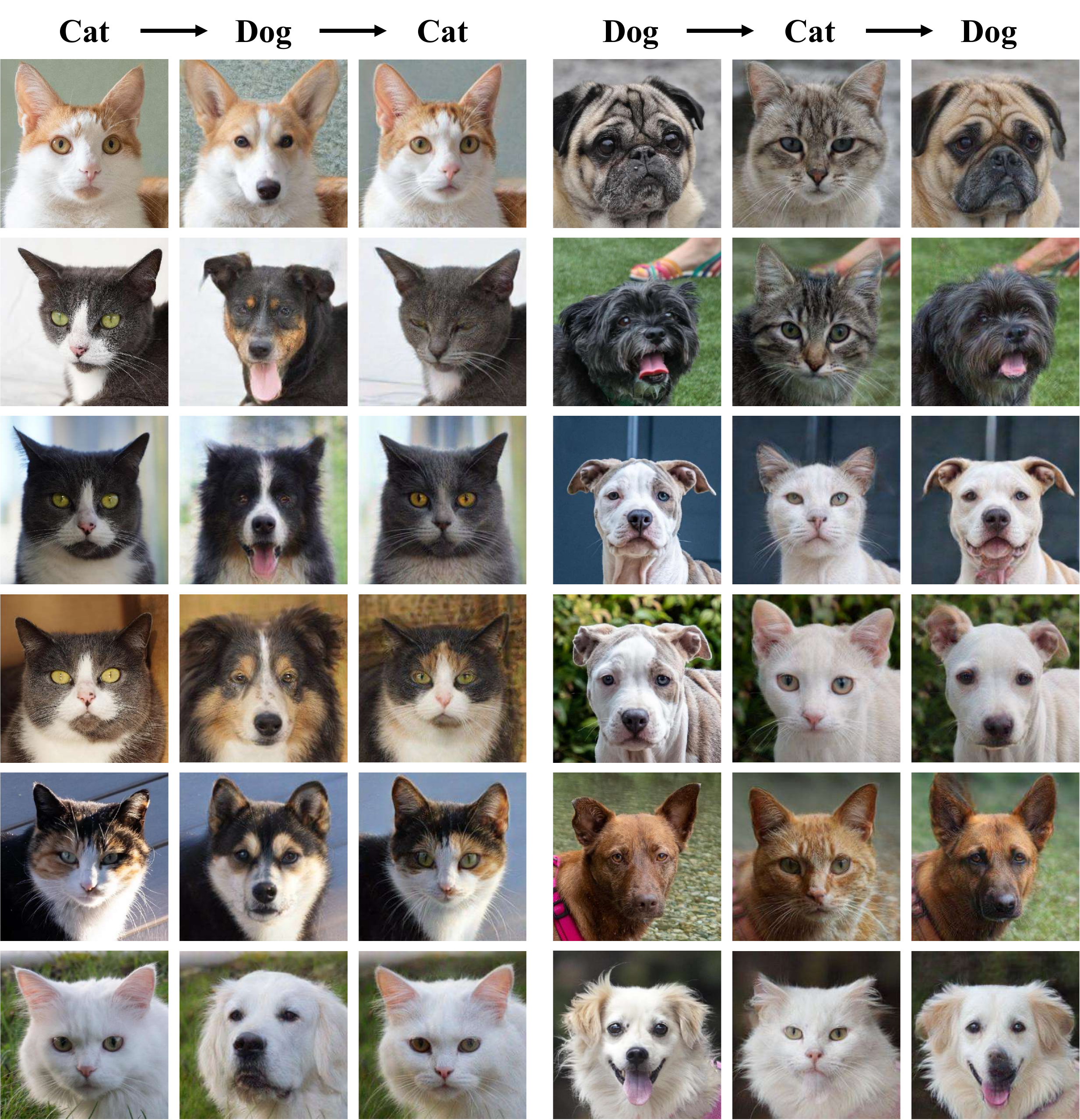}
    \caption{Result of the bi-directional cycle translation with a single model for the Cat$\leftrightarrow$Dog task (IBCD$^\dagger$).}
    \label{fig:supp_cycle_c2d}
\end{figure}

\begin{figure}
    \centering
    \includegraphics[width=0.8\textwidth]{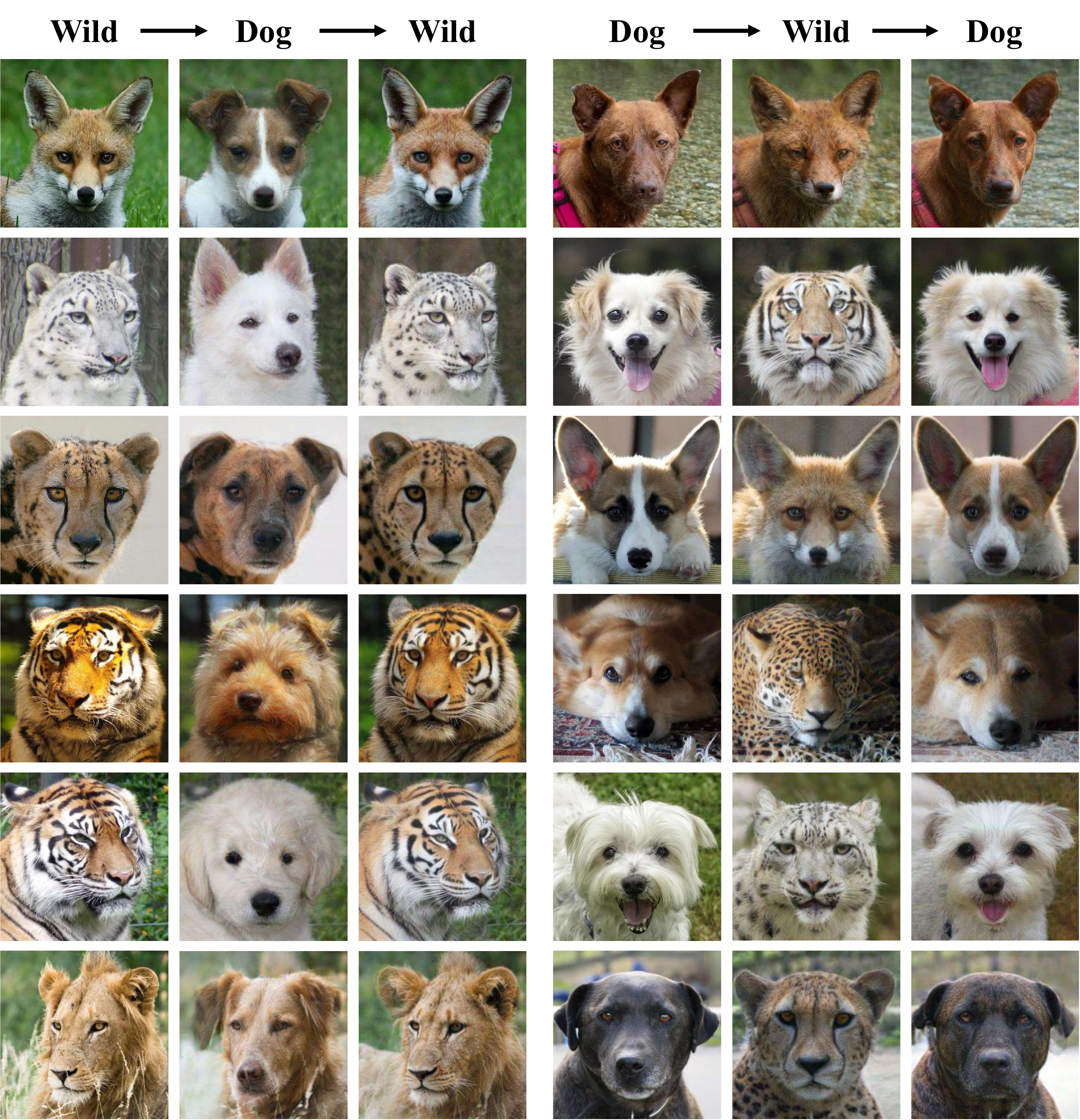}
    \caption{Result of the bi-directional cycle translation with a single model for the Wild$\leftrightarrow$Dog task (IBCD$^\dagger$).}
    \label{fig:supp_cycle_w2d}
\end{figure}

\begin{figure}
    \centering
    \includegraphics[width=0.8\textwidth]{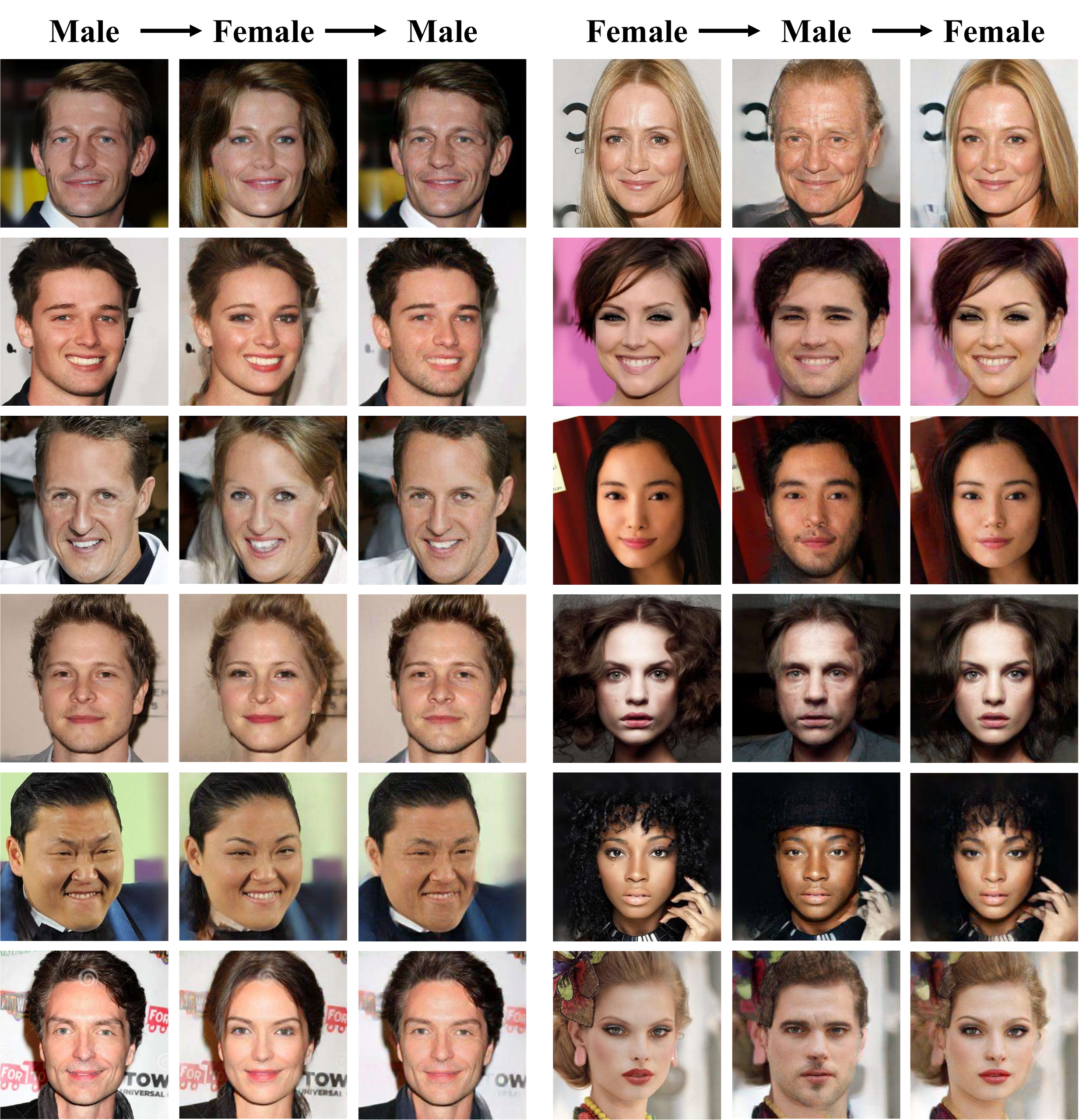}
    \caption{Result of the bi-directional cycle translation with a single model for the Male$\leftrightarrow$Female task (IBCD$^\dagger$).}
    \label{fig:supp_cycle_m2f}
\end{figure}

In this section, we present additional qualitative results obtained through cycle translation tasks (Cat$\rightarrow$Dog$\rightarrow$Cat, Wild$\rightarrow$Dog$\rightarrow$Wild, Male$\rightarrow$Female$\rightarrow$Male). The results of the Cat$\leftrightarrow$Dog, Wild$\leftrightarrow$Dog, and Male$\leftrightarrow$Female models are illustrated in Fig.~\ref{fig:supp_cycle_c2d},~\ref{fig:supp_cycle_w2d},~\ref{fig:supp_cycle_m2f}. These results highlight our model's one-way and bidirectional translation capabilities.

\clearpage

\end{document}